\documentclass[lettersize,journal]{IEEEtran}
\usepackage{amsmath,amssymb,amsfonts,bm}

\usepackage{algorithm}
\usepackage[noEnd=false,indLines=true]{algpseudocodex}
\algrenewcommand\algorithmicrequire{\textbf{Input:}}
\algrenewcommand\algorithmicensure{\textbf{Output:}}
\algrenewcommand\algorithmicfunction{\textbf{Function}}
\algrenewcommand\algorithmicend{\textbf{End}}
\algrenewcommand\algorithmicfor{\textbf{For}}
\algrenewcommand\algorithmicif{\textbf{If}}
\algrenewcommand\algorithmicelse{\textbf{Else}}
\algrenewcommand\algorithmicreturn{\textbf{Return}}
\algrenewcommand\textproc[1]{\texttt{#1}}

\usepackage{array}
\usepackage[caption=false,font=normalsize,labelfont=sf,textfont=sf]{subfig}
\usepackage{textcomp}
\usepackage{stfloats}
\usepackage{url}
\usepackage{verbatim}
\usepackage{graphicx}
\usepackage{booktabs}
\usepackage{tabularx}
\usepackage{amssymb}
\usepackage{multirow}
\usepackage{multicol}
\usepackage{cuted}
\usepackage{capt-of}
\newcommand{\cOne}{\cellcolor{orange!85}}
\newcommand{\cTwo}{\cellcolor{orange!50}}
\newcommand{\cThree}{\cellcolor{orange!15}}
\renewcommand{\arraystretch}{1.2}
\usepackage[table]{xcolor}

\usepackage[
    bookmarks=true,
    colorlinks=false,
    linkbordercolor={1 0 0}, 
    citebordercolor={0 1 0}, 
    urlbordercolor={0 0 1}, 
    pdfborder={0 0 1}
]{hyperref}
\def\BibTeX{{\rm B\kern-.05em{\sc i\kern-.025em b}\kern-.08em
    T\kern-.1667em\lower.7ex\hbox{E}\kern-.125emX}}
\usepackage{balance}
\begin{document}
\title{CF-Splat: Closed-Form Triangle Splatting with Mesh Priors}
\author{
Yixin Fang$^1$, Liang Zhao$^{1,\dagger}$% <-this % stops a space
\thanks{$^1$The University of Edinburgh}% <-this % stops a space
% \thanks{$^*$Anonymous Email}
% \thanks{$^{1}$Albert Author is with Faculty of Electrical Engineering, Mathematics and Computer Science,
%         University of Twente, 7500 AE Enschede, The Netherlands
%         {\tt\small albert.author@papercept.net}}%
% \thanks{$^{2}$Bernard D. Researcheris with the Department of Electrical Engineering, Wright State University,
%         Dayton, OH 45435, USA
%         {\tt\small b.d.researcher@ieee.org}}%
}

% \markboth{Journal of \LaTeX\ Class Files,~Vol.~18, No.~9, September~2020}%
% {How to Use the IEEEtran \LaTeX \ Templates}

\maketitle

\begin{abstract}
Splatting-based radiance-field representations have recently gained popularity in robotic perception, enabling high-quality novel-view synthesis (NVS) through explicit scene primitives. However, existing splatting methods typically rely on iterative nonlinear optimization that jointly optimizes millions of coupled primitive parameters, resulting in substantial computation time, peak GPU memory consumption, and susceptibility to local minima. In this paper, we propose CF-Splat, a novel closed-form triangle-splatting framework for efficient scene appearance estimation over a known mesh either pre-existing or provided by upstream mapping or reconstruction pipelines. Our key insight is that, by properly formulating the problem over the given mesh geometry and calibrated cameras, the rendered pixel colors can be expressed linearly with respect to the per-vertex spherical harmonics (SH) coefficients, allowing multi-view scene appearance estimation to be formulated as a global linear least-squares (LLS) problem without the need of iterative nonlinear optimization. To guarantee a well-posed solution under uneven multi-view observation coverage, we introduce adaptive SH regularization that constrains insufficiently supported high-order SH coefficients and yields a unique global minimizer. To ensure the scalability to large-scale scenes, we develop a triangle-structured matrix-free preconditioned conjugate-gradient solver that avoids explicitly constructing the large sparse normal system while producing the same LLS solution. Experiments on diverse real-world datasets demonstrate that CF-Splat achieves competitive or superior NVS quality compared with state-of-the-art splatting methods. While, CF-Splat requires approximately 27.5$\times$–378.0$\times$ less computation time and 1.8$\times$ lower peak GPU memory on average than 3DGS across scenes of different scales on a consumer-grade laptop.
Interactive live demos are available on an anonymous website%
\footnote{Anonymous Website: \url{https://github.com/Yixin-F}. Please allow about 5--10 minutes for the workspace to initialize the GPU resources.}. We will release the source code upon acceptance of this paper. 
\end{abstract}


\section{Introduction}
\IEEEPARstart{R}{adiance-field} representations have become increasingly important in robotic perception, including augmented reality~\cite{azinovic2022neural}, photorealistic simulation~\cite{yang2023unisim}, and autonomous navigation~\cite{adamkiewicz2022vision}. In these settings, an effective scene representation should capture reliable geometry while efficiently modeling appearance, enabling high-quality novel-view synthesis (NVS) with low computational overhead for downstream robotic tasks. Neural Radiance Field (NeRF)~\cite{mildenhall2021nerf} methods represent scenes as continuous volumetric functions parameterized by multilayer perceptrons (MLPs) and have demonstrated remarkable NVS quality, but typically require hours of per-scene optimization. Recently, 3D Gaussian Splatting (3DGS)~\cite{kerbl20233d} has achieved rendering quality comparable to NeRF while substantially reducing optimization time and enabling real-time rendering. It explicitly represents a scene as a set of anisotropic 3D Gaussian primitives and employs a differentiable tile-based rasterizer. However, 3DGS still faces four major bottlenecks: (1) its nonlinear optimization is susceptible to local minima, causing the solution to depend on initialization and optimization settings; (2) the optimized Gaussians do not accurately conform to the underlying scene surfaces; (3) per-scene optimization requires tens of thousands of iterations, making it still computationally expensive; and (4) the training process typically needs substantial GPU memory overhead.

\begin{figure}[!t]
    \centering
    \includegraphics[width=3.4in]{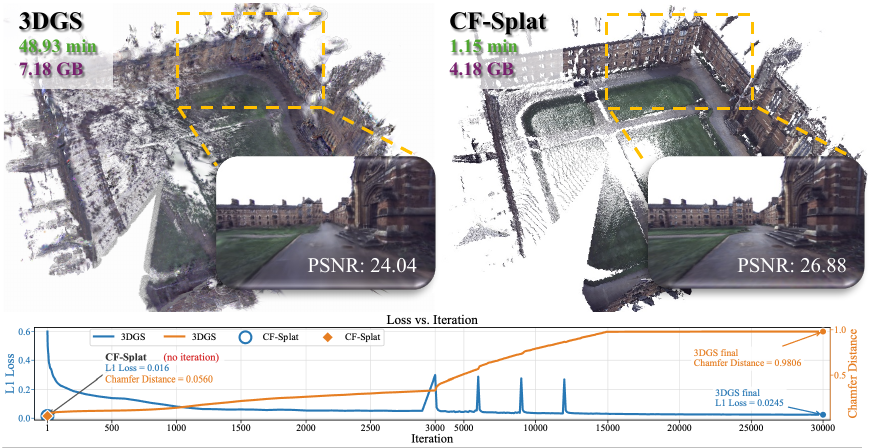}
    \captionof{figure}{Evaluation of CF-Splat against 3DGS~\cite{kerbl20233d} on the \textit{college} scene from the large-scale Oxford Spires dataset~\cite{tao2026oxford}. 
    The input consists of 760 views at 1080$\times$1440 resolution and a mesh with 1.55M vertices reconstructed using the SDF produced by VDBFusion~\cite{vizzo2022vdbfusion} and the LiDAR poses estimated by HBA~\cite{liu2023large}. Given the same input geometry source, 3DGS jointly optimizes millions of coupled Gaussian parameters through \textit{30K iterations} to minimize the photometric residual (\textcolor{orange}{L1 loss}), which improves image-space fitting at the cost of increased geometric error (\textcolor{blue}{Chamfer distance}). In contrast,
    CF-Splat directly admits a closed-form solution \textit{without any iteration}, while achieving 2.84 dB higher PSNR on the visualized test view, requiring
    only 1.15 min of \textcolor{green}{computation time} and 4.18 GB of \textcolor{purple}{peak GPU memory}, compared with 48.93 min and 7.18 GB for 3DGS.}
    \label{fig:teaser}
\end{figure}

These limitations of 3DGS primarily lie in its training pipeline, which \textit{jointly} optimizes millions of coupled Gaussian parameters, including geometry parameters such as positions, opacities, and covariances, as well as appearance parameters represented by spherical harmonics (SH), through iterative gradient-based optimization on the Adam optimizer~\cite{kingma2014adam}. During this nonlinear optimization shown in Fig.~\ref{fig:teaser}, additional Gaussians are progressively introduced, moved, reshaped, and recolored to better fit the appearance with photometric-only residual at each \textit{per-view} iteration. This inevitably increases optimization time and GPU memory overhead, while potentially trapping the optimization in view-specific appearance overfitting and degrading the underlying geometric structure.

Existing methods have attempted to alleviate these limitations from different perspectives. Multi-view regularization-based methods~\cite{du2024mvgs} introduce cross-view constraints to improve global consistency. Geometry-aware representations~\cite{huang20242d,chen2024pgsr} employ more compact primitives and surface-aligned constraints to better conform to the underlying geometric structure. Triangle-integrated approaches~\cite{waczynska2024games,held2026triangle,held2025meshsplatting}
incorporate mesh constraints or directly adopt triangle-based representations to achieve a better balance between geometry and appearance fitting. More recently, efficiency-oriented
methods~\cite{mallick2024taming,hanson2025speedy,ren2026fastgs} regulate Gaussian densification and pruning to suppress redundant primitives, thereby reducing computational and memory costs. In parallel, numerical optimization methods~\cite{hollein2025_3dgslm,pehlivan2026matrixfree} explore LM-based and matrix-free solvers to efficiently compute parameter updates within iterative nonlinear optimization. Despite these advances, they do not fundamentally alter the iterative nonlinear optimization framework of 3DGS, leaving its inherent bottlenecks unresolved. Therefore, a clear gap remains toward a closed-form splatting formulation for more efficient appearance estimation. 

To this end, we propose CF-Splat, a novel triangle-splatting framework that leverages a mesh prior to enable closed-form scene appearance estimation. In many real-world robotic systems, this scene geometry prior is readily available such as pre-operative CT scans in surgical robotics or CAD models for object manipulation, or can be obtained from upstream mapping or reconstruction pipelines, such as SLAM systems~\cite{ruan2023slamesh,jia2025cad,tian2022kimera,liu2024dynameshslam} and SDF-based reconstruction methods~\cite{dai2017bundlefusion,oleynikova2017voxblox,vizzo2022vdbfusion}. Rather than re-optimizing geometry together with appearance, by properly formulating view-dependent appearance over the given mesh and calibrated cameras, we can establish a linear relationship between rendered pixel colors and per-vertex spherical harmonics (SH) coefficients. Based on this, scene appearance estimation can be formulated as a global linear least-squares (LLS) problem and a closed-form solution can be obtained without any iteration. To further ensure a unique global minimizer, we introduce adaptive SH regularization for uneven observation coverage to make the LLS problem well posed. The proposed LLS formulation also results in a block-sparse structure corresponding to mesh topology that leads us to develop an efficient triangle-structured matrix-free preconditioned conjugate-gradient (PCG) solver, extending its scalability across different scene sizes. Extensive experiments on diverse real-world datasets demonstrate the accuracy and efficiency of CF-Splat. 

Our main contributions are summarized as follows: 

\begin{enumerate}
    \item We propose CF-Splat, a novel closed-form triangle-splatting framework for
    efficient scene appearance estimation over a known mesh.
    
    \item We derive a global LLS formulation for scene appearance estimation by exploiting the linear dependence of rendered pixel colors on per-vertex SH coefficients, thereby eliminating iterative nonlinear optimization and substantially reducing computation cost.
    
    \item We introduce adaptive SH regularization to address uneven multi-view
    observation coverage, resulting in a well-posed LLS problem with a unique
    global minimizer and improved cross-view stability of the solution.
    
    \item We develop a triangle-structured matrix-free PCG solver that avoids
    explicitly constructing the global normal system, reducing GPU memory consumption and enabling the proposed LLS formulation to large-scale scenes.
    
    \item We extensively evaluate CF-Splat on diverse real-world datasets against state-of-the-art splatting methods. CF-Splat achieves competitive or even superior NVS quality while maintaining high computational efficiency on a consumer-grade laptop. The source code will be released upon the acceptance of this paper.
\end{enumerate}

% \begin{figure}[!t]
%     \centering
%     \includegraphics[width=3.4in]{pipeline.png}
%     \captionof{figure}{Overview of CF-Splat. The blue regions highlight the closed-form
%     scene appearance estimation and the triangle-structured matrix-free
%     PCG solver, presented in Sections~\ref{sec:closed_form_formulation}
%     and~\ref{sec:scalable_solver}, respectively. The dashed branch denotes pre-system gradient-guided triangle densification.}
%     \label{fig:pipeline}
% \end{figure}

% Fig.~\ref{fig:pipeline} illustrates the flowchart of applying our CF-Splat in practice. The blue components represent our core approaches, while the dashed portions are preprocessing modules.

The remainder of this paper is organized as follows.
Section~\ref{sec:related_work} reviews related works. Section~\ref{sec:closed_form_formulation} presents our proposed closed-form scene appearance estimation.
Section~\ref{sec:scalable_solver} presents the triangle-structured matrix-free PCG solver.
Section~\ref{sec:experiments} presents the comprehensive experiments.
Finally, Section~\ref{sec:conclusion} concludes the paper with
its limitations and future directions.

\section{Related Work}
\label{sec:related_work}

In this section, we discuss some related works on neural radiance fields and explicit splatting representations, accelerated and memory-efficient splatting methods, optimization strategies for differentiable rendering, and classical mesh-based view-dependent appearance modeling.

\subsection{Neural Radiance Fields and Explicit Splatting}

NeRF~\cite{mildenhall2021nerf} represents scene density and view-dependent radiance using continuous neural functions optimized from posed images. Although capable of high-quality NVS, volumetric rendering and per-scene neural optimization remain computationally expensive. 3DGS~\cite{kerbl20233d} improves efficiency through visibility-aware rasterization of explicit anisotropic Gaussian primitives, whose geometry, opacity, and SH appearance coefficients are jointly optimized with adaptive densification.

Subsequent methods improve geometric fidelity using more surface-aware primitives and constraints. 2DGS~\cite{huang20242d} replaces volumetric Gaussians with oriented planar primitives and introduces geometric regularization, while PGSR~\cite{chen2024pgsr} exploits locally planar Gaussian structures with multi-view geometric constraints. MVGS~\cite{du2024mvgs} further incorporates multi-view regulated learning to improve cross-view consistency.

Explicit surface structures have also been incorporated into splatting representations. GaMeS~\cite{waczynska2024games} parameterizes Gaussian primitives with respect to triangles, allowing them to follow an underlying mesh while retaining differentiable Gaussian rendering. Triangle-Splat~\cite{held2026triangle} directly adopts triangles as rendering primitives, whereas Mesh-Splat~\cite{held2025meshsplatting} introduces connected and opaque mesh structures compatible with conventional graphics pipelines.

Despite their surface-aligned representations, these methods still estimate scene parameters through iterative nonlinear optimization. Moreover, incorporating additional constraints further increases optimization burdens. In contrast, CF-Splat builds upon a known mesh and formulates scene appearance estimation as an efficient closed-form problem.

\subsection{Accelerated and Memory-Efficient Gaussian Splatting}

The large number of Gaussian primitives generated during 3DGS optimization has motivated extensive research on reducing training, rendering, and storage costs. One direction focuses on compacting the final representation. Compact3DGS~\cite{lee2024compact} removes redundant Gaussians and compresses Gaussian attributes using compact appearance representations and vector quantization, while LightGaussian~\cite{fan2024lightgaussian} prunes insignificant primitives and further compresses SH and geometric attributes through distillation and quantization.

Another direction directly targets computational and memory costs during training. Taming-3DGS~\cite{mallick2024taming} controls Gaussian growth through resource-aware densification under prescribed primitive and memory budgets. Speedy-Splat~\cite{hanson2025speedy} combines sparse pixel processing with aggressive pruning to accelerate training and rendering, while FastGS~\cite{ren2026fastgs} regulates densification and pruning using multi-view consistency to suppress redundant Gaussian growth and shorten training time.

Although these approaches significantly improve efficiency, their gains mainly arise from reducing the number of Gaussians within the same gradient-based pipeline from 3DGS. In contrast, CF-Splat improves efficiency by replacing per-view gradient-based optimization with a global LLS formulation that does not require any iteration, substantially reducing computational cost.

\subsection{Optimization Strategies and Solvers for Splatting}

Most splatting methods formulate scene reconstruction as nonlinear optimization. In particular, 3DGS~\cite{kerbl20233d} and many of its derivatives employ first-order optimizers such as Adam~\cite{kingma2014adam} to update Gaussians through differentiable rendering. Although effective and highly parallelizable, their convergence and computational cost depend on initialization, learning rates, iteration schedules, densification, and other optimization settings.

Recent studies have explored more advanced numerical strategies for 3DGS. 3DGS$^2$~\cite{lan2025_3dgs2} replaces first-order updates with local Newton steps using small per-parameter systems. 3DGS-LM~\cite{hollein2025_3dgslm} introduces a Levenberg-Marquardt (LM) optimizer with GPU-parallelized PCG, while LM-RS~\cite{pehlivan2026matrixfree} further exploits sparse residual dependencies through matrix-free Jacobian-vector products and conjugate-gradient solves. More recently, 3DGS$^2$-TR~\cite{hsiao2026_3dgs2tr} adopts a matrix-free trust-region strategy, and Opt3DGS~\cite{huang2026opt3dgs} combines stochastic exploration with curvature-aware quasi-Newton guidance.

Despite their different numerical strategies, these methods retain a nonlinear formulation, where Newton, LM, or conjugate-gradient procedures compute local updates within an outer iterative optimization process. However in CF-Splat, a scalable solver is developed based on PCG for sufficiently solving large-scale global LLS problem. The proposed matrix-free solver avoids explicitly constructing the global normal system, reducing GPU memory overhead while converging to the same LLS solution within a prescribed numerical tolerance.

\subsection{Mesh-Based View-dependent Appearance Modeling}

Classical surface light field (SLF) methods model view-dependent appearance directly on reconstructed surfaces. \cite{wood2000surface} estimated directional radiance from registered images using regularized least squares, while \cite{chen2002light} introduced vertex-centered light fields on triangular meshes with barycentric interpolation. Later methods improved compactness and robustness by representing directional appearance with SH and estimating their coefficients through least-squares fitting~\cite{palma2013surface,vanhoey2013robust}. Gradient-based mesh rendering with differentiable rasterizers~\cite{kato2018neural,liu2019soft,liu2020general} provides another representative optimization paradigm for estimating mesh appearance.

These methods provide important foundations for mesh-based appearance modeling, but differ from CF-Splat in their estimation structure. Rather than independently fitting local surface appearance or relying on gradient-based optimization, CF-Splat derives a linear shading model directly from triangle rasterization and jointly estimates per-vertex appearance through a global LLS formulation.

\section{Closed-Form Scene Appearance Estimation}
\label{sec:closed_form_formulation}

Our approach enables closed-form scene appearance estimation by properly formulating view-dependent appearance over the given mesh and calibrated cameras. In this section, we derive the resulting global LLS formulation for multi-view appearance estimation by establishing the linear relationship between rendered pixel colors and per-vertex SH coefficients.

% We first define the optimization problem throughout CF-Splat. We then establish a linear pixel shading model on per-vertex SH coefficients and assemble this pixel-level linearity into a global multi-view
% observation model. We further introduce adaptive SH regularization based on
% the directional support available at each vertex, yielding a global LLS problem with a unique global minimizer. Finally,
% we describe gradient-guided triangle densification as a preprocessing step to improve the spatial resolution of triangles while preserving the closed-form formulation.

\subsection{Problem Statement}
\label{sec:problem_setup}

We begin by defining the triangle representation and calibrated observations, followed by the global SH appearance state and appearance estimation objective in CF-Splat. Throughout this section, we suppose that the mesh geometry and camera calibrations are already available.

Specifically, let
\begin{equation}
    \mathbb{M}\triangleq(\mathbb{V},\mathbb{F})
\end{equation}
denote the known mesh with $N_v$ vertices $\mathbb{V}\triangleq\{\mathbf{v}_i=(x_i,y_i,z_i)^\top\}_{i=1}^{N_v}$ and $N_f$ triangles (triangular faces) $\mathbb{F}\triangleq\{\mathbf{f}_j=(\iota(j,0),\iota(j,1),\iota(j,2))\}_{j=1}^{N_f}$. Here, $\iota(j,s)\in\{1,\ldots,N_v\}$, $s\in\{0,1,2\}$ is the global index of the $s$-th vertex of face $\mathbf{f}_j$.

We are additionally given $K$ calibrated RGB observations
\begin{equation}
    \mathbb{O}
    =
    \left\{
        \mathbf{I}_k,
        \mathbf{K}_k,
        \mathbf{T}_k
    \right\}_{k=1}^{K},
    \ \
    \mathbf{T}_k
    =
    [\mathbf{R}_k \mid \mathbf{t}_k],
\end{equation}
where
$\mathbf{I}_k\in\mathbb{R}^{H\times W\times3}$ is the $k$-th RGB image, $\mathbf{K}_k\in\mathbb{R}^{3\times3}$ is the intrinsic matrix, and $\mathbf{T}_k$ is the extrinsic transformation from the mesh coordinate frame to the $k$-th camera frame.

To model view-dependent appearance, each vertex in $\mathbb{V}$ is encoded by SH signals~\cite{seeley1966spherical,muller2006spherical}.
Given a maximum SH degree $L$, the per-channel color is
represented as
\begin{equation}
    f(\mathbf{c},\mathbf{d})
    =
    \sum_{l=0}^{L}
    \sum_{m=-l}^{l}
    c_l^m
    y_l^m(\mathbf{d}),
    \label{eq:sh}
\end{equation}
where
$\mathbf{c}\in\mathbb{R}^{(L+1)^2}$
is the SH coefficient vector,
$c_l^m$
is the coefficient associated with degree $l$ and order $m$,
$\mathbf{d}\in\mathbb{S}^{2}$
is the unit viewing direction, and
$y_l^m(\mathbf{d})$
denotes the corresponding SH basis function evaluated at $\mathbf{d}$. For a specific vertex $\mathbf{v}_i \in \mathbb{V}$ viewed from camera $k$, the viewing direction is
\begin{equation}
    \mathbf{d}_k^i
    =
    \frac{
        \mathbf{v}_i-\mathbf{o}_k
    }{
        \|\mathbf{v}_i-\mathbf{o}_k\|_2
    },
    \qquad
    \mathbf{o}_k
    =
    -\mathbf{R}_k^\top\mathbf{t}_k,
    \label{eq:view_direction}
\end{equation}
where
$\mathbf{o}_k$
is the camera center in the mesh coordinate frame. 

Let $q=(L+1)^2$ denote the dimension of $\mathbf{c}$ in \eqref{eq:sh}. The SH basis vector evaluated along $\mathbf{d}_k^i$ is collected as
\begin{equation}
    \mathbf{Y}(\mathbf{d}_k^i)
    \triangleq
    \begin{bmatrix}
        y_0^0(\mathbf{d}_k^i)
        &
        y_1^{-1}(\mathbf{d}_k^i)
        &
        \cdots
        &
        y_L^L(\mathbf{d}_k^i)
    \end{bmatrix}^\top
    \in\mathbb{R}^{q}.
\end{equation}
For each color channel
$\tau\in\{R,G,B\}$,
the SH coefficients of vertex $\mathbf{v}_i$ are collected as
\begin{equation}
    \mathbf{c}_i^{(\tau)}
    \triangleq
    \begin{bmatrix}
        c_{0,i}^{0(\tau)}
        &
        c_{1,i}^{-1(\tau)}
        &
        \cdots
        &
        c_{L,i}^{L(\tau)}
    \end{bmatrix}^\top
    \in\mathbb{R}^{q}.
\end{equation}
The RGB SH coefficient vectors of $\mathbf{v}_i$ are then assembled into
\begin{equation}
    \mathbf{C}_i
    \triangleq
    \begin{bmatrix}
        {\mathbf{c}_i^{(R)}}
        &
        {\mathbf{c}_i^{(G)}}
        &
        {\mathbf{c}_i^{(B)}}
    \end{bmatrix}
    \in
    \mathbb{R}^{q\times 3},
    \label{eq:vertex_sh_state}
\end{equation}
such that the view-dependent RGB color of $\mathbf{v}_i$ from the $k$-th camera is
\begin{equation}
    \mathbf{g}_i^k
    =
    \mathbf{C}_i^\top
    \mathbf{Y}(\mathbf{d}_k^i)
    \in
    \mathbb{R}^{3}.
    \label{eq:vertex_sh_color}
\end{equation}
Since $\mathbb{M}$ and $\mathbb{O}$ are known, $\mathbf{d}_k^i$ and hence $\mathbf{Y}(\mathbf{d}_k^i)$ in \eqref{eq:vertex_sh_color} is fully determined and independent of $\mathbf{C}_i$. Therefore, $\mathbf{g}_i^k$ depends linearly on $\mathbf{C}_i$.

We further stack $\mathbf{C}_i$ of all vertices and define the global SH appearance state as
\begin{equation}
    \mathbf{X}
    \triangleq
    \begin{bmatrix}
        \mathbf{C}_1^\top
        &
        \cdots
        &
        \mathbf{C}_{N_v}^\top
    \end{bmatrix}^{\!\top}
    \in
    \mathbb{R}^{qN_v\times3}.
    \label{eq:global_appearance_state}
\end{equation}
Thus, the photometric objective for scene appearance estimation is expressed as
\begin{equation}
    \mathcal{E}_{\mathrm{photo}}(\mathbf{X})
    \triangleq
    \frac{1}{2}
    \sum_{k=1}^{K}
    \left\|
        \widehat{\mathbf{I}}_k(\mathbf{X}\mid\mathbb{M},\mathbb{O})
        -
        \mathbf{I}_k
    \right\|_F^2.
    \label{eq:appearance_objective}
\end{equation}
where
$\mathcal{E}_{\mathrm{photo}}(\mathbf{X})$
denotes the total photometric residual over $K$ calibrated cameras,
and $\widehat{\mathbf{I}}_k(\mathbf{X}\mid\mathbb{M},\mathbb{O})$
denotes the image rendered from $\mathbb{M}$ and $\mathbb{O}$ under the current
appearance state $\mathbf{X}$.

In \eqref{eq:appearance_objective}, $\mathbf{X}$ is the only unknown to be estimated, avoiding the joint nonlinear optimization over multiple primitive parameters required by 3DGS~\cite{kerbl20233d}. In Section~\ref{sec:linear_triangle}, we show that the rendered pixel colors in $\widehat{\mathbf{I}}_k(\mathbf{X}\mid\mathbb{M},\mathbb{O})$ depend linearly on $\mathbf{C}_i$, and hence on $\mathbf{X}$. Building on this property, Section~\ref{sec:global_observation} extends the linear pixel shading model to a global multi-view observation model, enabling \eqref{eq:appearance_objective} to be reformulated as a closed-form problem that can be solved by a global LLS formulation derived in Section~\ref{sec:global_lls}.

\subsection{Linear Pixel Shading Model}
\label{sec:linear_triangle}

This section establishes the linear dependence of the rendered image
$\widehat{\mathbf{I}}_k(\mathbf{X}\mid\mathbb{M},\mathbb{O})$
on $\mathbf{X}$.

For the $k$-th calibrated image $\mathbf{I}_k$, we arrange its $HW$ pixels
in raster-scan pixel ordering and denote the $r$-th pixel by
$
    \mathbf{p}_{k,r}
    \triangleq
    [u_{k,r}, v_{k,r}]^{\top}
$, where $r\in\{1,\ldots,HW\}$.
We first perform standard z-buffer triangle rasterization to establish the
correspondence between these image pixels and the mesh triangles.
The projection of vertex $\mathbf{v}_i$ onto the $k$-th image is
\begin{equation}
    \mathbf{v}_i^{k}
    =
    \pi
    \left(
        \mathbf{K}_k
        (
        \mathbf{R}_k
        \mathbf{v}_i
        +
        \mathbf{t}_k
        )
    \right),
    \label{eq:vertex_projection}
\end{equation}
where $\pi(\cdot)$ denotes perspective division.

Let $\mathbf{f}_{\rho_k(\mathbf{p}_{k,r})}$ denote the triangle covering
$\mathbf{p}_{k,r}$, where $\rho_k(\cdot)$ returns its global triangle index.
For compact notation, define
\[
    i_{k,r}^{s}
    \triangleq
    \iota\!\left(\rho_k(\mathbf{p}_{k,r}),s\right),
    \ \
    s\in\{0,1,2\},
\]
as the global index of the $s$-th vertex of the covering triangle.
The pixel color is interpolated from the three corresponding vertex colors
$\{\mathbf{g}_{i_{k,r}^{s}}^{k}\}_{s=0}^{2}$ using barycentric weights
$\{\lambda_{k,r}^{s}\}_{s=0}^{2}$.

Since the projection in \eqref{eq:vertex_projection} is fully determined
by $\mathbb{M}$ and $\mathbb{O}$, the barycentric weights are independent
of $\mathbf{X}$. The rendered color at
$\mathbf{p}_{k,r}$ is therefore
\begin{equation}
    \widehat{\mathbf{I}}_k(\mathbf{p}_{k,r})
    =
    \sum_{s=0}^{2}
    \lambda_{k,r}^{s}
    \mathbf{g}_{i_{k,r}^{s}}^{k}.
    \label{eq:pixel_barycentric_color}
\end{equation}

We further define the local observation feature
\begin{equation}
    \boldsymbol{\phi}_{k,r}^{s}
    \triangleq
    \lambda_{k,r}^{s}
    \mathbf{Y}\!\left(\mathbf{d}_k^{i_{k,r}^{s}}\right)
    \in\mathbb{R}^{q}.
    \label{eq:local_observation_feature}
\end{equation}
Substituting the vertex SH color model in
\eqref{eq:vertex_sh_color} into \eqref{eq:pixel_barycentric_color}
gives
\begin{equation}
\begin{aligned}
    \widehat{\mathbf{I}}_k(\mathbf{p}_{k,r})
    &=
    \sum_{s=0}^{2}
    \lambda_{k,r}^{s}
    \mathbf{C}_{i_{k,r}^{s}}^{\top}
    \mathbf{Y}\!\left(\mathbf{d}_k^{i_{k,r}^{s}}\right)
    \\
    &=
    \sum_{s=0}^{2}
    \mathbf{C}_{i_{k,r}^{s}}^{\top}
    \boldsymbol{\phi}_{k,r}^{s}.
\end{aligned}
\label{eq:linear_pixel_model}
\end{equation}
Here, $\boldsymbol{\phi}_{k,r}^{s}$ is constant (known), and
$\mathbf{C}_{i_{k,r}^{s}} \in \mathbf{X}$.

\eqref{eq:linear_pixel_model} shows that each rendered pixel $\widehat{\mathbf{I}}_k(\mathbf{p}_{k,r})$ is a
linear combination of only the three SH coefficient blocks $\{\mathbf{C}_{i_{k,r}^{s}}\}_{s=0}^2$ associated with
its covering triangle. Therefore, each rendered pixel depends linearly on
$\mathbf{X}$, and this linear dependence extends to the entire rendered
image $\widehat{\mathbf{I}}_k(\mathbf{X}\mid\mathbb{M},\mathbb{O})$.
We next assemble these pixel-level relations into a global multi-view
observation model in Section~\ref{sec:global_observation}.

\subsection{Global Multi-View Observation Model}
\label{sec:global_observation}

We now extend the pixel-level linear color in
\eqref{eq:linear_pixel_model} first to each calibrated view and then to all
$K$ views in $\mathbb{O}$. 

For the $k$-th view, we construct
the per-view observation matrix
$\mathbf{A}_k\in\mathbb{R}^{HW\times qN_v}$.
For the row corresponding to $\mathbf{p}_{k,r}$, only the three
$q$-dimensional vertex blocks associated with its covering triangle are
nonzero
\begin{equation}
    [\mathbf{A}_k]_{r,i_{k,r}^{s}}
    =
    \left(\boldsymbol{\phi}_{k,r}^{s}\right)^{\top}.
    \label{eq:per_view_observation_blocks}
\end{equation}
where 
$[\mathbf{A}_k]_{r,i_{k,r}^{s}}\in\mathbb{R}^{1\times q}$
denotes the vertex block indexed by $i_{k,r}^{s}$ in the $r$-th row.

Using the raster-scan pixel ordering, the observed RGB colors in the
$k$-th view are stacked as
\begin{equation}
\begin{aligned}
    \mathbf{B}_k
    &\triangleq
    \begin{bmatrix}
        \mathbf{I}_k(\mathbf{p}_{k,1})
        &
        \cdots
        &
        \mathbf{I}_k(\mathbf{p}_{k,HW})
    \end{bmatrix}^{\!\top},
\end{aligned}
\label{eq:per_view_rgb_matrices}
\end{equation}
where
$\mathbf{B}_k
\in\mathbb{R}^{HW\times3}$.
The pixel-level shading in \eqref{eq:linear_pixel_model} therefore forms the per-view linear model
$\widehat{\mathbf{B}}_k
    =
    \mathbf{A}_k\mathbf{X}
$,
which are then assembled into the global matrices
\begin{equation}
\begin{aligned}
    \mathbf{A}
    &\triangleq
    \begin{bmatrix}
        \mathbf{A}_1^{\top}
        &
        \cdots
        &
        \mathbf{A}_K^{\top}
    \end{bmatrix}^{\!\top}
    \in
    \mathbb{R}^{KHW\times qN_v},
    \\
    \mathbf{B}
    &\triangleq
    \begin{bmatrix}
        \mathbf{B}_1^{\top}
        &
        \cdots
        &
        \mathbf{B}_K^{\top}
    \end{bmatrix}^{\!\top}
    \in
    \mathbb{R}^{KHW\times3}.
\end{aligned}
\label{eq:global_matrix_assembly}
\end{equation}
Accordingly, the global multi-view observation model is
\begin{equation}
    \widehat{\mathbf{B}}
    =
    \mathbf{A}\mathbf{X}.
    \label{eq:global_linear_observation}
\end{equation}
Since each pixel depends on only the three vertices of its covering triangle,
every row of $\mathbf{A}$ contains at most three nonzero $q$-dimensional
blocks. The global observation matrix is therefore highly sparse.

To reduce the influence of grazing-angle observations, we assign each pixel
an angle-dependent confidence according to the angle between the normal of
its covering triangle and the corresponding camera-to-surface direction
\begin{equation}
    w_{k,r}
    \triangleq
    \max\!\left(
        \epsilon_w,
        \left|
        \mathbf{n}_{\rho_k(\mathbf{p}_{k,r})}^{\top}
        \frac{
            \displaystyle
            \sum_{s=0}^{2}
            \lambda_{k,r}^{s}\mathbf{v}_{i_{k,r}^{s}}
            -
            \mathbf{o}_k
        }{
            \displaystyle
            \left\|
            \sum_{s=0}^{2}
            \lambda_{k,r}^{s}\mathbf{v}_{i_{k,r}^{s}}
            -
            \mathbf{o}_k
            \right\|_2
        }
        \right|
    \right).
    \label{eq:observation_weight}
\end{equation}
Here,
$\mathbf{n}_{\rho_k(\mathbf{p}_{k,r})}$
is the unit normal of the covering triangle,
$0<\epsilon_w<1$
is the minimizer confidence, and
$0<w_{k,r}\leq1$.
\eqref{eq:observation_weight} assigns higher confidence to near-frontal observations
and lower confidence to grazing-angle observations.

Following the pixel ordering in \eqref{eq:per_view_rgb_matrices}, the
confidence matrix for the $k$-th view is
\begin{equation}
    \mathbf{W}_k
    \triangleq
    \operatorname{diag}\!\left(
        \begin{bmatrix}
            w_{k,1}
            &
            \cdots
            &
            w_{k,HW}
        \end{bmatrix}^{\!\top}
    \right)
    \in
    \mathbb{R}^{HW\times HW}.
    \label{eq:per_view_confidence_matrix}
\end{equation}
The global confidence matrix is assembled from all views as
\begin{equation}
    \mathbf{W}
    \triangleq
    \operatorname{blkdiag}
    \left(
        \mathbf{W}_1,
        \ldots,
        \mathbf{W}_K
    \right)
    \in
    \mathbb{R}^{KHW\times KHW}.
    \label{eq:global_confidence_matrix}
\end{equation}

The matrices $\mathbf{A}$, $\mathbf{B}$, and $\mathbf{W}$ therefore define
the weighted global residual
$\mathbf{W}^{1/2}(\mathbf{A}\mathbf{X}-\mathbf{B})$.
We next discuss the directional constraints on the per-vertex SH
coefficients and introduce an adaptive regularization in Section~\ref{sec:adaptive_sh}.

\subsection{Adaptive SH Regularization}
\label{sec:adaptive_sh}

Although the global observation model in \eqref{eq:global_linear_observation} combines all pixels from all views, multiple pixels from the same camera do not provide multiple independent SH directions for a given vertex. Consequently, high-order SH coefficients can remain weakly constrained when a vertex is observed from only a small number of cameras. As a result, $\mathbf{A}$ may become rank-deficient under uneven multi-view observations.

Suppose
$\mathbf{A}_k^{(i)}\in\mathbb{R}^{HW\times q}$
denote the $q$-column block of $\mathbf{A}_k$ associated with
$\mathbf{v}_i$. Its corresponding block in the global observation matrix is
\begin{equation}
    \mathbf{A}^{(i)}
    \triangleq
    \begin{bmatrix}
        {\mathbf{A}_1^{(i)}}^{\top}
        &
        \cdots
        &
        {\mathbf{A}_K^{(i)}}^{\top}
    \end{bmatrix}^{\!\top}
    \in\mathbb{R}^{KHW\times q}.
    \label{eq:vertex_observation_block}
\end{equation}

Let $n_i^{\mathrm{view}}$ denote the number of cameras for which
$\mathbf{A}_k^{(i)}$ contains at least one nonzero row.

\noindent\textbf{Proposition 1 (Per-vertex directional-rank bound).}
The necessary condition for uniquely estimating the SH coefficients at vertex $\mathbf{v}_i$ is that its observation rank satisfies
\begin{equation}
    \operatorname{rank}\!\left(\mathbf{A}^{(i)}\right)
    \leq
    n_i^{\mathrm{view}}.
    \label{eq:vertex_directional_rank}
\end{equation}

\noindent\textit{Proof.}
For a camera $k$ observing $\mathbf{v}_i$, every nonzero row of
$\mathbf{A}_k^{(i)}$ has the form
$
    \lambda_k^s
    \mathbf{Y}\!\left(\mathbf{d}_k^i\right)^{\top},
$
where the barycentric weight $\lambda_k^s$ varies across pixels, while the SH
basis vector remains unchanged. Hence, all such rows are scalar multiples of
the same vector and
$\operatorname{rank}(\mathbf{A}_k^{(i)})\leq1$.
Since only $n_i^{\mathrm{view}}$ per-view blocks contain nonzero rows
\begin{equation}
    \operatorname{rank}\!\left(\mathbf{A}^{(i)}\right)
    \leq
    \sum_{k=1}^{K}
    \operatorname{rank}\!\left(\mathbf{A}_k^{(i)}\right)
    \leq
    n_i^{\mathrm{view}},
\end{equation}
which proves \eqref{eq:vertex_directional_rank}.
\hfill$\square$

An SH representation through degree $l$ contains $(l+1)^2$ coefficients
per color channel. Therefore, a necessary dimensional condition for
independently constraining these coefficients is
\begin{equation}
    (l+1)^2
    \leq
    n_i^{\mathrm{view}}.
    \label{eq:sh_dimensional_condition}
\end{equation}
This condition does not require the viewing directions to be well
distributed and is therefore used as a soft regularization criterion rather
than as a hard SH truncation rule.

Accordingly, the band-wise regularizer for vertex $\mathbf{v}_i$ is defined
as
\begin{equation}
\begin{aligned}
    \boldsymbol{\Lambda}_i
    &\triangleq
    \operatorname{diag}
    \left(
        \gamma_{i,0}\mathbf{I}_{1},
        \gamma_{i,1}\mathbf{I}_{3},
        \ldots,
        \gamma_{i,L}\mathbf{I}_{2L+1}
    \right)
    \in\mathbb{R}^{q\times q},
    \\
    \gamma_{i,l}
    &=
    \begin{cases}
        \gamma_s,
        & (l+1)^2\leq n_i^{\mathrm{view}}
        \\[2pt]
        \gamma_u,
        & (l+1)^2>n_i^{\mathrm{view}},
    \end{cases}
\end{aligned}
\label{eq:adaptive_sh_regularizer}
\end{equation}
where $0<\gamma_s\ll\gamma_u$.
Thus, SH bands with sufficient dimensional support receive weak
regularization, whereas bands lacking sufficient directional support are strongly constrained.

Following the vertex ordering in $\mathbf{X}$ in \eqref{eq:global_appearance_state}, the global SH regularizer is
assembled as
\begin{equation}
    \boldsymbol{\Lambda}
    \triangleq
    \operatorname{blkdiag}
    \left(
        \boldsymbol{\Lambda}_1,
        \ldots,
        \boldsymbol{\Lambda}_{N_v}
    \right)
    \in\mathbb{R}^{qN_v\times qN_v}.
    \label{eq:global_sh_regularizer}
\end{equation}
The adaptive SH regularization retains the same $q$-dimensional appearance
state for each vertex while suppressing high-order components that are
weakly constrained by the available views. It consequently reduces the
sensitivity of the global estimation to uneven directional coverage and
improves its numerical stability for a unique global minimizer.

\subsection{Global LLS Formulation}
\label{sec:global_lls}

Combining the global multi-view observation model in \eqref{eq:global_linear_observation} with the adaptive SH regularization in \eqref{eq:global_sh_regularizer} yields the final global LLS formulation
\begin{equation}
    \mathbf{X}^{\star}
    =
    \arg\min_{\mathbf{X}}
    \frac{1}{2}
    \left\|
        \mathbf{W}^{1/2}
        \left(
            \mathbf{A}\mathbf{X}
            -
            \mathbf{B}
        \right)
    \right\|_F^2
    +
    \frac{1}{2}
    \left\|
        \boldsymbol{\Lambda}^{1/2}
        \mathbf{X}
    \right\|_F^2.
    \label{eq:global_lls}
\end{equation}
The first term jointly fits the RGB observations from all calibrated views,
with grazing-angle observations down-weighted by $\mathbf{W}$. The second
term adaptively constrains the SH bands according to the directional support
available at each vertex.

Setting the gradient of \eqref{eq:global_lls} with respect to $\mathbf{X}$
to zero gives the global normal system
\begin{equation}
    \mathbf{H}\mathbf{X}^{\star}
    =
    \mathbf{G},
    \ \
    \mathbf{H}
    =
    \mathbf{A}^{\top}\mathbf{W}\mathbf{A}
    +
    \boldsymbol{\Lambda},
    \ \
    \mathbf{G}
    =
    \mathbf{A}^{\top}\mathbf{W}\mathbf{B}.
    \label{eq:normal_system}
\end{equation}
Here,
$\mathbf{H}\in\mathbb{R}^{qN_v\times qN_v}$
is the shared normal matrix for the three color channels, and
$\mathbf{G}\in\mathbb{R}^{qN_v\times3}$
is the corresponding right-hand-side matrix, whose three columns contain the weighted observation terms for the R, G, and B channels, respectively.

We now establish below that the proposed adaptive SH regularization in \eqref{eq:global_sh_regularizer} makes the resulting normal system strictly
positive definite and guarantees a unique global minimizer.

\noindent\textbf{Proposition 2 (Unique global minimizer).}
The global LLS problem in \eqref{eq:global_lls} admits a unique global
minimizer.

\noindent\textit{Proof.}
By construction,
$w_{k,r}>0$
for every pixel, and hence
$\mathbf{W}\succ0$.
Moreover, the adaptive SH regularization assigns either
$\gamma_s$ or $\gamma_u$ to every SH coefficient, where
$0<\gamma_s\ll\gamma_u$.
Therefore,
$\boldsymbol{\Lambda}\succ0$.

For any nonzero vector
$\mathbf{z}\in\mathbb{R}^{qN_v}$,
we have
\begin{equation}
\begin{aligned}
    \mathbf{z}^{\top}\mathbf{H}\mathbf{z}
    &=
    \mathbf{z}^{\top}
    \mathbf{A}^{\top}\mathbf{W}\mathbf{A}
    \mathbf{z}
    +
    \mathbf{z}^{\top}
    \boldsymbol{\Lambda}
    \mathbf{z}
    \\
    &=
    \left\|
        \mathbf{W}^{1/2}
        \mathbf{A}\mathbf{z}
    \right\|_2^2
    +
    \left\|
        \boldsymbol{\Lambda}^{1/2}
        \mathbf{z}
    \right\|_2^2>0.
\end{aligned}
\label{eq:normal_matrix_spd}
\end{equation}
The strict inequality follows from the SH regularization term, which is
positive for every nonzero $\mathbf{z}$. Thus,
$\mathbf{H}\succ0$.
Since the three RGB channels share the same matrix $\mathbf{H}$ and are
separable in \eqref{eq:global_lls}, each channel has a unique minimizer.
Therefore, \eqref{eq:global_lls} admits a unique global minimizer
$\mathbf{X}^{\star}$.
\hfill$\square$

The resulting estimator has the analytical form
\begin{equation}
    \mathbf{X}^{\star}
    =
    \mathbf{H}^{-1}\mathbf{G}
    =
    \left(
        \mathbf{A}^{\top}
        \mathbf{W}
        \mathbf{A}
        +
        \boldsymbol{\Lambda}
    \right)^{-1}
    \mathbf{A}^{\top}
    \mathbf{W}
    \mathbf{B},
    \label{eq:closed_form_solution}
\end{equation}
which characterizes the global closed-form
solution of the appearance estimation problem in \eqref{eq:appearance_objective}.

Algorithm~\ref{alg:closed_form_appearance} summarizes the complete closed-form
scene appearance estimation pipeline. For each calibrated view, triangle
rasterization establishes the pixel-to-triangle correspondence, from which the
local SH observation features, confidence weights, and per-view matrices
$\mathbf{A}_k$, $\mathbf{B}_k$, and $\mathbf{W}_k$ are constructed. These
per-view components are then assembled into the global matrices
$\mathbf{A}$, $\mathbf{B}$, and $\mathbf{W}$. Based on the available view
support at each vertex, the adaptive SH regularizer
$\boldsymbol{\Lambda}$ is constructed, yielding the global normal system
defined by $\mathbf{H}$ and $\mathbf{G}$. In practice, the
inverse of $\mathbf{H}$ is not explicitly formed; instead, the resulting
system is solved using the triangle-structured matrix-free PCG solver proposed in
Section~\ref{sec:scalable_solver} to obtain the final appearance estimate
$\mathbf{X}^{*}$.

\begin{algorithm}[t]
\caption{\small Closed-Form Scene Appearance Estimation}
\label{alg:closed_form_appearance}
\small
\begin{algorithmic}[1]
\Require Mesh $\mathbb{M}$, calibrated RGB observations $\mathbb{O}$,
         SH degree $L$, supported/under-supported SH regularization
         $\gamma_s,\gamma_u$, minimum confidence weight $\epsilon_w$.
\Ensure Estimated appearance $\mathbf{X}^{*}$

\State Initialize rasterization records $\mathcal{R}$ and
       $\{n_i^{\mathrm{view}}\}_{i=1}^{N_v}$

\For{$k=1,\ldots,K$}
    \State Rasterize $\mathbb{M}$ in view $k$ and initialize $\mathbf{A}_k$

    \For{each valid pixel $\mathbf{p}_{k,r}$}
        \State Obtain $\{i_{k,r}^{s},\lambda_{k,r}^{s}\}_{s=0}^{2}$

        \State
        $\boldsymbol{\phi}_{k,r}^{s}
        \gets
        \lambda_{k,r}^{s}
        \mathbf{Y}(\mathbf{d}_k^{i_{k,r}^{s}})$,
        $s=0,1,2$
        % \Comment{\eqref{eq:local_observation_feature}}

        \State
        $[\mathbf{A}_k]_{r,i_{k,r}^{s}}
        \gets
        (\boldsymbol{\phi}_{k,r}^{s})^\top$
        % \Comment{\eqref{eq:per_view_observation_blocks}}

        \State Compute $w_{k,r}$ using
        \eqref{eq:observation_weight}

        \State Store
        $\{i_{k,r}^{s},\boldsymbol{\phi}_{k,r}^{s}\}_{s=0}^{2},w_{k,r}$
        in $\mathcal{R}$
    \EndFor

    \State
    $\mathbf{B}_k
    \gets
    [\mathbf{I}_k(\mathbf{p}_{k,1})\ \cdots\
    \mathbf{I}_k(\mathbf{p}_{k,HW})]^\top$

    \State
    $\mathbf{W}_k
    \gets
    \operatorname{diag}(w_{k,1},\ldots,w_{k,HW})$

    \State Update $\{n_i^{\mathrm{view}}\}$
\EndFor

\State
$\mathbf{A}\gets
[\mathbf{A}_1^\top\ \cdots\ \mathbf{A}_K^\top]^\top$,
$\quad
\mathbf{B}\gets
[\mathbf{B}_1^\top\ \cdots\ \mathbf{B}_K^\top]^\top$
% \Comment{\eqref{eq:global_matrix_assembly}}

\State
$\mathbf{W}\gets
\operatorname{blkdiag}(\mathbf{W}_1,\ldots,\mathbf{W}_K)$
% \Comment{\eqref{eq:global_confidence_matrix}}

\For{$i=1,\ldots,N_v$}
    \State Construct $\boldsymbol{\Lambda}_i$ from
    $n_i^{\mathrm{view}}$, $\gamma_s$, and $\gamma_u$
    using \eqref{eq:adaptive_sh_regularizer}
\EndFor

\State
$\boldsymbol{\Lambda}
\gets
\operatorname{blkdiag}
(\boldsymbol{\Lambda}_1,\ldots,\boldsymbol{\Lambda}_{N_v})$
% \Comment{\eqref{eq:global_sh_regularizer}}

\State
$\mathbf{H}\gets
\mathbf{A}^{\top}\mathbf{W}\mathbf{A}
+\boldsymbol{\Lambda}$,
$\quad
\mathbf{G}\gets
\mathbf{A}^{\top}\mathbf{W}\mathbf{B}$
% \Comment{\eqref{eq:normal_system}}

\State
$\mathbf{X}^{*}
\gets
\mathbf{H}^{-1}\mathbf{G}$

\State \Return $\mathbf{X}^{*}$

\end{algorithmic}
\end{algorithm}

\subsection{Gradient-Guided Triangle Densification}
\label{sec:densification}

Although the mesh provides sufficient geometric fidelity, coarse triangles
may still limit the spatial resolution of the vertex appearance
representation. We therefore apply a
lightweight gradient-guided triangle densification as a preprocessing step before
constructing the global LLS system.

Using the pixel-to-triangle correspondence established in
Section~\ref{sec:linear_triangle}, let
$\Omega_{k,j}\triangleq
\{\mathbf{p}_{k,r}\mid \rho_k(\mathbf{p}_{k,r})=j\}$
denote the pixels covered by triangle $\mathbf{f}_j$ in view $\mathbb{O}_k$.
We measure its multi-view image-space variation as
\begin{equation}
    S_j
    \triangleq
    \sum_{k=1}^{K}
    \sum_{\mathbf{p}\in\Omega_{k,j}}
    \left\|
        \nabla\mathbf{I}_k(\mathbf{p})
    \right\|_2.
    \label{eq:face_gradient_score}
\end{equation}
Triangles with large $S_j$ are selected for
subdivision by concentrating additional vertices.

For each selected triangle, the barycentric weights
introduced in Section~\ref{sec:linear_triangle} are used to map its covered
pixels back to the triangle surface. A new vertex is placed at the mean of
these surface samples, and the parent triangle is replaced by three child
triangles sharing the new vertex. 

The refined mesh is re-rasterized after each densification round. After a number of rounds, densification is terminated and the resulting mesh
is used to construct the linear pixel model in
\eqref{eq:linear_pixel_model} and the global LLS system in
\eqref{eq:global_lls}. Importantly, densification is completed before
appearance estimation and thus does not introduce extra iterations when solving the global LLS problem.

\section{Triangle-Structured Matrix-Free PCG Solver}
\label{sec:scalable_solver}

The closed-form formulation in
Section~\ref{sec:closed_form_formulation} defines the global appearance
estimator through the LLS problem in \eqref{eq:global_lls}.
For large-scale scenes, explicitly constructing the global normal matrix
$\mathbf{H}$ and solving the linear system in \eqref{eq:closed_form_solution} lead to substantial computation and memory costs.
We therefore implement the same the linear system using a
triangle-structured matrix-free PCG solver. By exploiting the
three-vertex support of each pixel, the normal operator is evaluated
directly from rasterization records without explicitly constructing
$\mathbf{H}$, while vertex-block preconditioning supports efficient
numerical solution.

\subsection{Triangle-Local Normal Operator}
\label{sec:matrix_free}

PCG repeatedly requires the product between the normal matrix and a
vector.  In CF-Splat, each valid pixel is covered by
one triangle and therefore depends only on its three vertices.
We use this property to evaluate the required matrix-vector
product directly from the rasterization records, without explicitly constructing $\mathbf{H}$.

We first construct the right-hand side $\mathbf{G}$ in
\eqref{eq:normal_system}. Let
$\mathbf{G}_i\in\mathbb{R}^{q\times3}$ denote the block
corresponding to vertex $\mathbf{v}_i$. Starting from
$\mathbf{G}_i=\mathbf{0}$, each valid pixel contributes to the three
vertices of its covering triangle as
\begin{equation}
    \mathbf{G}_{i_{k,r}^{s}}
    \gets
    \mathbf{G}_{i_{k,r}^{s}}
    +
    w_{k,r}
    \boldsymbol{\phi}_{k,r}^{s}
    \mathbf{I}_k(\mathbf{p}_{k,r})^\top.
    \label{eq:triangle_rhs}
\end{equation}
Accumulating \eqref{eq:triangle_rhs} over all valid pixels and views
gives the right-hand side $\mathbf{G}$ of the global normal system.

For the matrix-vector product, consider an observation vector
$\mathbf{u}\in\mathbb{R}^{qN_v}$ with vertex blocks
$\mathbf{u}_i\in\mathbb{R}^{q}$.
We first initialize each output block using the regularization term
\begin{equation}
    \mathbf{h}_i
    \gets
    \boldsymbol{\Lambda}_i\mathbf{u}_i.
    \label{eq:matrix_free_regularization}
\end{equation}
For a valid pixel $\mathbf{p}_{k,r}$, rasterization provides its three
vertex indices $\{i_{k,r}^{s}\}_{s=0}^{2}$ and the corresponding
observation vectors
$\{\boldsymbol{\phi}_{k,r}^{s}\}_{s=0}^{2}$.
The contribution of this pixel is accumulated as
\begin{equation}
    \mathbf{h}_{i_{k,r}^{s}}
    \gets
    \mathbf{h}_{i_{k,r}^{s}}
    +
    w_{k,r}\boldsymbol{\phi}_{k,r}^{s}
    \left(
        \sum_{t=0}^{2}
        \left(\boldsymbol{\phi}_{k,r}^{t}\right)^\top
        \mathbf{u}_{i_{k,r}^{t}}
    \right).
    \label{eq:matrix_free_operator}
\end{equation}
For each pixel, only the three vertex blocks of its covering triangle
are involved. Their contributions are first combined for the current
pixel and then accumulated back to the same three vertices.
Thus, repeating this operation over all valid pixels gives
\begin{equation}
    \mathcal{H}(\mathbf{u})
    =
    \underbrace{
    \begin{bmatrix}
        \mathbf{h}_1^\top &
        \cdots &
        \mathbf{h}_{N_v}^\top
    \end{bmatrix}^{\top}
    }_{\mathbf{H}\mathbf{u}}.
    \label{eq:matrix_free}
\end{equation}
Thus, the normal-matrix product required by PCG is evaluated without
explicitly constructing $\mathbf{H}$ in \eqref{eq:normal_system}.

\subsection{PCG with Vertex-Block Preconditioner}
\label{sec:pcg}

To improve the numerical efficiency of PCG, we use a vertex-block
preconditioner that retains the coupling among the SH coefficients
associated with each vertex.

We use one $q\times q$ preconditioner block for each vertex.
Starting from
$\mathbf{P}_i=\boldsymbol{\Lambda}_i$, each pixel contributes
\begin{equation}
    \mathbf{P}_{i_{k,r}^{s}}
    \gets
    \mathbf{P}_{i_{k,r}^{s}}
    +
    w_{k,r}
    \boldsymbol{\phi}_{k,r}^{s}
    \left(\boldsymbol{\phi}_{k,r}^{s}\right)^\top.
    \label{eq:block_jacobi}
\end{equation}
After accumulation, $\mathbf{P}_i$ is the complete diagonal block
of the normal matrix associated with vertex $\mathbf{v}_i$.
Compared with a scalar Jacobi preconditioner, this preserves the
coupling among all SH coefficients of the same vertex.

Since the adaptive SH regularization makes each
$\mathbf{P}_i$ positive definite, we compute
\begin{equation}
    \mathbf{P}_i
    =
    \mathbf{L}_i\mathbf{L}_i^\top
    \label{eq:block_cholesky}
\end{equation}
once during system setup.
Applying the preconditioner during PCG then only requires solving the
small $q\times q$ system
$\mathbf{P}_i\mathbf{z}_i=\mathbf{r}_i$ independently for each
vertex using the Cholesky factor $\mathbf{L}_i$.

The right-hand side in \eqref{eq:triangle_rhs} and the preconditioner
blocks in \eqref{eq:block_jacobi} depend on the same rasterized pixel
records and are therefore constructed in a shared traversal of the
input observations. The three RGB channels also share the same
triangle indices, observation vectors, normal operator, and
preconditioner. Hence, the Cholesky factors $\{\mathbf{L}_i\}$ are
computed once and reused for all three channel-wise PCG solves.

For channel $\tau\in\{R,G,B\}$, PCG is terminated when the relative
residual satisfies
\begin{equation}
    \sum_{i=1}^{N_v}
    \left\|
        \mathbf{G}_i^{(\tau)}
        -
        \mathcal{H}_i\!\left(\mathbf{X}^{(\tau)}\right)
    \right\|_2^2
    \leq
    \delta^2
    \sum_{i=1}^{N_v}
    \left\|
        \mathbf{G}_i^{(\tau)}
    \right\|_2^2,
    \label{eq:pcg_stopping}
\end{equation}
where $\delta$ is the prescribed relative residual tolerance and
$T_{\max}$ is the maximum iteration count. The tolerance $\delta$
controls the numerical residual of the global LLS solve.

Algorithm~\ref{alg:matrix_free_pcg} summarizes the triangle-structured matrix-free PCG solver of the global LLS system in
Algorithm~\ref{alg:closed_form_appearance}.
\Call{TriangleSetup}{} constructs the right-hand side and vertex-block
preconditioner, while \Call{TriangleNormal}{} evaluates the
normal-matrix product without explicitly constructing $\mathbf{H}$.
The resulting system is then solved channel-wise by PCG.

\begin{algorithm}[t]
\caption{\small Triangle-Structured Matrix-Free PCG}
\label{alg:matrix_free_pcg}
\small

% =========================================================
% Main PCG
% =========================================================
\begin{algorithmic}[1]
\Require Images $\{\mathbf{I}_k\}$, rasterization records $\mathcal{R}$,
         global SH regularizer $\boldsymbol{\Lambda}$, tolerance $\delta$,
         and maximum iteration count $T_{\max}$
\Ensure Estimated appearance $\mathbf{X}^{*}$

\Function{TrianglePCG}{$\{\mathbf{I}_k\},\mathcal{R},
 \boldsymbol{\Lambda},\delta,T_{\max}$}

    \State
    $(\mathbf{G},\{\mathbf{L}_i\})
    \gets
    \Call{TriangleSetup}{}$,
    $\mathbf{X}\gets\mathbf{0}$

    \For{$\tau\in\{R,G,B\}$}
        \State
        $\mathbf{g}\gets\mathbf{G}^{(\tau)}$,
        $\mathbf{r}\gets\mathbf{g}$,
        $\eta\gets\|\mathbf{g}\|_2^2$

        \State
        Solve $\mathbf{P}\mathbf{z}=\mathbf{r}$ using
        $\{\mathbf{L}_i\}$,
        $\mathbf{p}\gets\mathbf{z}$,
        $\sigma\gets\mathbf{r}^{\top}\mathbf{z}$

        \For{$t=1,\ldots,T_{\max}$}
            \State
            $\mathbf{h}
            \gets
            \Call{TriangleNormal}{\mathbf{p}}$,
            $\displaystyle
            \alpha
            \gets
            \frac{\sigma}{\mathbf{p}^{\top}\mathbf{h}}$

            \State
            $\mathbf{X}^{(\tau)}
            \gets
            \mathbf{X}^{(\tau)}+\alpha\mathbf{p}$,
            $\mathbf{r}
            \gets
            \mathbf{r}-\alpha\mathbf{h}$

            \If{$\|\mathbf{r}\|_2^2\leq\delta^2\eta$}
                \State \textbf{break}
            \EndIf

            \State
            Solve $\mathbf{P}\mathbf{z}=\mathbf{r}$
            using $\{\mathbf{L}_i\}$

            \State
            $\sigma_{\mathrm{new}}
            \gets
            \mathbf{r}^{\top}\mathbf{z}$,
            $\displaystyle
            \beta
            \gets
            \frac{\sigma_{\mathrm{new}}}{\sigma}$

            \State
            $\mathbf{p}
            \gets
            \mathbf{z}+\beta\mathbf{p}$,
            $\sigma\gets\sigma_{\mathrm{new}}$
        \EndFor

        \State
        $\mathbf{h}
        \gets
        \Call{TriangleNormal}{\mathbf{X}^{(\tau)}}$
    \EndFor

    \State
    $\mathbf{X}^{*}\gets\mathbf{X}$

    \State
    \Return $\mathbf{X}^{*}$
\EndFunction
\end{algorithmic}

\vspace{1mm}

% =========================================================
% Side-by-side auxiliary functions
% =========================================================
\noindent
\begin{minipage}[t]{0.49\linewidth}
\scriptsize
\begin{algorithmic}[1]

\Function{TriangleSetup}{}
    \State
    $\mathbf{G}\gets\mathbf{0}$,
    $\mathbf{P}_i\gets\boldsymbol{\Lambda}_i,\ \forall i$

    \For{$k=1,\ldots,K$}
        \For{each valid pixel $\mathbf{p}_{k,r}$}
            \For{$s=0,1,2$}
                \State
                Update $\mathbf{G}_{i_{k,r}^{s}}$
                using \eqref{eq:triangle_rhs}

                \State
                Update $\mathbf{P}_{i_{k,r}^{s}}$
                using \eqref{eq:block_jacobi}
            \EndFor
        \EndFor
    \EndFor

    \For{$i=1,\ldots,N_v$}
        \State
        Compute $\mathbf{L}_i$
        using \eqref{eq:block_cholesky}
    \EndFor

    \State
    \Return $\mathbf{G},\{\mathbf{L}_i\}$
\EndFunction

\end{algorithmic}
\end{minipage}
\hfill
\begin{minipage}[t]{0.49\linewidth}
\scriptsize
\begin{algorithmic}[1]

\Function{TriangleNormal}{$\mathbf{u}$}
    \For{$i=1,\ldots,N_v$}
        \State
        Initialize $\mathbf{h}_i$
        using \eqref{eq:matrix_free_regularization}
    \EndFor

    \For{$k=1,\ldots,K$}
        \For{each valid pixel $\mathbf{p}_{k,r}$}
            \For{$s=0,1,2$}
                \State
                Update $\mathbf{h}_{i_{k,r}^{s}}$
                using \eqref{eq:matrix_free_operator}
            \EndFor
        \EndFor
    \EndFor

    \State
    Assemble $\mathcal{H}(\mathbf{u})$
    using \eqref{eq:matrix_free}

    \State
    \Return $\mathcal{H}(\mathbf{u})$
\EndFunction

\end{algorithmic}
\end{minipage}

\end{algorithm}

\subsection{Scalability Analysis}
\label{sec:solver_scalability}

We next analyze the computational and memory scalability of this
matrix-free implementation of the global LLS estimator in \eqref{eq:closed_form_solution}.

\paragraph{Computational complexity}
The computation time mainly depends on the number of input observations,
the number of mesh vertices, and the number of PCG iterations.
During \Call{TriangleSetup}{}, all $K$ views are traversed once to
construct the right-hand side $\mathbf{G}$ using
\eqref{eq:triangle_rhs} and the vertex-block preconditioners
$\{\mathbf{P}_i\}$ using \eqref{eq:block_jacobi}.
These operations require $O(KHWq)$ and $O(KHWq^2)$, respectively.
The Cholesky factorizations in \eqref{eq:block_cholesky} further require
$O(N_vq^3)$ operations.

At each PCG iteration, \Call{TriangleNormal}{} evaluates
\eqref{eq:matrix_free_operator} over all valid pixels, requiring
$O(KHWq)$ operations. Applying the vertex-block preconditioner requires
$O(N_vq^2)$ operations, while the remaining vector updates require
$O(N_vq)$.

Let $T_{\mathrm{pcg}}\leq T_{\max}$ denote the largest iteration count
among the three color channels. The total computation is
\begin{equation}
\begin{aligned}
    T_{\mathrm{solve}}
    &=
    O\!\left(
        KHWq^2 + N_vq^3
    \right)
    \\
    &\quad+
    O\!\left(
        T_{\mathrm{pcg}}
        \left(
            KHWq + N_vq^2
        \right)
    \right)
    \\
    &\quad+
    O\!\left(
        KHWq + N_vq
    \right).
\end{aligned}
\label{eq:solver_time_complexity}
\end{equation}
The last term accounts for the final residual evaluation by
\Call{TriangleNormal}{} according to \eqref{eq:pcg_stopping}.

For a prescribed SH degree $L$, $q$ is constant. The cost of
\Call{TriangleSetup}{}, each \Call{TriangleNormal}{} evaluation, and
the PCG solve therefore grows linearly with the number of image
observations $KHW$. Thus,
\textit{the computation time grows approximately linearly with $K$}.

\paragraph{Memory complexity}
The peak GPU memory mainly depends on the number of mesh vertices and
the image resolution. Since \Call{TriangleNormal}{} evaluates the
normal-matrix product directly from the rasterization records according
to \eqref{eq:matrix_free}, neither the global observation matrix
$\mathbf{A}$ nor the normal matrix $\mathbf{H}$ needs to be explicitly
stored on the GPU.

The SH coefficients, right-hand side, and PCG working vectors require
$O(N_vq)$ GPU memory. With view-wise processing, only the data required
for the current view needs to reside on the GPU, requiring
$O(HWq)$ additional memory. Therefore, the peak GPU memory complexity is
\begin{equation}
    M_{\mathrm{GPU}}
    =
    O\!\left(
        N_vq + HWq
    \right).
    \label{eq:solver_memory_complexity}
\end{equation}

For a constant $q$, increasing the number of input views $K$ does not
increase the peak GPU memory, since the views are processed
sequentially during each matrix-free evaluation. Thus,
\textit{the peak GPU memory grows linearly with $N_v$}. Given a certain number of input views, the peak GPU memory only depends on the number of vertices of the input mesh.

\begin{table}[t]
\centering
\caption{The Datasets Used for Evaluation.}
\label{tab:dataset_summary}
\scriptsize
\setlength{\tabcolsep}{2.5pt}
\renewcommand{\arraystretch}{1.08}
\resizebox{\columnwidth}{!}{%
\begin{tabular}{lccccc}
\toprule
Dataset
& Real-world
& Type
& Scene Scale
& RGB Resolution
& Data Source \\
\midrule

C3VD~\cite{bobrow2023}
& $\checkmark$
& surgical
& tiny
& $540\times675$
& colonoscopic video / phantom \\

OmniObject3D~\cite{wu2023omniobject3d}
& $\checkmark$
& object-centric
& small
& $800\times800$
& structured-light scanner \\

BundleFusion~\cite{dai2017bundlefusion}
& $\checkmark$
& indoor
& medium
& $480\times640$
& RGB-D camera \\

Oxford Spires~\cite{tao2026oxford}
& $\checkmark$
& outdoor
& large
& $1080\times1440$
& RGB camera / LiDAR \\

\bottomrule
\end{tabular}%
}
\end{table}

\section{Experiments}
\label{sec:experiments}
In this section, we evaluate CF-Splat on 33 scenes from four real-world datasets spanning different scene scales, and compare it against 12 state-of-the-art baselines. All experiments are conducted on a consumer-grade laptop equipped with an NVIDIA RTX 4060 GPU (8 GB) and an AMD Ryzen 9 7940HS CPU (48 GB).

\subsection{Baselines}
\label{sec:baselines}

We select 12 baselines from complementary lines of splatting research discussed
in Section~\ref{sec:related_work} to comprehensively evaluate CF-Splat in
terms of representation for NVS, computational efficiency, and optimization strategy.

\paragraph{Primitive representations}
We compare against Gaussian-based methods, including
3DGS~\cite{kerbl20233d}, 2DGS~\cite{huang20242d},
PGSR~\cite{chen2024pgsr}, and MVGS~\cite{du2024mvgs};
triangle-based methods, including
Triangle-Splat~\cite{held2026triangle} and
Mesh-Splat~\cite{held2025meshsplatting}; and the hybrid
Gaussian--triangle representation GaMeS~\cite{waczynska2024games}.

\paragraph{Accelerated and memory-efficient Methods}
We include FastGS~\cite{ren2026fastgs},
Speedy-Splat~\cite{hanson2025speedy} and Taming-3DGS~\cite{mallick2024taming} to evaluate computing efficiency
and memory consumption.

\paragraph{Optimization strategies and solvers}
We further compare with 3DGS-LM~\cite{hollein2025_3dgslm} and
LM-RS~\cite{pehlivan2026matrixfree}, which represent alternative
optimization and numerical solution strategies for splatting-based
representations.

In addition, we implement CF-Splat (Adam), a 
gradient-based variant from CF-Splat. It optimizes the same objective in \eqref{eq:global_lls} over $\mathbf{X}$ using Adam~\cite{kingma2014adam}, instead of
solving its normal system using the matrix-free PCG solver in Section~\ref{sec:scalable_solver}. It also serves as a representative gradient-based mesh-rendering baseline following the optimization paradigm commonly used
with differentiable rasterizers~\cite{kato2018neural,liu2019soft,liu2020general}.

\subsection{Datasets}
\label{sec:datasets}
The datasets used for evaluation are summarized in Table~\ref{tab:dataset_summary}. 

C3VD~\cite{bobrow2023} is a colonoscopy dataset acquired using a high-definition clinical colonoscope and high-fidelity colon phantoms. We select 10 sequences covering three anatomical regions: cecum (\textit{c\_t1\_b}, \textit{c\_t2\_b}, and \textit{c\_t3\_a}), sigmoid colon (\textit{s\_t1\_a}, \textit{s\_t2\_a}, and \textit{s\_t3\_a}), and transverse colon (\textit{t\_t1\_b}, \textit{t\_t2\_c}, \textit{t\_t4\_a}, and \textit{t\_t4\_b}). 

OmniObject3D~\cite{wu2023omniobject3d} is a real-world 3D object dataset acquired using structured-light scanning, covering a wide range of object categories. We select 10 objects: \textit{apple}, \textit{backpack}, \textit{ball}, \textit{bed}, \textit{banana}, \textit{battery}, \textit{book}, \textit{bottle}, \textit{bowl}, and \textit{box}.

BundleFusion~\cite{dai2017bundlefusion} provides RGB-D sequences of real indoor environments captured using an iPad-based RGB-D scanning system. We evaluate all 8 released scenes: \textit{apt0}, \textit{apt1}, \textit{apt2}, \textit{copyroom}, \textit{office0}, \textit{office1}, \textit{office2}, and \textit{office3}. 

Oxford Spires~\cite{tao2026oxford} is a large-scale multimodal dataset captured across Oxford using a custom-built multi-sensor platform equipped with RGB cameras and LiDAR. We select 5 representative sites: \textit{college}, \textit{quarter}, \textit{library}, \textit{church}, and \textit{palace}.

These four datasets span surgical, object-centric, indoor, and outdoor environments across scene scales ranging from tiny to large, enabling a comprehensive evaluation of the performance and scalability of the compared methods.

\subsection{Experimental Setup}
\label{sec:experimental_setup}
To ensure a fair comparison, all evaluated methods use the same posed RGB observations and geometry source obtained from identical sources. 

For OmniObject3D, every 10-th frame is held out as a test view. For C3VD, BundleFusion, and Oxford Spires, every 20-th frame is selected as a keyframe, and every 10-th keyframe is held out as a test view. The remaining frames or keyframes are used as input observations. For Oxford Spires, regions without valid geometric support, such as sky areas, are excluded from both optimization and quantitative evaluation for all compared methods. For qualitative visualization only, the corresponding invalid regions in the rendered images are replaced with ground-truth pixels to avoid distracting empty regions. 

For C3VD and OmniObject3D, we use the provided meshes and camera poses. For BundleFusion, we reconstruct the indoor geometry by applying TSDF fusion~\cite{curless1996volumetric} with a 3cm voxel resolution to the RGB-D observations using camera poses estimated with COLMAP~\cite{schoenberger2016mvs,schoenberger2016sfm}. For Oxford Spires, we reconstruct the large-scale outdoor meshes from LiDAR measurements using VDBFusion~\cite{vizzo2022vdbfusion} with a 0.5m voxel resolution, with the LiDAR poses estimated by Vilens-SLAM~\cite{wisth2022vilens} and refined by HBA~\cite{liu2023large}. The resulting meshes are used directly by GaMeS, CF-Splat (Adam), and CF-Splat, while vertices extracted from the same meshes are used to initialize the remaining baselines. 

For CF-Splat and CF-Splat (Adam), unless otherwise specified, we use
a maximum SH degree of $L=3$ for C3VD and OmniObject3D, and $L=2$ for BundleFusion and Oxford Spires. Before constructing the LLS system, we perform 10 gradient-guided
densification rounds. We set the supported- and unsupported-band
regularization strengths to $\gamma_s=10^{-3}$ and
$\gamma_u=10^{3}$, respectively, with the confidence
lower bound $\epsilon_w=10^{-3}$.
The matrix-free PCG solver uses a relative residual tolerance
of $\delta=10^{-6}$ and a maximum of $T_{\max}=300$
iterations. All other baselines are evaluated using their publicly released implementations and default configurations, with 30K optimization iterations. 

\subsection{Metrics}
We evaluate NVS quality using peak signal-to-noise ratio (PSNR), structural similarity index measure (SSIM), and learned perceptual image patch similarity (LPIPS). PSNR measures pixel-level reconstruction fidelity, SSIM evaluates structural similarity, and LPIPS assesses perceptual similarity using deep visual features. Computational efficiency is measured in terms of peak GPU memory consumption (GB) and computation time (min). To qualitatively assess the resulting scene appearances, we directly visualize the reconstructed models. For the Gaussian-based methods, the resulting Gaussian models are visualized using SuperSplat\footnote{\url{https://superspl.at/}}. For the triangle-based methods, we extract the corresponding meshes using their officially provided extraction code and visualize them using MeshLab\footnote{\url{https://www.meshlab.net/}}. The mesh representations produced by CF-Splat and CF-Splat (Adam) are likewise visualized using MeshLab for consistent comparison.

\subsection{Novel-View Synthesis Results}
\label{sec:nvs_results}

Table~\ref{tab:c3vd_quantitative}, \ref{tab:omniobject_quantitative},
\ref{tab:bundlefusion_quantitative}, and
\ref{tab:oxford_spires_quantitative} report the quantitative NVS results,
with the corresponding qualitative comparisons shown in
Fig.~\ref{fig:c3vd}, \ref{fig:omni}, \ref{fig:bundle}, and
\ref{fig:oxford}.  Overall,
CF-Splat achieves the best or near-best average NVS quality across
the four datasets.

\subsubsection{C3VD Dataset}

\begin{figure*}[!t]
    \centering
    \includegraphics[width=6.8in]{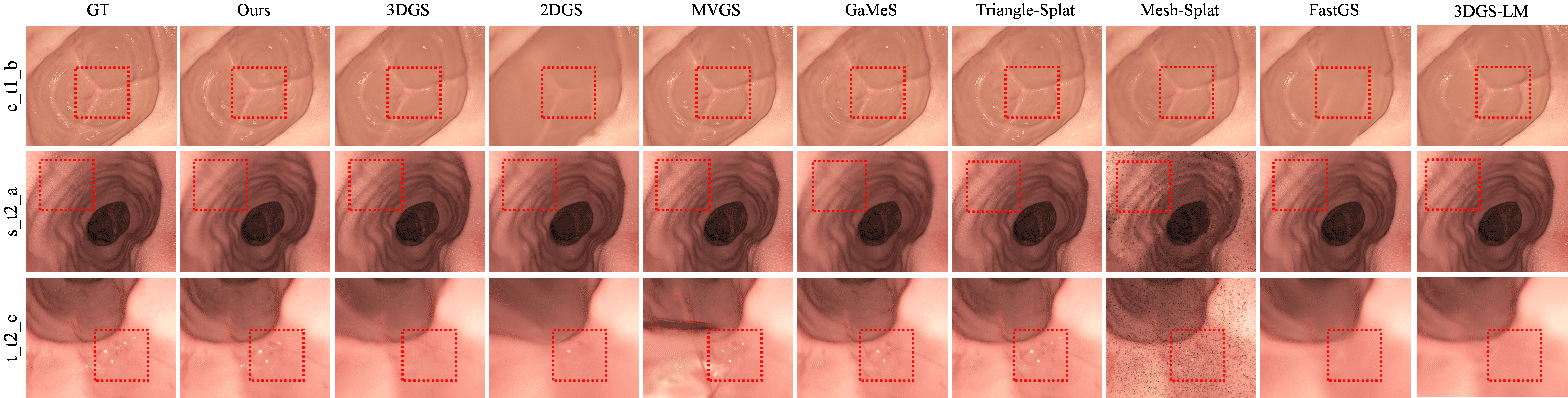}
    \caption{Qualitative NVS comparison results on \textit{c\_t1\_b}, \textit{s\_t2\_a} and \textit{t\_t2\_c} scenes from C3VD dataset~\cite{bobrow2023}. The red boxes highlight the differences.}
    \label{fig:c3vd}
\end{figure*}

C3VD contains narrow endoscopic scenes with strong specular highlights and
substantial cross-view appearance variation, making view-dependent appearance
particularly challenging. As shown in Table~\ref{tab:c3vd_quantitative},
CF-Splat achieves the strongest overall PSNR and SSIM while remaining
competitive in LPIPS. The qualitative results of CF-Splat in
Fig.~\ref{fig:c3vd} also show stable local structures under the dynamic
view-dependent illumination.

Compared Gaussian- and triangle-based baselines provide greater flexibility for fitting dynamic lighting through iterative adjustments of primitive positions, shapes, and opacity. However, such nonlinear optimization may converge to a local minimizer that better fits the training views but produces inconsistent appearance in novel views. In particular, irregular primitive distributions and opacity can introduce noticeable image-space blurring, as observed on the \textit{c\_t1\_b} and \textit{t\_t2\_c} scenes in Fig.~\ref{fig:c3vd}. 

CF-Splat instead estimates the per-vertex SH coefficients jointly from all calibrated observations through the global LLS formulation in \eqref{eq:global_lls}. The adaptive SH regularization further constrains directional components that are insufficiently supported by the available views. As proved in Section~\ref{sec:global_lls}, the LLS formulation admits a unique global minimizer, avoiding the local-minimizer ambiguity of iterative nonlinear fitting. This results that CF-Splat remains effective under strong cross-view photometric variation while preserving consistent novel-view appearance.

\begin{table*}[t]
\centering
\caption{Quantitative NVS Comparison Results on C3VD Dataset~\cite{bobrow2023}. {\setlength{\fboxsep}{0.5pt}\colorbox{orange!85}{First}}, {\setlength{\fboxsep}{0.5pt}\colorbox{orange!50}{Second}}, and {\setlength{\fboxsep}{0.5pt}\colorbox{orange!15}{Third}} Indicate Ranking Results.}
\label{tab:c3vd_quantitative}
\resizebox{\textwidth}{!}{%
\begin{tabular}{l *{33}{c}}
\toprule
Method
& \multicolumn{3}{c}{\textit{c\_t1\_b}}
& \multicolumn{3}{c}{\textit{c\_t2\_b}}
& \multicolumn{3}{c}{\textit{c\_t3\_a}}
& \multicolumn{3}{c}{\textit{s\_t1\_a}}
& \multicolumn{3}{c}{\textit{s\_t2\_a}}
& \multicolumn{3}{c}{\textit{s\_t3\_a}}
& \multicolumn{3}{c}{\textit{t\_t1\_b}}
& \multicolumn{3}{c}{\textit{t\_t2\_c}}
& \multicolumn{3}{c}{\textit{t\_t4\_a}}
& \multicolumn{3}{c}{\textit{t\_t4\_b}}
& \multicolumn{3}{c}{\textit{Average}} \\
\cmidrule(lr){2-4}
\cmidrule(lr){5-7}
\cmidrule(lr){8-10}
\cmidrule(lr){11-13}
\cmidrule(lr){14-16}
\cmidrule(lr){17-19}
\cmidrule(lr){20-22}
\cmidrule(lr){23-25}
\cmidrule(lr){26-28}
\cmidrule(lr){29-31}
\cmidrule(lr){32-34}
& PSNR & SSIM & LPIPS
& PSNR & SSIM & LPIPS
& PSNR & SSIM & LPIPS
& PSNR & SSIM & LPIPS
& PSNR & SSIM & LPIPS
& PSNR & SSIM & LPIPS
& PSNR & SSIM & LPIPS
& PSNR & SSIM & LPIPS
& PSNR & SSIM & LPIPS
& PSNR & SSIM & LPIPS
& PSNR & SSIM & LPIPS \\
\midrule

3DGS~\cite{kerbl20233d}
& 31.95 & 0.902 & 0.356
& \cTwo 34.31 & \cThree 0.918 & 0.267
& 33.49 & \cThree 0.912 & 0.235
& 37.17 & 0.914 & 0.394
& 35.65 & 0.893 & 0.263
& 36.17 & 0.895 & 0.325
& 37.67 & 0.924 & 0.306
& 32.11 & 0.909 & 0.416
& 32.64 & 0.888 & 0.246
& 32.24 & 0.847 & 0.332
& 34.34 & 0.900 & 0.314 \\

2DGS~\cite{huang20242d}
& 28.01 & 0.870 & 0.450
& 31.77 & 0.903 & 0.308
& 30.43 & 0.886 & 0.304
& 36.01 & 0.911 & 0.404
& 34.37 & 0.876 & 0.305
& 33.86 & 0.883 & 0.367
& 37.01 & 0.920 & 0.317
& 30.71 & 0.906 & 0.417
& 29.81 & 0.860 & 0.302
& 30.90 & 0.823 & 0.373
& 32.29 & 0.884 & 0.355 \\

PGSR~\cite{chen2024pgsr}
& 30.72 & 0.892 & 0.369
& 32.57 & 0.904 & 0.284
& 31.81 & 0.889 & 0.274
& 36.57 & 0.913 & 0.374
& 35.10 & 0.887 & 0.262
& 35.40 & 0.892 & 0.314
& 37.24 & 0.920 & 0.289
& 31.61 & 0.906 & 0.415
& 31.98 & 0.874 & 0.259
& 32.13 & 0.836 & 0.337
& 33.51 & 0.891 & 0.318 \\

MVGS~\cite{du2024mvgs}
& 32.63 & 0.907 & 0.326
& \cOne 34.90 & \cOne 0.923 & \cThree 0.249
& \cThree 33.95 & \cTwo 0.915 & \cThree 0.222
& \cOne 37.67 & \cTwo 0.918 & 0.377
& \cOne 36.83 & \cOne 0.913 & \cOne 0.208
& \cOne 36.67 & \cTwo 0.903 & 0.298
& \cTwo 38.48 & \cThree 0.930 & 0.278
& 30.88 & 0.903 & 0.399
& \cTwo 33.35 & \cTwo 0.895 & \cThree 0.218
& 32.92 & 0.855 & 0.306
& 34.83 & 0.906 & 0.288 \\

GaMeS~\cite{waczynska2024games}
& 30.79 & 0.896 & 0.335
& 29.50 & 0.888 & 0.327
& 31.39 & 0.887 & 0.271
& 36.41 & 0.913 & 0.384
& 32.70 & 0.845 & 0.344
& 34.22 & 0.873 & 0.360
& 35.88 & 0.909 & 0.316
& 33.65 & 0.920 & 0.358
& 30.36 & 0.849 & 0.336
& 20.71 & 0.752 & 0.476
& 31.56 & 0.873 & 0.351 \\

Triangle-Splat~\cite{held2026triangle}
& \cTwo 33.29 & \cTwo 0.911 & \cTwo 0.261
& 32.40 & 0.908 & \cTwo 0.237
& \cTwo 34.11 & \cOne 0.919 & \cOne 0.181
& 36.99 & \cTwo 0.918 & \cOne 0.340
& 35.65 & \cThree 0.899 & \cTwo 0.210
& 35.55 & 0.894 & \cOne 0.248
& 38.10 & \cTwo 0.931 & \cTwo 0.234
& 34.91 & 0.925 & \cOne 0.286
& 32.10 & 0.886 & \cTwo 0.206
& 33.30 & 0.884 & 0.213
& 34.64 & \cThree 0.907 & \cOne 0.242 \\

Mesh-Splat~\cite{held2025meshsplatting}
& 29.71 & 0.890 & 0.321
& 27.85 & 0.878 & 0.310
& 29.42 & 0.878 & 0.269
& 25.58 & 0.606 & 0.470
& 22.91 & 0.582 & 0.476
& 22.42 & 0.568 & 0.491
& 22.79 & 0.527 & 0.501
& 20.97 & 0.436 & 0.535
& 21.17 & 0.526 & 0.490
& 21.92 & 0.588 & 0.500
& 24.47 & 0.648 & 0.436 \\

FastGS~\cite{ren2026fastgs}
& 32.95 & \cThree 0.910 & 0.316
& 33.50 & 0.913 & 0.279
& 33.82 & 0.911 & 0.229
& \cThree 37.32 & \cThree 0.917 & 0.376
& \cTwo 36.10 & \cTwo 0.901 & 0.245
& \cTwo 36.62 & \cOne 0.905 & \cThree 0.286
& \cThree 38.45 & \cTwo 0.931 & \cThree 0.268
& \cThree 35.55 & \cThree 0.926 & \cThree 0.333
& 32.69 & \cThree 0.892 & 0.234
& 35.54 & \cThree 0.905 & 0.207
& \cThree 35.25 & \cTwo 0.911 & \cThree 0.277 \\

Speedy-Splat~\cite{hanson2025speedy}
& 30.93 & 0.890 & 0.401
& 33.43 & 0.905 & 0.315
& 32.54 & 0.896 & 0.282
& 36.65 & 0.911 & 0.409
& 35.38 & 0.883 & 0.297
& 35.91 & 0.894 & 0.331
& 37.36 & 0.918 & 0.330
& 32.28 & 0.910 & 0.419
& 32.31 & 0.882 & 0.264
& 31.68 & 0.828 & 0.369
& 33.85 & 0.892 & 0.342 \\

Taming-3DGS~\cite{mallick2024taming}
& \cThree 33.04 & 0.897 & \cThree 0.312
& 33.99 & 0.900 & 0.279
& \cOne 34.39 & 0.906 & \cTwo 0.202
& 37.15 & 0.898 & 0.392
& 35.98 & 0.886 & 0.251
& 36.38 & 0.886 & 0.304
& 37.97 & 0.911 & 0.289
& \cOne 35.89 & 0.910 & 0.338
& \cThree 32.70 & 0.876 & 0.239
& \cThree 35.86 & 0.897 & \cThree 0.195
& \cTwo 35.34 & 0.897 & 0.280 \\

3DGS-LM~\cite{hollein2025_3dgslm}
& 31.98 & 0.903 & 0.346
& \cThree 34.25 & \cThree 0.918 & 0.264
& 32.45 & 0.904 & 0.256
& \cThree 37.32 & \cThree 0.917 & 0.380
& \cThree 36.02 & 0.897 & 0.250
& \cThree 36.44 & \cThree 0.899 & 0.306
& 38.06 & 0.926 & 0.292
& 32.08 & 0.909 & 0.422
& \cThree 32.70 & 0.890 & 0.232
& 32.53 & 0.851 & 0.316
& 34.38 & 0.902 & 0.306 \\

LM-RS~\cite{pehlivan2026matrixfree}
& 32.85 & 0.895 & 0.335
& 33.13 & 0.892 & 0.316
& 32.44 & 0.876 & 0.304
& 36.88 & 0.910 & 0.402
& 33.90 & 0.845 & 0.370
& 35.22 & 0.869 & 0.386
& 36.38 & 0.902 & 0.358
& 35.13 & 0.919 & 0.369
& 31.31 & 0.834 & 0.386
& 29.90 & 0.810 & 0.383
& 33.71 & 0.875 & 0.361 \\

\midrule

CF-Splat (Adam)
& 32.46 & 0.900 & 0.322
& 29.58 & 0.881 & 0.326
& 30.90 & 0.869 & 0.302
& \cTwo 37.40 & \cOne 0.927 & \cTwo 0.358
& 27.38 & 0.820 & 0.362
& 29.52 & 0.869 & 0.355
& 32.56 & 0.902 & 0.340
& 34.76 & \cOne 0.931 & 0.336
& 27.42 & 0.822 & 0.369
& \cTwo 38.56 & \cOne 0.971 & \cOne 0.090
& 32.05 & 0.889 & 0.316 \\

CF-Splat
& \cOne 34.28 & \cOne 0.926 & \cOne 0.238
& 32.92 & \cTwo 0.920 & \cOne 0.232
& 32.89 & 0.878 & 0.295
& 36.57 & \cThree 0.917 & \cThree 0.373
& 34.99 & 0.887 & \cThree 0.220
& 35.26 & 0.887 & \cTwo 0.259
& \cOne 38.57 & \cOne 0.937 & \cOne 0.219
& \cTwo 35.70 & \cTwo 0.928 & \cTwo 0.321
& \cOne 33.89 & \cOne 0.905 & \cOne 0.197
& \cOne 39.30 & \cTwo 0.947 & \cTwo 0.162
& \cOne 35.44 & \cOne 0.913 & \cTwo 0.252 \\

\bottomrule
\end{tabular}%
}
\end{table*}

\subsubsection{OmniObject3D Dataset}

\begin{table*}[t]
\centering
\caption{Quantitative NVS Comparison Results on OmniObject3D Dataset~\cite{wu2023omniobject3d}. {\setlength{\fboxsep}{0.5pt}\colorbox{orange!85}{First}}, {\setlength{\fboxsep}{0.5pt}\colorbox{orange!50}{Second}}, and {\setlength{\fboxsep}{0.5pt}\colorbox{orange!15}{Third}} Indicate Ranking Results.}
\label{tab:omniobject_quantitative}
\resizebox{\textwidth}{!}{%
\begin{tabular}{l *{33}{c}}
\toprule
Method
& \multicolumn{3}{c}{\textit{apple}}
& \multicolumn{3}{c}{\textit{backpack}}
& \multicolumn{3}{c}{\textit{ball}}
& \multicolumn{3}{c}{\textit{bed}}
& \multicolumn{3}{c}{\textit{banana}}
& \multicolumn{3}{c}{\textit{battery}}
& \multicolumn{3}{c}{\textit{book}}
& \multicolumn{3}{c}{\textit{bottle}}
& \multicolumn{3}{c}{\textit{bowl}}
& \multicolumn{3}{c}{\textit{box}}
& \multicolumn{3}{c}{\textit{Average}} \\
\cmidrule(lr){2-4}
\cmidrule(lr){5-7}
\cmidrule(lr){8-10}
\cmidrule(lr){11-13}
\cmidrule(lr){14-16}
\cmidrule(lr){17-19}
\cmidrule(lr){20-22}
\cmidrule(lr){23-25}
\cmidrule(lr){26-28}
\cmidrule(lr){29-31}
\cmidrule(lr){32-34}
& PSNR & SSIM & LPIPS
& PSNR & SSIM & LPIPS
& PSNR & SSIM & LPIPS
& PSNR & SSIM & LPIPS
& PSNR & SSIM & LPIPS
& PSNR & SSIM & LPIPS
& PSNR & SSIM & LPIPS
& PSNR & SSIM & LPIPS
& PSNR & SSIM & LPIPS
& PSNR & SSIM & LPIPS
& PSNR & SSIM & LPIPS \\
\midrule

3DGS~\cite{kerbl20233d}
& 24.91 & 0.577 & 0.119
& 38.03 & 0.982 & 0.054
& 34.22 & 0.961 & 0.090
& 37.62 & 0.982 & 0.048
& 33.74 & 0.926 & 0.033
& 33.82 & 0.981 & 0.019
& 40.64 & \cTwo 0.996 & 0.020
& 25.57 & 0.679 & 0.127
& 38.45 & 0.990 & 0.058
& 23.98 & 0.859 & 0.138
& 33.10 & 0.893 & 0.070 \\

2DGS~\cite{huang20242d}
& 41.46 & 0.985 & 0.035
& 39.45 & \cThree 0.988 & 0.038
& 38.05 & 0.986 & 0.038
& 38.15 & \cTwo 0.989 & \cThree 0.027
& 38.38 & 0.982 & 0.024
& 37.63 & 0.992 & 0.011
& 40.77 & \cOne 0.997 & 0.016
& 43.52 & \cOne 0.998 & \cTwo 0.007
& 39.57 & \cThree 0.994 & 0.035
& 37.08 & 0.984 & 0.026
& 39.41 & 0.989 & 0.026 \\

PGSR~\cite{chen2024pgsr}
& 40.36 & 0.986 & 0.033
& 38.99 & 0.986 & 0.038
& 35.94 & 0.979 & 0.046
& 39.31 & 0.987 & 0.029
& 42.19 & 0.991 & 0.009
& 43.23 & 0.996 & 0.005
& 40.50 & \cThree 0.995 & 0.016
& 42.82 & \cTwo 0.996 & \cThree 0.008
& 39.49 & 0.991 & 0.040
& 39.10 & 0.988 & 0.020
& 40.19 & 0.990 & 0.024 \\

MVGS~\cite{du2024mvgs}
& \cOne 46.04 & \cTwo 0.994 & \cTwo 0.013
& \cTwo 40.08 & \cTwo 0.989 & \cOne 0.024
& \cTwo 39.36 & \cThree 0.989 & \cOne 0.021
& \cOne 40.94 & \cThree 0.988 & \cOne 0.019
& \cOne 46.30 & \cThree 0.996 & \cOne 0.005
& \cOne 46.65 & 0.996 & \cOne 0.002
& \cThree 42.40 & \cThree 0.995 & \cTwo 0.012
& 43.81 & \cThree 0.995 & \cThree 0.008
& \cThree 40.99 & 0.993 & \cTwo 0.024
& \cTwo 42.50 & \cThree 0.994 & \cTwo 0.009
& \cTwo 42.91 & \cThree 0.993 & \cTwo 0.014 \\

GaMeS~\cite{waczynska2024games}
& 34.51 & 0.981 & 0.037
& 37.75 & 0.981 & 0.053
& 34.47 & 0.943 & 0.117
& 36.54 & 0.977 & 0.056
& 43.68 & \cThree 0.996 & 0.008
& 42.39 & \cThree 0.997 & 0.006
& 40.54 & \cTwo 0.996 & 0.018
& 43.76 & \cOne 0.998 & \cThree 0.008
& 38.26 & 0.991 & 0.042
& 37.95 & 0.989 & 0.029
& 38.98 & 0.985 & 0.037 \\

Triangle-Splat~\cite{held2026triangle}
& 44.89 & \cOne 0.996 & \cTwo 0.013
& \cThree 39.91 & \cOne 0.991 & \cOne 0.024
& 38.15 & \cThree 0.989 & \cThree 0.023
& 36.13 & 0.978 & 0.047
& 41.19 & 0.993 & 0.014
& 40.67 & 0.995 & 0.009
& 37.29 & 0.993 & 0.025
& 39.50 & 0.993 & 0.016
& 35.85 & 0.987 & 0.052
& 36.48 & 0.989 & 0.026
& 39.01 & 0.991 & 0.025 \\

Mesh-Splat~\cite{held2025meshsplatting}
& 33.67 & 0.944 & 0.093
& 35.06 & 0.960 & 0.083
& 27.23 & 0.875 & 0.133
& 27.84 & 0.903 & 0.109
& 34.67 & 0.970 & 0.035
& 35.38 & 0.984 & 0.019
& 33.43 & 0.962 & 0.071
& 35.32 & 0.977 & 0.030
& 28.82 & 0.905 & 0.151
& 29.17 & 0.928 & 0.089
& 32.06 & 0.941 & 0.081 \\

FastGS~\cite{ren2026fastgs}
& 44.65 & \cTwo 0.994 & 0.022
& 39.68 & \cTwo 0.989 & 0.038
& 37.79 & 0.980 & 0.058
& 39.95 & \cTwo 0.989 & 0.032
& 44.90 & \cTwo 0.997 & 0.008
& 45.52 & \cTwo 0.998 & \cThree 0.004
& 41.46 & \cOne 0.997 & 0.018
& 43.88 & \cOne 0.998 & 0.009
& 39.76 & 0.993 & 0.046
& 40.79 & 0.993 & 0.024
& 41.84 & \cThree 0.993 & 0.026 \\

Speedy-Splat~\cite{hanson2025speedy}
& 39.58 & 0.970 & 0.095
& 37.75 & 0.976 & 0.076
& 35.41 & 0.956 & 0.109
& 38.60 & 0.981 & 0.056
& 40.50 & 0.986 & 0.028
& 42.92 & 0.994 & 0.013
& 40.13 & 0.992 & 0.037
& 41.53 & 0.993 & 0.021
& 38.12 & 0.987 & 0.077
& 38.19 & 0.984 & 0.047
& 39.27 & 0.982 & 0.056 \\

Taming-3DGS~\cite{mallick2024taming}
& 44.80 & 0.976 & 0.035
& 39.65 & 0.971 & 0.050
& 38.52 & 0.970 & 0.049
& \cThree 40.25 & 0.971 & 0.043
& 45.33 & 0.978 & 0.026
& \cThree 45.88 & 0.979 & 0.023
& 41.60 & 0.977 & 0.034
& 43.92 & 0.978 & 0.027
& 39.98 & 0.975 & 0.055
& 41.54 & 0.977 & 0.032
& 42.15 & 0.975 & 0.037 \\

3DGS-LM~\cite{hollein2025_3dgslm}
& \cTwo 45.41 & \cOne 0.996 & \cThree 0.016
& \cOne 40.13 & \cOne 0.991 & \cThree 0.029
& \cThree 38.90 & \cTwo 0.990 & 0.027
& \cTwo 40.63 & \cOne 0.991 & \cTwo 0.022
& \cTwo 45.89 & \cOne 0.998 & \cTwo 0.006
& \cTwo 46.14 & \cOne 0.999 & \cTwo 0.003
& 41.83 & \cOne 0.997 & \cThree 0.014
& \cThree 44.30 & \cOne 0.998 & \cTwo 0.007
& 40.17 & \cTwo 0.995 & \cThree 0.032
& \cThree 41.71 & \cTwo 0.996 & \cThree 0.013
& \cThree 42.51 & \cOne 0.995 & \cThree 0.017 \\

LM-RS~\cite{pehlivan2026matrixfree}
& 43.06 & 0.988 & 0.041
& 37.32 & 0.976 & 0.068
& 35.83 & 0.961 & 0.093
& 36.33 & 0.969 & 0.071
& 43.28 & 0.993 & 0.012
& 43.87 & \cThree 0.997 & 0.005
& 40.11 & 0.994 & 0.027
& 42.65 & \cTwo 0.996 & 0.013
& 37.03 & 0.982 & 0.085
& 38.70 & 0.987 & 0.034
& 39.82 & 0.984 & 0.045 \\

\midrule

CF-Splat (Adam)
& 40.12 & 0.973 & 0.036
& 36.72 & 0.972 & 0.039
& 36.50 & 0.951 & 0.053
& 35.48 & 0.963 & 0.065
& 44.34 & 0.985 & 0.013
& 40.00 & 0.983 & 0.021
& \cTwo 44.82 & 0.984 & 0.031
& \cTwo 47.35 & 0.990 & 0.012
& \cTwo 41.52 & 0.985 & 0.040
& 40.28 & 0.976 & 0.049
& 40.71 & 0.976 & 0.036 \\

CF-Splat
& \cThree 45.13 & \cThree 0.992 & \cOne 0.012
& 39.61 & \cTwo 0.989 & \cTwo 0.025
& \cOne 39.38 & \cOne 0.992 & \cTwo 0.022
& 39.67 & 0.985 & 0.030
& \cThree 45.76 & 0.995 & \cThree 0.007
& \cThree 45.88 & 0.996 & \cThree 0.004
& \cOne 46.21 & \cTwo 0.996 & \cOne 0.011
& \cOne 48.35 & \cOne 0.998 & \cOne 0.006
& \cOne 42.04 & \cOne 0.996 & \cOne 0.012
& \cOne 43.12 & \cOne 0.998 & \cOne 0.005
& \cOne 43.52 & \cTwo 0.994 & \cOne 0.013 \\

\bottomrule
\end{tabular}%
}
\end{table*}

\begin{figure*}[!t]
    \centering
    \includegraphics[width=6.8in]{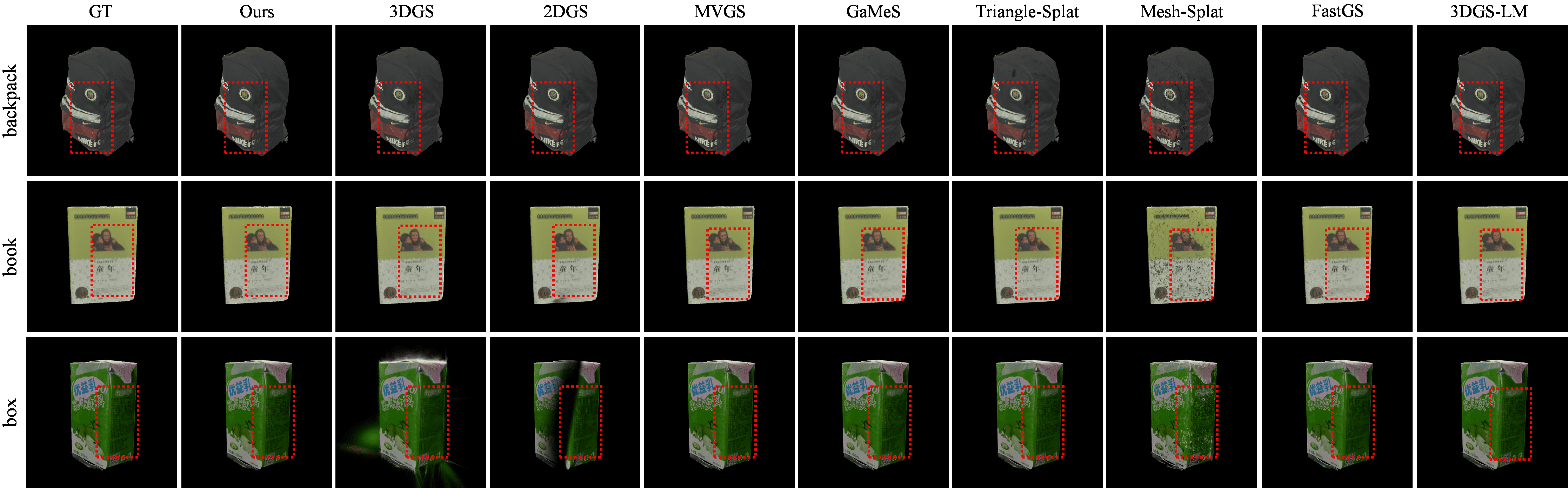}
    \caption{Qualitative NVS comparison results on \textit{backpack}, \textit{book} and \textit{box} from OmniObject3D Dataset~\cite{wu2023omniobject3d}. The red boxes highlight the rendering differences.}
    \label{fig:omni}
\end{figure*}

OmniObject3D provides accurate object geometry together with sharp boundaries
and high-frequency surface textures. It therefore offers a favorable setting
for isolating the quality of the appearance representation with limited
geometric uncertainty. CF-Splat achieves the strongest overall PSNR and
LPIPS while remaining highly competitive in SSIM, as reported in
Table~\ref{tab:omniobject_quantitative}.

The qualitative comparisons in Fig.~\ref{fig:omni} show that triangle-based
representations generally preserve fine texture structures more clearly than
spatially extended Gaussian primitives. For CF-Splat, the gradient-guided
densification in Section~\ref{sec:densification} further allocates additional
vertices to regions with stronger image-space variation. This increases local
appearance resolution without introducing additional geometric variables into
the subsequent appearance estimation. The strong performance on
OmniObject3D therefore demonstrates the benefit of combining an explicit
surface representation with adaptive spatial appearance capacity. At the same time, this fully exploits the advantage of the closed-form formulation by enabling CF-Splat to recover the unique global optimum of the appearance estimation problem.

\subsubsection{BundleFusion}

\begin{table*}[t]
\centering
\tiny
\setlength{\tabcolsep}{2pt}
\renewcommand{\arraystretch}{0.95}
\caption{Quantitative NVS Comparison Results on BundleFusion Dataset~\cite{dai2017bundlefusion}. {\setlength{\fboxsep}{0.5pt}\colorbox{orange!85}{First}}, {\setlength{\fboxsep}{0.5pt}\colorbox{orange!50}{Second}}, and {\setlength{\fboxsep}{0.5pt}\colorbox{orange!15}{Third}} Indicate Ranking Results.}
\label{tab:bundlefusion_quantitative}
\resizebox{0.95\textwidth}{!}{%
\begin{tabular}{l *{27}{c}}
\toprule
Method
& \multicolumn{3}{c}{\textit{apt0}}
& \multicolumn{3}{c}{\textit{apt1}}
& \multicolumn{3}{c}{\textit{apt2}}
& \multicolumn{3}{c}{\textit{copyroom}}
& \multicolumn{3}{c}{\textit{office0}}
& \multicolumn{3}{c}{\textit{office1}}
& \multicolumn{3}{c}{\textit{office2}}
& \multicolumn{3}{c}{\textit{office3}}
& \multicolumn{3}{c}{\textit{Average}} \\
\cmidrule(lr){2-4}
\cmidrule(lr){5-7}
\cmidrule(lr){8-10}
\cmidrule(lr){11-13}
\cmidrule(lr){14-16}
\cmidrule(lr){17-19}
\cmidrule(lr){20-22}
\cmidrule(lr){23-25}
\cmidrule(lr){26-28}
& PSNR & SSIM & LPIPS
& PSNR & SSIM & LPIPS
& PSNR & SSIM & LPIPS
& PSNR & SSIM & LPIPS
& PSNR & SSIM & LPIPS
& PSNR & SSIM & LPIPS
& PSNR & SSIM & LPIPS
& PSNR & SSIM & LPIPS
& PSNR & SSIM & LPIPS \\
\midrule

3DGS~\cite{kerbl20233d}
& 27.49 & 0.838 & \cTwo 0.231
& 27.32 & 0.815 & \cTwo 0.269
& 22.28 & 0.709 & 0.347
& 28.29 & 0.903 & \cThree 0.194
& 27.09 & 0.853 & \cTwo 0.258
& 25.36 & 0.791 & \cTwo 0.278
& 24.87 & 0.742 & \cTwo 0.314
& 25.00 & 0.823 & \cTwo 0.265
& 25.96 & 0.809 & \cTwo 0.270 \\

2DGS~\cite{huang20242d}
& 27.26 & 0.846 & 0.271
& 27.34 & 0.831 & 0.308
& 24.01 & 0.766 & 0.347
& 28.38 & 0.910 & 0.219
& \cTwo 27.27 & \cThree 0.857 & 0.298
& 25.56 & \cThree 0.804 & 0.317
& 25.34 & 0.774 & 0.365
& \cTwo 25.69 & \cThree 0.844 & 0.288
& 26.36 & 0.829 & 0.302 \\

PGSR~\cite{chen2024pgsr}
& 27.37 & 0.841 & 0.261
& 26.69 & 0.815 & 0.302
& 23.62 & 0.753 & 0.339
& 28.25 & 0.908 & 0.209
& 26.89 & 0.853 & 0.280
& 25.38 & 0.799 & 0.301
& 24.51 & 0.756 & 0.348
& 25.42 & 0.835 & 0.285
& 26.02 & 0.820 & \cThree 0.291 \\

MVGS~\cite{du2024mvgs}
& \cThree 28.45 & \cThree 0.852 & \cThree 0.236
& \cThree 28.03 & 0.828 & \cThree 0.275
& 23.98 & 0.746 & \cThree 0.335
& \cOne 29.67 & \cThree 0.916 & \cOne 0.185
& 25.14 & 0.848 & 0.308
& 24.69 & 0.796 & 0.325
& 24.47 & 0.769 & 0.381
& 24.97 & 0.842 & 0.293
& 26.18 & 0.825 & 0.292 \\

GaMeS~\cite{waczynska2024games}
& 26.71 & 0.838 & 0.259
& 26.29 & 0.820 & 0.299
& 23.50 & 0.753 & 0.353
& 25.76 & 0.890 & 0.232
& 25.21 & 0.848 & 0.292
& 24.50 & 0.789 & 0.317
& 24.09 & 0.756 & 0.357
& 25.03 & 0.840 & 0.289
& 25.14 & 0.817 & 0.300 \\

Triangle-Splat~\cite{held2026triangle}
& 21.04 & 0.763 & 0.445
& 21.23 & 0.763 & 0.454
& 19.88 & 0.718 & 0.487
& 18.50 & 0.739 & 0.447
& 22.50 & 0.743 & 0.395
& 21.74 & 0.713 & 0.396
& 20.73 & 0.666 & 0.452
& 18.92 & 0.679 & 0.459
& 20.57 & 0.723 & 0.442 \\

Mesh-Splat~\cite{held2025meshsplatting}
& 21.40 & 0.716 & 0.464
& 20.23 & 0.671 & 0.508
& 18.16 & 0.542 & 0.554
& 18.71 & 0.669 & 0.497
& 17.99 & 0.500 & 0.528
& 18.04 & 0.531 & 0.547
& 18.34 & 0.542 & 0.580
& 18.25 & 0.562 & 0.508
& 18.89 & 0.592 & 0.523 \\

FastGS~\cite{ren2026fastgs}
& 28.18 & \cThree 0.852 & 0.264
& 27.88 & \cThree 0.835 & 0.296
& \cThree 24.40 & \cTwo 0.771 & 0.344
& 28.61 & 0.914 & 0.211
& \cThree 27.16 & \cTwo 0.865 & \cThree 0.277
& \cThree 25.62 & \cTwo 0.809 & 0.308
& \cThree 25.60 & \cTwo 0.782 & 0.349
& 25.63 & \cOne 0.846 & \cThree 0.281
& \cThree 26.64 & \cThree 0.834 & \cThree 0.291 \\

Speedy-Splat~\cite{hanson2025speedy}
& 27.17 & 0.841 & 0.299
& 27.33 & 0.827 & 0.331
& 23.77 & \cThree 0.770 & 0.379
& 28.16 & 0.907 & 0.242
& 27.10 & \cThree 0.857 & 0.310
& 25.57 & 0.803 & 0.338
& 25.37 & \cThree 0.775 & 0.390
& 25.55 & 0.843 & 0.301
& 26.25 & 0.828 & 0.324 \\

Taming-3DGS~\cite{mallick2024taming}
& \cTwo 28.46 & \cTwo 0.856 & 0.251
& \cTwo 28.17 & \cTwo 0.837 & 0.281
& \cTwo 24.46 & \cThree 0.770 & \cTwo 0.330
& \cTwo 29.35 & \cOne 0.919 & \cOne 0.185
& \cOne 27.83 & \cOne 0.870 & \cOne 0.249
& \cTwo 25.91 & \cOne 0.812 & \cThree 0.279
& \cTwo 25.94 & \cTwo 0.782 & \cThree 0.327
& \cOne 25.94 & \cTwo 0.845 & \cOne 0.256
& \cTwo 27.01 & \cTwo 0.836 & \cTwo 0.270 \\

3DGS-LM~\cite{hollein2025_3dgslm}
& 27.81 & 0.847 & 0.264
& 27.69 & 0.829 & 0.299
& 23.80 & 0.756 & 0.350
& 28.42 & 0.906 & 0.221
& 26.78 & \cThree 0.857 & 0.293
& 25.35 & 0.801 & 0.313
& 25.11 & 0.771 & 0.361
& 25.32 & 0.839 & 0.288
& 26.28 & 0.826 & 0.299 \\

LM-RS~\cite{pehlivan2026matrixfree}
& 26.79 & 0.821 & 0.332
& 26.87 & 0.808 & 0.357
& 23.84 & 0.756 & 0.414
& 28.11 & 0.894 & 0.254
& 26.46 & 0.839 & 0.333
& 25.15 & 0.777 & 0.371
& 25.43 & 0.757 & 0.413
& 25.29 & 0.828 & 0.330
& 25.99 & 0.810 & 0.351 \\

\midrule

CF-Splat (Adam)
& 25.30 & 0.800 & 0.287
& 25.21 & 0.786 & 0.312
& 22.67 & 0.685 & 0.402
& 25.50 & 0.866 & 0.252
& 24.15 & 0.817 & 0.301
& 23.32 & 0.745 & 0.327
& 23.53 & 0.710 & 0.357
& 24.40 & 0.796 & 0.340
& 23.26 & 0.776 & 0.322 \\

CF-Splat
& \cOne 28.67 & \cOne 0.866 & \cOne 0.223
& \cOne 28.47 & \cOne 0.840 & \cOne 0.260
& \cOne 26.05 & \cOne 0.786 & \cOne 0.315
& \cThree 28.95 & \cTwo 0.918 & \cTwo 0.189
& 25.96 & 0.839 & 0.291
& \cOne 27.77 & \cTwo 0.809 & \cOne 0.272
& \cOne 26.31 & \cOne 0.797 & \cOne 0.279
& \cThree 25.66 & 0.843 & 0.303
& \cOne 27.22 & \cOne 0.837 & \cOne 0.267 \\

\bottomrule
\end{tabular}%
}
\end{table*}

\begin{figure*}[!t]
    \centering
    \includegraphics[width=7.0in]{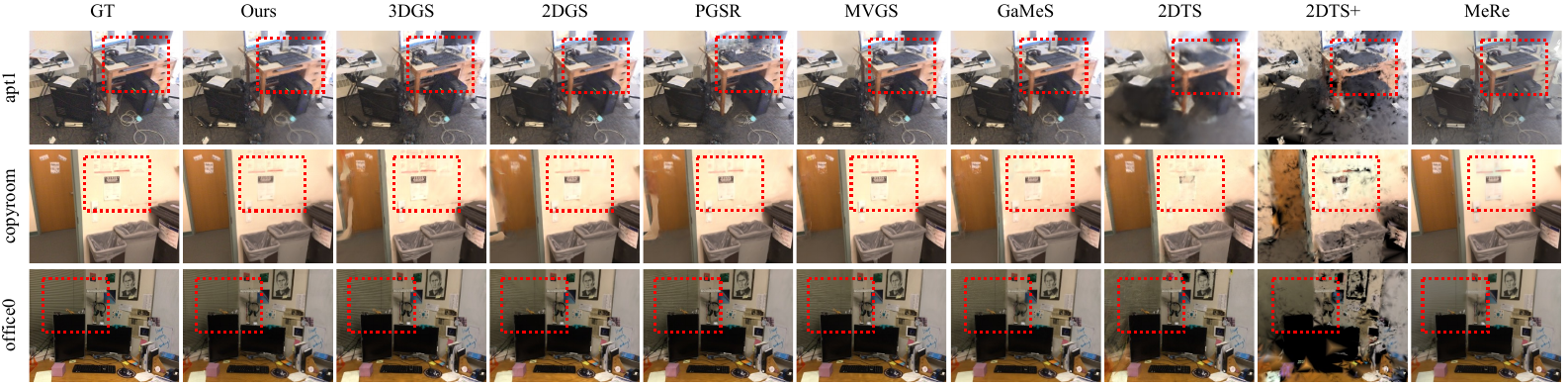}
    \caption{Qualitative NVS comparison results on \textit{apt1}, \textit{copyroom} and \textit{office0} from BundleFusion Dataset~\cite{dai2017bundlefusion}. The red boxes highlight the differences.}
    \label{fig:bundle}
\end{figure*}

BundleFusion presents a different challenge because its input geometry is
obtained from RGB-D reconstruction and therefore contains residual surface
and registration errors. Despite these imperfections, CF-Splat achieves the
best average performance across PSNR, SSIM, and LPIPS, indicating that it
does not require near-perfect input geometry.

Compared Gaussian- and triangle-based methods tend to rely on densification
and flexible primitive reshaping or moving for image-space fitting. However, in
regions with many object boundaries, as commonly observed in BundleFusion,
freely distributed primitives can be more sensitive to residual geometry
and registration errors. During optimization, these methods may compensate
for such inconsistencies by redistributing primitives and reducing their
opacity, which increases blending across object boundaries. This can produce
visible floating or blurred structures around the desk and monitor edges in
the \textit{apt1} and \textit{office0} scenes in
Fig.~\ref{fig:bundle}. Residual pose errors can further amplify these
artifacts by introducing additional misalignment among the training views.

CF-Splat handles these inconsistencies during appearance estimation without
altering the given scene geometry. The observation weighting in
\eqref{eq:observation_weight} reduces the influence of grazing-angle
measurements, while the remaining observations are jointly incorporated
into the global LLS problem in \eqref{eq:global_lls}. As illustrated in
Fig.~\ref{fig:bundle}, this formulation produces more stable appearance and
cleaner object boundaries despite moderate surface and registration errors.

\subsubsection{Oxford Spires}

\begin{figure*}[!t]
    \centering
    \includegraphics[width=\textwidth]{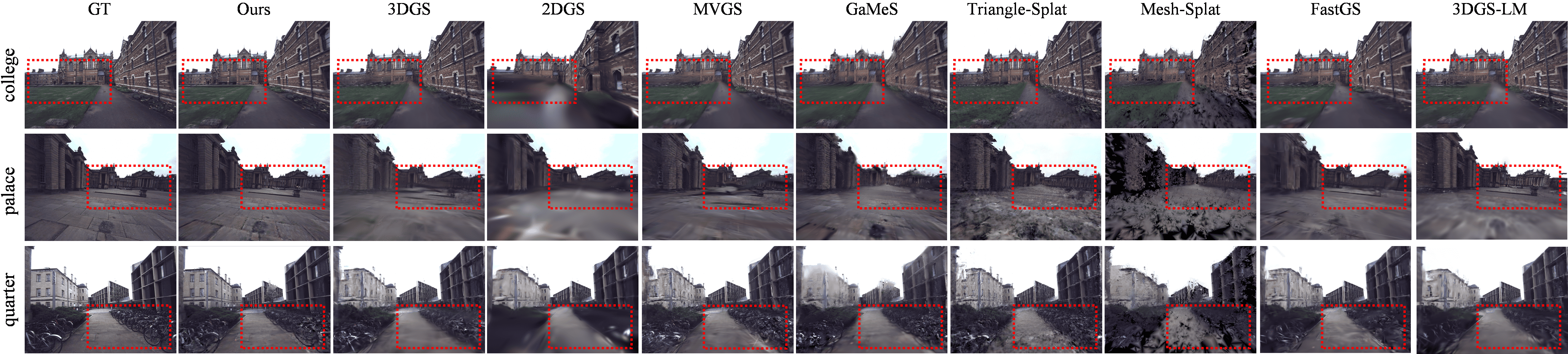}
    \caption{Qualitative NVS comparison results on \textit{college}, \textit{quarter} and \textit{palace} from Oxford Spires Dataset~\cite{tao2026oxford}. The red boxes highlight the rendering differences.}
    \label{fig:oxford}
\end{figure*}

Oxford Spires evaluates the methods at a substantially larger scale, with
large outdoor geometry, high-resolution images, and extensive multi-view
observations. Under this setting, CF-Splat maintains a clear advantage in
average PSNR and LPIPS while remaining competitive in SSIM. The qualitative
comparison in Fig.~\ref{fig:oxford} further shows that this advantage extends
to challenging large-scale structures.

For iterative splatting baselines, increasing the scene scale introduces a
larger number of primitives and correspondingly enlarges the nonlinear
optimization problem. Redundant primitives can be difficult to sufficiently
optimize or prune, resulting in visible floaters and unstable structures.
This behavior is particularly noticeable in the \textit{palace} and
\textit{quarter} scenes in Fig.~\ref{fig:oxford}. Although FastGS removes a
large number of redundant Gaussians during optimization, the remaining
primitives still need to be jointly optimized through the nonlinear
splatting process, making accurate appearance fitting increasingly difficult
at large scale.

In CF-Splat, increasing the scene scale mainly increases the dimension of
the global SH appearance state $\mathbf{X}$ in \eqref{eq:global_appearance_state}, while the global LLS formulation
in \eqref{eq:global_lls} remains linear. Therefore, the larger problem does
not introduce additional nonlinear optimization variables or local minima.
The proposed matrix-free PCG solver further solves the resulting large normal system without explicitly constructing the normal system and achieves stable numerical convergence. This allows CF-Splat to retain stable appearance
estimation as the scene scale increases. Detailed analyses of efficiency and scalability will be discussed in Section~\ref{sec:efficiency} and Section~\ref{sec:scalability}, respectively.

% Across all four datasets, the results reveal a common advantage of CF-Splat.
% Rather than relying on additional geometric freedom to absorb photometric
% residuals, CF-Splat retains the input surface and allocates model capacity
% directly to its appearance through per-vertex SH coefficients and
% gradient-guided triangle densification in Section~\ref{sec:densification}. These coefficients are jointly estimated from
% the multi-view observations through the global LLS
% formulation in \eqref{eq:global_lls}. Consequently, CF-Splat achieves competitive or even superior NVS quality across diverse imaging conditions and scene scales.

\begin{table*}[t]
\centering
\tiny
\setlength{\tabcolsep}{2pt}
\renewcommand{\arraystretch}{0.95}
\caption{Quantitative NVS Comparison Results on Oxford Spires Dataset~\cite{tao2026oxford}. {\setlength{\fboxsep}{0.5pt}\colorbox{orange!85}{First}}, {\setlength{\fboxsep}{0.5pt}\colorbox{orange!50}{Second}}, and {\setlength{\fboxsep}{0.5pt}\colorbox{orange!15}{Third}} Indicate Ranking Results.} 
\label{tab:oxford_spires_quantitative}
\resizebox{0.75\textwidth}{!}{%
\begin{tabular}{l *{18}{c}}
\toprule
Method
& \multicolumn{3}{c}{\textit{college}}
& \multicolumn{3}{c}{\textit{quarter}}
& \multicolumn{3}{c}{library}
& \multicolumn{3}{c}{\textit{church}}
& \multicolumn{3}{c}{\textit{palace}}
& \multicolumn{3}{c}{\textit{Average}} \\
\cmidrule(lr){2-4}
\cmidrule(lr){5-7}
\cmidrule(lr){8-10}
\cmidrule(lr){11-13}
\cmidrule(lr){14-16}
\cmidrule(lr){17-19}
& PSNR & SSIM & LPIPS
& PSNR & SSIM & LPIPS
& PSNR & SSIM & LPIPS
& PSNR & SSIM & LPIPS
& PSNR & SSIM & LPIPS
& PSNR & SSIM & LPIPS \\
\midrule

3DGS~\cite{kerbl20233d}
& 24.55 & 0.754 & 0.351
& 18.86 & 0.685 & \cOne 0.402
& 17.72 & 0.689 & \cThree 0.434
& 17.26 & 0.680 & \cThree 0.432
& 19.71 & 0.743 & \cThree 0.404
& 19.62 & 0.710 & \cTwo 0.405 \\

2DGS~\cite{huang20242d}
& 22.82 & 0.731 & 0.430
& 18.55 & 0.710 & 0.456
& 17.09 & 0.696 & 0.492
& 17.11 & 0.708 & 0.548
& 20.09 & \cThree 0.773 & 0.512
& 19.13 & 0.724 & 0.488 \\

PGSR~\cite{chen2024pgsr}
& 22.77 & 0.728 & 0.442
& \cThree 19.56 & \cThree 0.722 & 0.490
& \cTwo 19.49 & \cThree 0.730 & 0.513
& 18.04 & 0.713 & 0.548
& \cThree 20.22 & \cTwo 0.777 & 0.537
& 20.02 & 0.734 & 0.506 \\

MVGS~\cite{du2024mvgs}
& \cTwo 25.08 & 0.767 & 0.376
& \cTwo 19.88 & 0.720 & 0.442
& 18.96 & 0.726 & 0.490
& 18.04 & 0.712 & 0.526
& 19.67 & 0.758 & 0.512
& \cThree 20.33 & \cThree 0.737 & 0.469 \\

GaMeS~\cite{waczynska2024games} $^*$
& 22.39 & 0.730 & 0.485
& 18.46 & 0.681 & 0.492
& -- & -- & --
& 19.31 & 0.686 & 0.558
& 20.21 & 0.734 & 0.612
& 20.09 & 0.708 & 0.537 \\

Triangle-Splat~\cite{held2026triangle}
& 22.52 & 0.701 & 0.393
& 18.49 & 0.650 & \cThree 0.434
& 16.76 & 0.631 & 0.473
& 17.90 & 0.658 & 0.475
& 18.90 & 0.679 & 0.465
& 18.91 & 0.664 & 0.448 \\

Mesh-Splat~\cite{held2025meshsplatting} $^*$
& -- & -- & --
& 17.31 & 0.506 & 0.631
& -- & -- & --
& 16.21 & 0.481 & 0.635
& 16.80 & 0.543 & 0.667
& 16.77 & 0.510 & 0.644 \\

FastGS~\cite{ren2026fastgs}
& 24.37 & \cThree 0.785 & 0.367
& 17.46 & \cTwo 0.731 & 0.466
& 17.36 & 0.695 & 0.463
& 19.69 & \cTwo 0.758 & 0.463
& 19.22 & 0.710 & 0.427
& 19.62 & 0.736 & 0.437 \\

Speedy-Splat~\cite{hanson2025speedy}
& 23.60 & 0.773 & 0.388
& 15.10 & 0.688 & 0.484
& 17.03 & 0.689 & 0.470
& 17.97 & 0.754 & 0.476
& 18.76 & 0.730 & 0.436
& 18.49 & 0.727 & 0.451 \\

Taming-3DGS~\cite{mallick2024taming}
& 24.73 & \cThree 0.785 & \cThree 0.345
& 17.50 & 0.720 & 0.457
& 16.93 & 0.683 & 0.454
& \cThree 19.86 & \cThree 0.757 & 0.451
& 19.55 & 0.703 & \cTwo 0.403
& 19.71 & 0.730 & 0.422 \\

3DGS-LM~\cite{hollein2025_3dgslm}
& 19.70 & 0.704 & 0.377
& 15.11 & 0.687 & 0.486
& 15.87 & 0.672 & 0.470
& 17.75 & 0.745 & 0.467
& 18.67 & 0.711 & 0.425
& 17.42 & 0.704 & 0.445 \\

LM-RS~\cite{pehlivan2026matrixfree}
& 23.28 & \cTwo 0.791 & 0.366
& 17.93 & \cOne 0.746 & 0.473
& \cThree 19.46 & \cTwo 0.747 & \cTwo 0.425
& 17.68 & \cOne 0.778 & 0.452
& \cTwo 21.17 & 0.761 & 0.411
& 19.90 & \cOne 0.765 & 0.425 \\

\midrule

CF-Splat (Adam)
& \cThree 24.86 & 0.761 & \cTwo 0.342
& 18.63 & 0.676 & 0.436
& 19.23 & 0.680 & 0.438
& \cTwo 20.53 & 0.725 & \cOne 0.400
& 19.71 & 0.680 & 0.417
& \cTwo 20.59 & 0.705 & \cThree 0.407 \\

CF-Splat
& \cOne 27.48 & \cOne 0.831 & \cOne 0.316
& \cOne 21.81 & 0.706 & \cTwo 0.403
& \cOne 22.70 & \cOne 0.756 & \cOne 0.388
& \cOne 20.97 & 0.705 & \cTwo 0.410
& \cOne 25.46 & \cOne 0.784 & \cOne 0.333
& \cOne 23.68 & \cTwo 0.756 & \cOne 0.370 \\

\bottomrule
\end{tabular}%
}
\vspace{1.0mm}

\parbox{0.75\textwidth}{\scriptsize
$-$ indicates out-of-memory on GPU. $^*$ means that the average results are calculated over the successfully running scenes only. 
}
\end{table*}

\subsection{Scene Appearance Estimation Results}

\begin{figure*}[!t]
    \centering
    \includegraphics[width=6.2in]{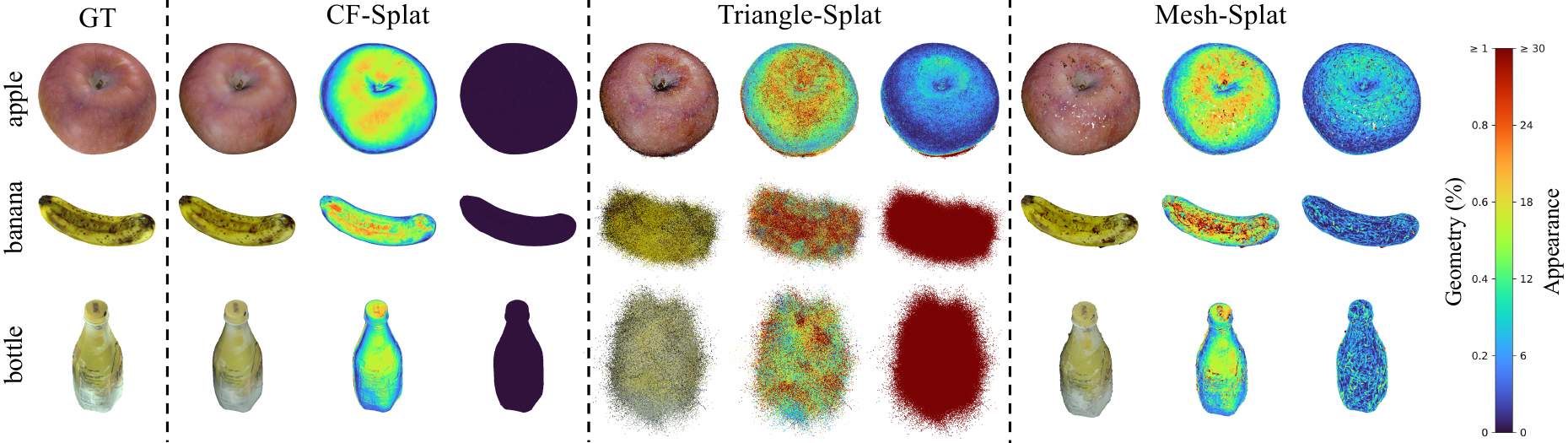}
    \caption{Surface geometry and appearance evaluation on the \textit{apple}, \textit{banana}, and \textit{bottle} scenes from OmniObject3D~\cite{wu2023omniobject3d}. For each method, we show the reconstructed surface, appearance-error map and geometric-error map from left to right. CIEDE2000 reference is attached.}
    \label{fig:omni_mesh_eva}
\end{figure*}

\begin{figure}[!t]
    \centering
    \includegraphics[width=3.5in]{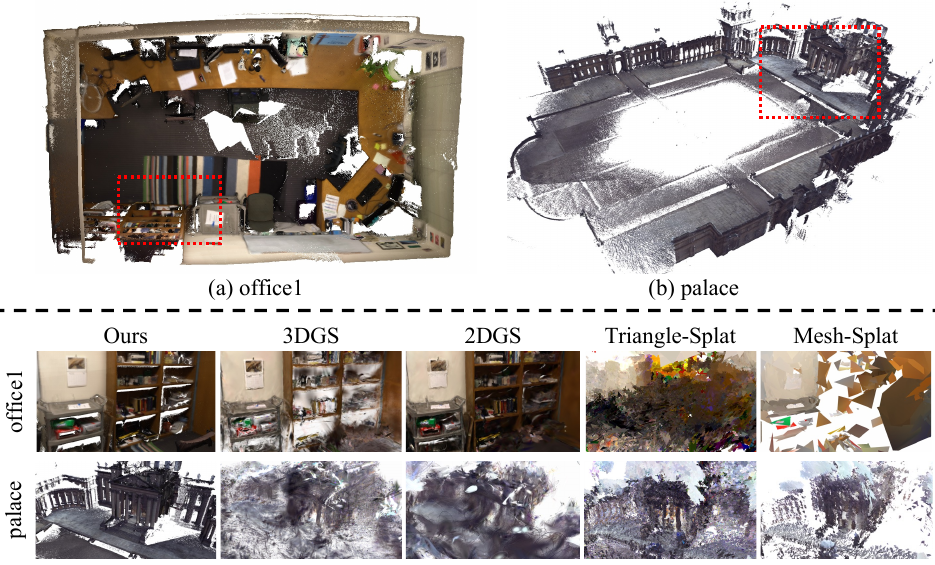}
    \caption{Surface geometry and appearance evaluation on the \textit{office1} scene from BundleFusion~\cite{dai2017bundlefusion} and the \textit{palace}
    scene from Oxford Spires~\cite{tao2026oxford}. The top pictures are the estimated scene of our CF-Splat.}
    \label{fig:oxford_mesh_eva}
\end{figure}

Beyond image-space NVS metrics, we quantitatively evaluate the geometric accuracy and appearance fidelity of the reconstructed surface representations. We restrict this evaluation to triangle-based baselines, since Gaussian-based representations do not explicitly define a surface, and extracting reliable surfaces can be affected by floating Gaussians. We conduct this evaluation on OmniObject3D, which provides high-quality textured meshes as reference geometry and appearance.

For geometric evaluation, we measure the bidirectional mean surface distance between the ground-truth and reconstructed meshes. Specifically, we compute
$
    \mathrm{CD}
    =
    \frac{1}{2}
    (
        F+B
    )
$,
where \(F\) denotes the mean reference-to-prediction surface distance and \(B\) denotes the mean prediction-to-reference surface distance. For appearance evaluation, we reuse the prediction-to-reference surface correspondences established when computing \(B\). For each sampled location on the predicted surface, we compare its color with that at the corresponding location on the ground-truth surface. Both colors are converted to the CIELAB color space, and their appearance discrepancy is quantified by the mean CIEDE2000 color difference~\cite{sharma2005ciede2000}.

Fig.~\ref{fig:omni_mesh_eva} shows that CF-Splat consistently achieves the lowest bidirectional geometric error across all evaluated scenes while also producing lower surface appearance error. In contrast, the compared methods exhibit larger geometric deviations and more spatially irregular error patterns in several local regions, which further lead to noticeable appearance discrepancies around fine structures and high-frequency texture boundaries.

To further visualize the quality of the reconstructed representations, Fig.~\ref{fig:oxford_mesh_eva} presents representative comparisons on BundleFusion and Oxford Spires. CF-Splat produces cleaner surface-attached appearance and better preserves local texture structures, whereas the compared Gaussian- and triangle-based methods exhibit more surface artifacts and texture distortions.

Overall, these advantages stem from our closed-form formulation, which estimates view-dependent appearance directly over the input mesh without iterative geometry optimization. This allows CF-Splat to preserve the underlying surface structure while avoiding geometry drift during appearance fitting, leading to more faithful geometric and appearance reconstruction.

\subsection{Efficiency Analysis}
\label{sec:efficiency}

\begin{figure}[!t]
    \centering
    \includegraphics[width=3.45in]{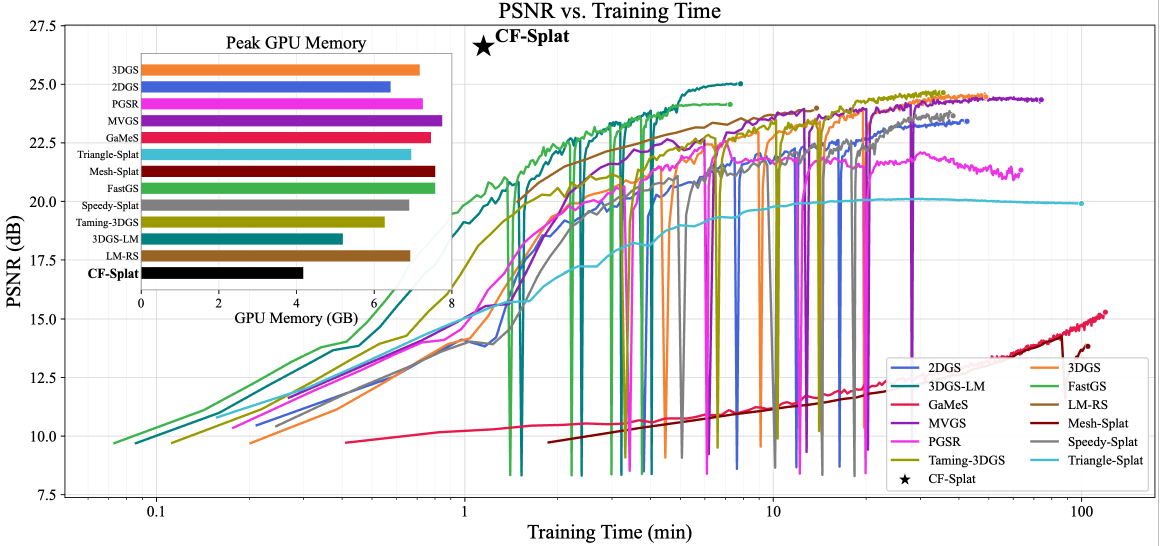}
    \caption{Efficiency comparison on the \textit{college} scene
    from Oxford Spires~\cite{tao2026oxford}. The curves show PSNR as a
    function of computation time for the iterative baselines, while the
     marker denotes the result obtained by CF-Splat. The inset reports the corresponding peak GPU memory consumption.}
    \label{fig:time_psnr}
\end{figure}

\begin{table}[t]
\centering
\caption{
Efficiency comparison across the evaluated datasets.
{\setlength{\fboxsep}{0.5pt}\colorbox{orange!85}{First}},
{\setlength{\fboxsep}{0.5pt}\colorbox{orange!50}{Second}}, and
{\setlength{\fboxsep}{0.5pt}\colorbox{orange!15}{Third}}
indicate the ranking results.
}
\label{tab:efficiency_comparison}

\scriptsize
\setlength{\tabcolsep}{1.5pt}
\renewcommand{\arraystretch}{1.0}

\resizebox{\columnwidth}{!}{%
\begin{tabular}{@{}l *{8}{c}@{}}
\toprule

Method
& \multicolumn{2}{c}{\makebox[2.1cm][c]{C3VD~\cite{bobrow2023}}}
& \multicolumn{2}{c}{\makebox[2.1cm][c]{OmniObject3D~\cite{wu2023omniobject3d}}}
& \multicolumn{2}{c}{\makebox[2.1cm][c]{BundleFusion~\cite{dai2017bundlefusion}}}
& \multicolumn{2}{c}{\makebox[2.1cm][c]{Oxford Spires~\cite{tao2026oxford}}} \\

\cmidrule(lr){2-3}
\cmidrule(lr){4-5}
\cmidrule(lr){6-7}
\cmidrule(lr){8-9}

& Time (min) & GPU (GB)
& Time (min) & GPU (GB)
& Time (min) & GPU (GB)
& Time (min) & GPU (GB) \\

\midrule

3DGS~\cite{kerbl20233d}
& 11.52 & 2.18
& 11.34 & 1.47
& 25.53 & 5.93
& 48.90 & 7.14 \\

2DGS~\cite{huang20242d}
& 14.74 & 2.47
& 14.94 & 1.83
& 41.42 & 6.96
& 42.50 & 6.43 \\

PGSR~\cite{chen2024pgsr}
& 51.32 & 3.06
& 43.17 & 2.24
& 85.19 & 7.07
& 63.58 & 7.26 \\

MVGS~\cite{du2024mvgs}
& 43.99 & 4.28
& 36.69 & 3.07
& 77.69 & 7.43
& 74.00 & 7.76 \\

GaMeS~\cite{waczynska2024games}
& 45.67 & 2.54
& 30.90 & 1.74
& 87.05 & \cTwo 5.08
& 119.40 & 7.47 \\

Triangle-Splat~\cite{held2026triangle}
& 41.07 & 3.08
& 39.67 & 2.06
& 76.90 & 7.12
& 99.85 & 6.96 \\

Mesh-Splat~\cite{held2025meshsplatting}
& 49.60 & 3.46
& 46.82 & 2.43
& 64.62 & 7.46
& 104.81 & 7.58 \\

FastGS~\cite{ren2026fastgs}
& \cTwo 2.94 & \cThree 2.04
& \cTwo 2.53 & \cThree 1.28
& \cTwo 3.26 & \cTwo 5.08
& \cTwo 7.25 & 6.58 \\

Speedy-Splat~\cite{hanson2025speedy}
& 7.93 & 2.40
& 7.54 & 2.65
& \cThree 18.02 & 6.31
& 38.30 & 6.91 \\

Taming-3DGS~\cite{mallick2024taming}
& 8.41 & 2.39
& \cThree 5.60 & 1.47
& 26.16 & 5.81
& \cThree 35.58 & \cThree 6.28 \\

3DGS-LM~\cite{hollein2025_3dgslm}
& 9.36 & 4.66
& 8.95 & 2.74
& 28.77 & 6.21
& \cThree 40.72 & \cThree 7.55 \\

LM-RS~\cite{pehlivan2026matrixfree}
& 7.62 & 2.14
& 6.12 & 1.88
& 14.25 & 3.63
& 16.61 & 6.61 \\

\midrule

CF-Splat (Adam)
& \cThree 5.96 & \cOne 0.47
& 12.24 & \cOne 0.61
& 22.90 & \cThree 4.14
& 44.45 & \cTwo 4.72 \\

CF-Splat
& \cOne 0.09 & \cTwo 0.63
& \cOne 0.03 & \cTwo 0.88
& \cOne 0.93 & \cOne 3.62
& \cOne 1.55 & \cOne 4.18 \\

\bottomrule
\end{tabular}%
}
\end{table}

Fig.~\ref{fig:time_psnr} and Table~\ref{tab:efficiency_comparison}
evaluate the computational efficiency of CF-Splat in terms of computation
time and peak GPU memory consumption. Fig.~\ref{fig:time_psnr} illustrates the
accuracy--time behavior on the large-scale \textit{college} scene, while
Table~\ref{tab:efficiency_comparison} summarizes the efficiency results
across all four datasets. CF-Splat achieves substantially shorter
computation time than the compared methods while maintaining a smaller peak GPU
memory consumption across scene scales.

\subsubsection{Computation Time}

CF-Splat achieves the shortest computation time across all four datasets. Compared with 3DGS, it provides a speedup of approximately $27.5\times$--$378.0\times$. Meanwhile, it only requires 1.55 min on large-scale Oxford Spires.
More notably, CF-Splat is approximately $24.6\times$--$408.0\times$ faster than CF-Splat (Adam), demonstrating that the efficiency gain is not merely due to the triangle-based representation, but primarily results from the proposed closed-form formulation.

As illustrated in Fig.~\ref{fig:time_psnr}, iterative splatting methods progressively improve reconstruction quality through repeated rendering, gradient back-propagation, and parameter updates. In contrast, CF-Splat exploits the linear dependence of image observations on the per-vertex SH coefficients and formulates appearance estimation as a global LLS formulation in \eqref{eq:global_lls}. The advantage becomes increasingly pronounced for large-scale Oxford Spires, where repeatedly processing high-resolution multi-view observations dominates the computational cost of iterative methods.

\subsubsection{Peak GPU Memory Consumption}

CF-Splat also maintains low peak GPU memory consumption across all evaluated datasets, achieving a $1.6\times$--$3.5\times$ reduction relative to 3DGS. Even on the large-scale Oxford Spires dataset, CF-Splat requires only $4.18$ GB of peak GPU memory.

The memory advantage originates from two complementary aspects. First, CF-Splat estimates only the per-vertex SH appearance coefficients over the given mesh. Unlike gradient-based splatting methods, it does not require maintaining a large collection of independently optimized primitive parameters together with the intermediate gradients and optimizer states. Second, the matrix-free solver introduced in Section~\ref{sec:matrix_free} applies the normal operator implicitly without explicitly materializing the global normal matrix in \eqref{eq:normal_system}. 

Overall, these results highlight a favorable scaling behavior of CF-Splat. Increasing the number of observations primarily introduces additional linear measurements rather than additional trainable scene parameters or optimization states. Combined with the matrix-free solution strategy, this allows CF-Splat to retain both low computation time and moderate GPU memory consumption as the scene scale increases. More detailed scalability analyses are presented in Section~\ref{sec:scalability}.

\subsection{Analysis of matrix-free PCG Solver}

\begin{figure}[!t]
    \centering
    \includegraphics[width=3.5in]{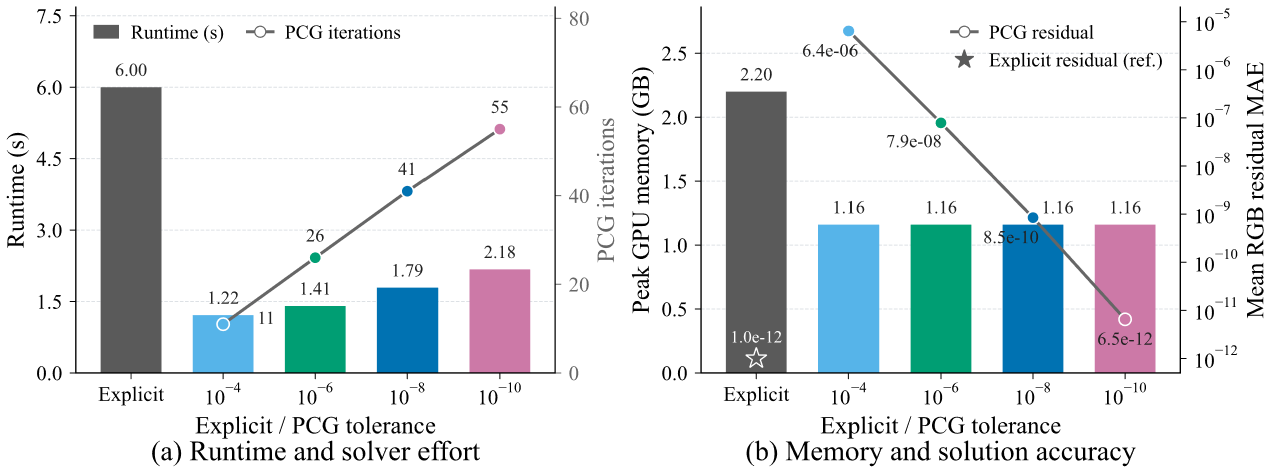}
    \caption{Numerical accuracy and efficiency of matrix-free PCG under
    different stopping tolerances on the \textit{backpack} scene from
    OmniObject3D~\cite{wu2023omniobject3d}, with the explicitly assembled
    solver used as the numerical reference. Explicit means solving the same global LLS formulation through a factorization-based solver.
    (a) Runtime and PCG iteration count.
    (b) GPU memory and mean RGB residual MAE.}
    \label{fig:explicit_pcg}
\end{figure}

We evaluate the numerical accuracy of the matrix-free PCG solver by comparing
it with an explicitly assembled solution of the same global normal system.
As established in Section~\ref{sec:matrix_free}, the matrix-free operator
preserves the normal system exactly; therefore, the PCG stopping tolerance
controls the numerical approximation to the same LLS solution.

As shown in Fig.~\ref{fig:explicit_pcg}, tightening the stopping tolerance
requires more PCG iterations and computation, but progressively reduces the
final residual error. At the tighter tolerances, the residual MAE approaches
that of the factorization-based solver, demonstrating that matrix-free PCG can recover
the same LLS solution to high numerical accuracy. This behavior agrees with
\eqref{eq:pcg_stopping}, where a smaller residual provides a tighter
bound on the error relative to the unique global minimizer. Meanwhile, matrix-free PCG maintains a smaller GPU memory because the normal matrices are not explicitly constructed.

\subsection{Analysis of Scalability on Large-Scale Scenes}
\label{sec:scalability}
In this section, we evaluate the scalability of CF-Splat with respect to the number of input views and mesh vertices, on the large-scale \textit{college} scene from  Oxford Spires. For view scalability, we keep the input mesh at 1.55M vertices, while for vertex scalability, we use 760 input views across all settings. To isolate the computational scalability of each method, densification function is disabled for both CF-Splat and the baselines throughout these experiments.

\subsubsection{Scalability to The Number of Input Views}

\begin{figure*}[!t]
    \centering
    \includegraphics[width=7.1in]{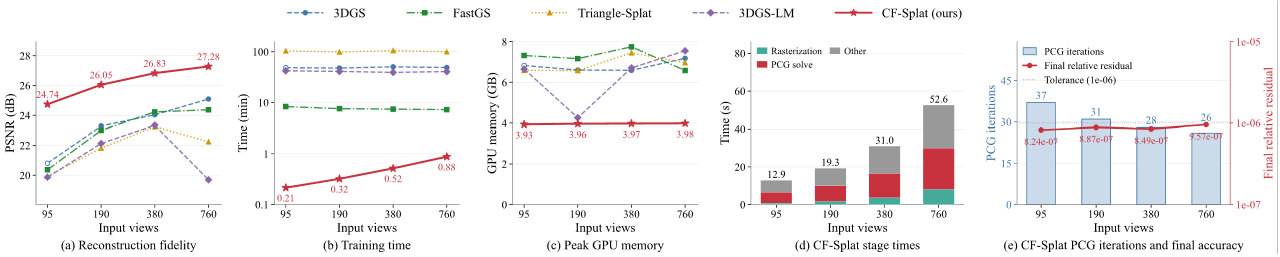}
    \caption{Scalability of CF-Splat (w/o densification) with respect to the number of input views on the \textit{college} scene from Oxford Spires~\cite{tao2026oxford}, with 1.55M input mesh vertices. (a--c) Reconstruction quality, runtime, and peak GPU memory compared with baseline methods. (d) Runtime breakdown of CF-Splat. (e) PCG iteration counts and final relative residuals, with the dashed line indicating the convergence tolerance.}
    \label{fig:view_scalability}
\end{figure*}

Fig.~\ref{fig:view_scalability} evaluates the scalability of CF-Splat with respect to the number of input views. As shown in Fig.~\ref{fig:view_scalability}(a), the PSNR of CF-Splat steadily improves as more input views become available and remains consistently higher than those of the compared baselines. This trend is consistent with the directional support underlying our global SH appearance estimation. Additional cameras introduce new viewing directions for the per-vertex SH coefficients, enabling more SH bands to satisfy the support criterion in \eqref{eq:sh_dimensional_condition} and receive weaker regularization through \eqref{eq:adaptive_sh_regularizer}. Although 3DGS and FastGS also benefit substantially from additional views, CF-Splat achieves the highest PSNR across all settings, while Triangle-Splat and 3DGS-LM exhibit non-monotonic behavior at larger input scales. Unlike CF-Splat, the baselines do not explicitly adapt SH complexity to the available directional support, but instead incorporate additional observations through iterative scene optimization, which may contribute to their less stable fidelity gains as the number of views increases.

Fig.~\ref{fig:view_scalability}(b), (d), and (e) jointly demonstrate the
computational scalability of CF-Splat. CF-Splat maintains substantially
lower overall runtime than the baselines by solving the global LLS system,
rather than repeatedly optimizing a large set of scene parameters. As the
number of input views increases, the solving time grows with the number of
observations, consistent with \eqref{eq:solver_time_complexity}, while the
global LLS solution remains efficiently obtained across the evaluated
settings.

Fig.~\ref{fig:view_scalability}(c) further shows the memory scalability of CF-Splat. The matrix-free formulation avoids explicitly storing \(\mathbf{A}\) and \(\mathbf{H}\) in \eqref{eq:normal_system}, while view-wise streaming retains only the observation features of the current view. As indicated by \eqref{eq:solver_memory_complexity}, the working memory is therefore governed primarily by the number of vertices \(N_v\), rather than the number of input views \(K\). Consequently, increasing \(K\) leads to only minor memory variation when the mesh resolution is unchanged.

\subsubsection{Scalability to The Number of Input Mesh Vertices}

\begin{figure}[!t]
    \centering
    \includegraphics[width=3.5in]{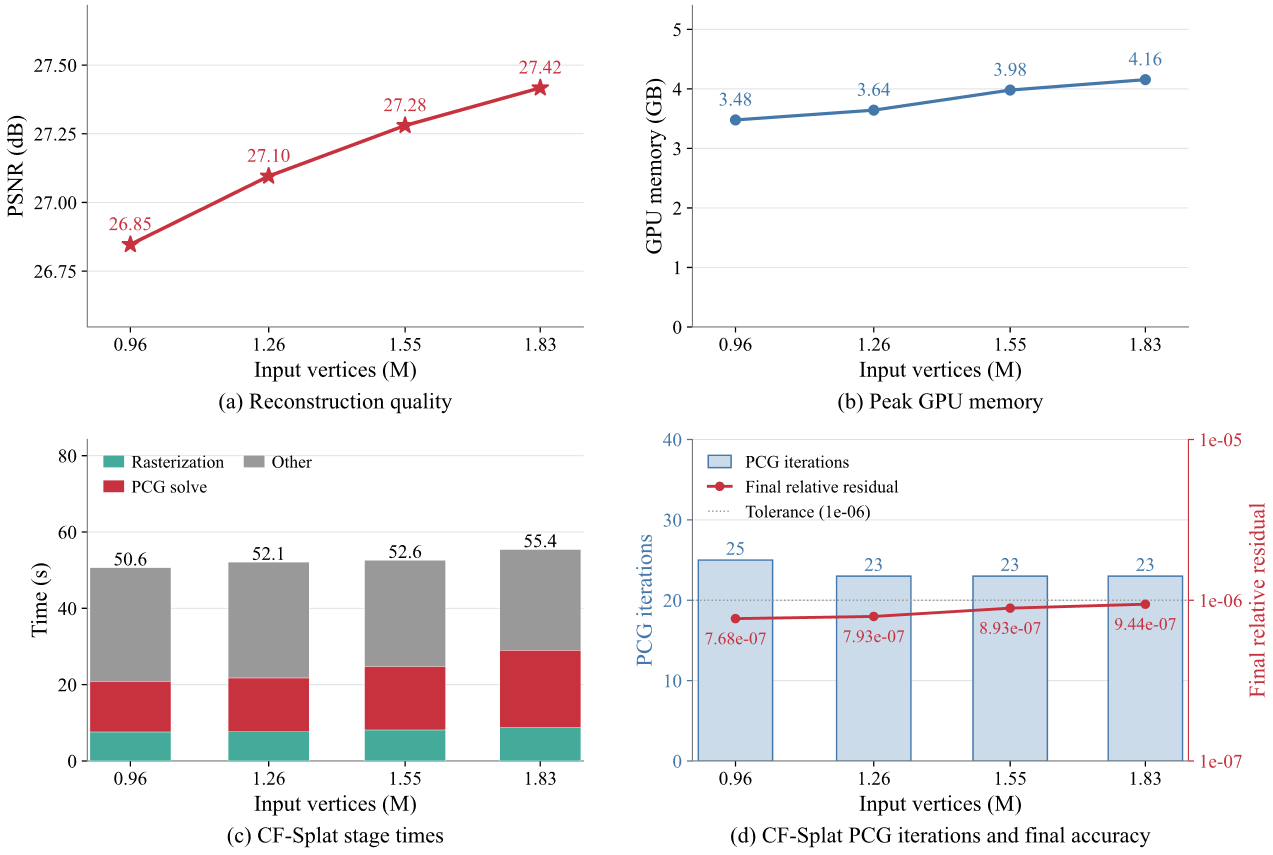}
    \caption{Scalability of CF-Splat (w/o densification) with respect to the number of input mesh vertices on the \textit{college} scene from Oxford Spires~\cite{tao2026oxford}, with 760 input views. (a) Reconstruction quality. (b) Peak GPU memory. (c) Runtime breakdown of CF-Splat. (d) PCG iteration counts and final relative residuals, with the dashed line indicating the convergence tolerance.}
    \label{fig:vertices_scalability}
\end{figure}

Fig.~\ref{fig:vertices_scalability} evaluates the scalability of CF-Splat with respect to the number of input mesh vertices. As shown in Fig.~\ref{fig:vertices_scalability}(a), PSNR steadily improves as the mesh resolution increases. We do not directly compare CF-Splat with other splatting-based methods under a controlled primitive count, as their NVS performance relies heavily on adaptive densification, making it difficult to enforce a comparable number of primitives without altering their native optimization behavior.

Fig.~\ref{fig:vertices_scalability}(b), (c), and (d) show that this
improvement is achieved with moderate computational growth. Peak GPU memory
increases gradually with $N_v$, since the global LLS formulation optimizes
only the per-vertex SH coefficients, while its matrix-free solver
processes the observations view-wise without explicitly storing the global
observation or normal matrices. This behavior is consistent with
\eqref{eq:solver_memory_complexity}. The overall runtime also grows
moderately with $N_v$, mainly due to the increased vertex-wise workload
described in \eqref{eq:solver_time_complexity}. These results demonstrate
that CF-Splat maintains better computational and memory scalability as
the mesh resolution increases.

\subsection{Analysis of Adaptive SH Regularization}

\begin{figure}[!t]
    \centering
    \includegraphics[width=3.5in]{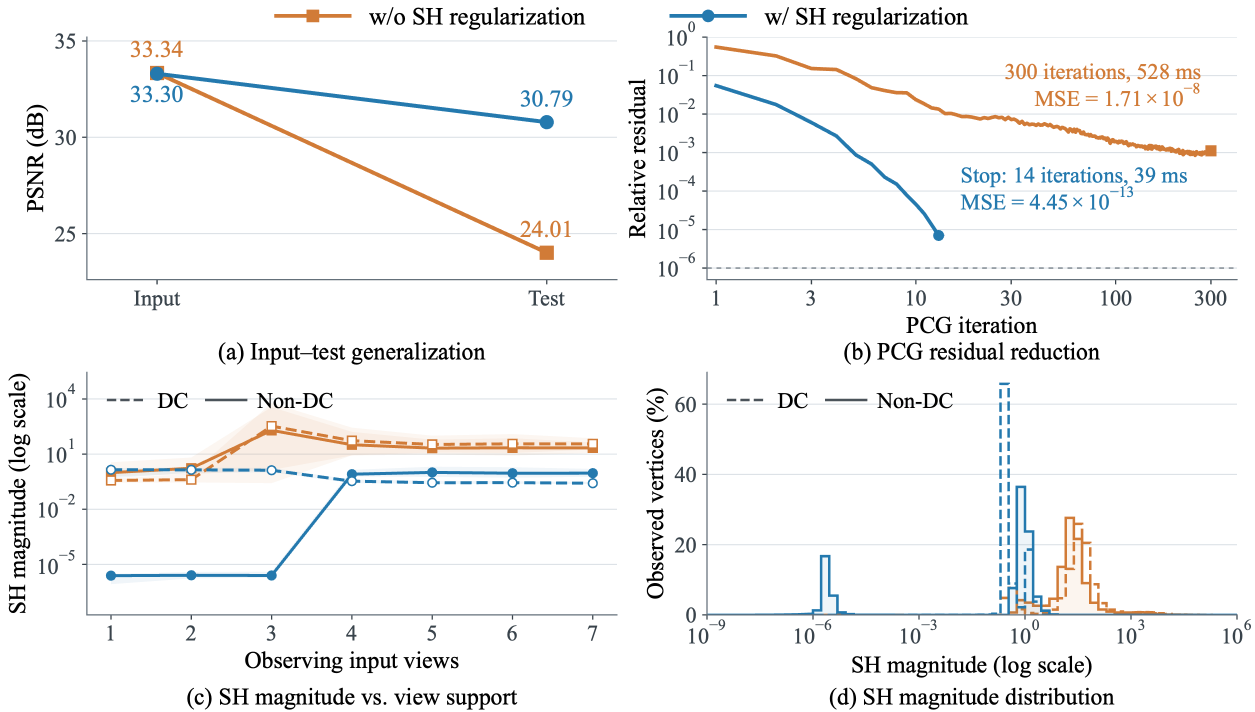}
    \captionof{figure}{Effect of adaptive SH regularization under uneven observations, on the \textit{c\_t1\_b} scene from C3VD~\cite{bobrow2023}, using $K=7$ input views and maximum SH degree \(L=3\). (a) PSNR on input and test views. (b) PCG solving records, with solve time, iteration count, and final residual MSE. (c) Per-vertex SH magnitudes over the number of observing input views. (d) SH magnitude distributions over observed vertices. Dashed and solid lines in (c) and (d) denote the DC (\(l=0\)) and non-DC (\(l\geq1\)) SH components, respectively.}
    \label{fig:adaptive_sh1}
\end{figure}

This section evaluates the role of the adaptive SH regularization in Section~\ref{sec:adaptive_sh} under uneven observations on the \textit{c\_t1\_b} scene from C3VD. We set the maximum SH degree \(L=3\) and use $K=7$ input views, resulting in a sparse-view setting with limited directional support for many vertices.

\begin{figure}[!t]
    \centering
    \includegraphics[width=3.4in]{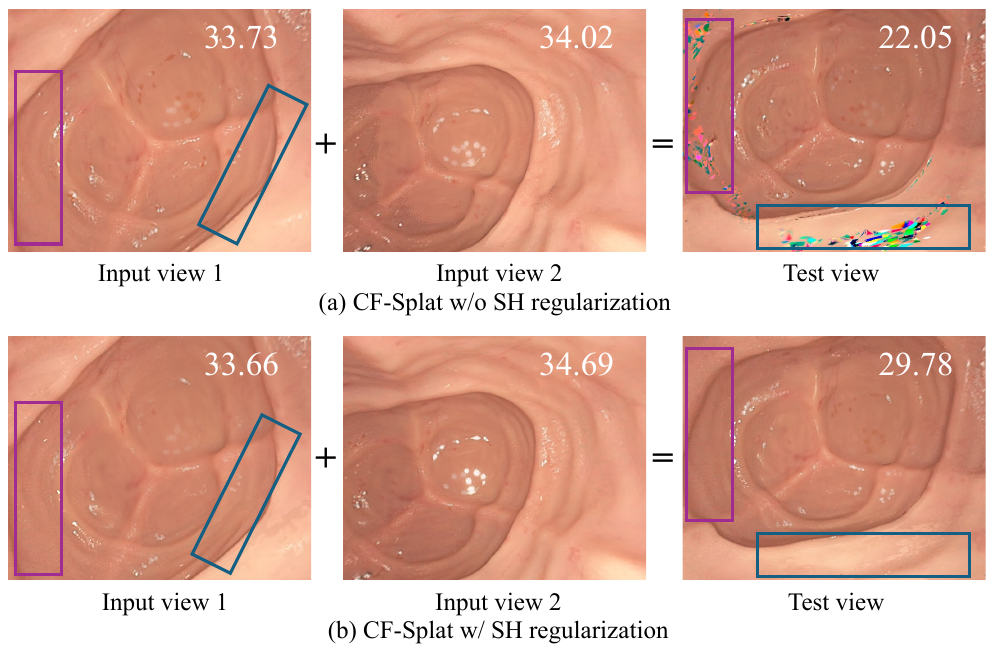}
    \captionof{figure}{Qualitative effect of adaptive SH regularization in Fig.~\ref{fig:adaptive_sh1}. The regions highlighted by the \textcolor{purple}{purple} and \textcolor{blue}{blue} boxes are observed in input view 1 but not in input view 2, resulting in insufficient directional support for higher-order SH components according to \eqref{eq:sh_dimensional_condition}. Without adaptive SH regularization, the weakly constrained SH components produce noisy colors in the test view.}
    \label{fig:adaptive_sh2}
\end{figure}

Fig.~\ref{fig:adaptive_sh1} quantitatively evaluates the effect of adaptive SH regularization under sparse and uneven observations, while Fig.~\ref{fig:adaptive_sh2} provides a visualization of the corresponding case.

As shown in Fig.~\ref{fig:adaptive_sh1}(a), both configurations fit the input views similarly, whereas adaptive regularization substantially improves the test-view PSNR. This indicates that, without sufficient directional constraints, unconstrained SH coefficients that do not satisfy the condition in \eqref{eq:sh_dimensional_condition} can overfit the observed views and generalize poorly to unseen viewpoints.

Fig.~\ref{fig:adaptive_sh1}(b) further shows the numerical effect of the
adaptive SH regularization. With regularization, the PCG residual decreases
more rapidly, while the unregularized system remains difficult to solve within
the maximum iteration count $T_{\max}=300$. This observation is consistent
with the role of the proposed regularization in constraining weakly supported
SH components and improving the conditioning of the appearance estimation
problem.

Fig.~\ref{fig:adaptive_sh1}(c) and (d) further illustrate how the proposed regularization controls the DC (zeroth-order, $l=0$) and non-DC (higher-order, $l>0$) components of the SH appearance vector. For vertices observed from only a few views, the non-DC components are strongly suppressed while the DC component is preserved. As view support increases, more directional components are retained according to \eqref{eq:adaptive_sh_regularizer}. The resulting coefficient distribution also exhibits substantially smaller large-magnitude parts, reducing unstable view-dependent appearance.

\subsection{Robustness to Input Noise}
Since CF-Splat builds upon a given mesh and calibrated observations, we further evaluate its robustness to camera pose perturbations and geometric inaccuracies in the input mesh on the \textit{apt0} scene from BundleFusion. Camera pose perturbations are applied to both the input and test views.

\subsubsection{Pose Noise}

\begin{table}[t]
    \centering
    \setlength{\tabcolsep}{3.0pt}
    \renewcommand{\arraystretch}{1.05}
    
    \caption{Effect of pose noise on the \textit{Apt0} scene from
    BundleFusion~\cite{dai2017bundlefusion}. Rotation and translation
    noise denote per-axis standard deviations.}
    \label{tab:pose_noise}

    \begin{tabular}{@{}c@{\hspace{8pt}}l@{\hspace{10pt}}c@{\hspace{10pt}}c@{\hspace{10pt}}c@{}}
        \toprule
        \shortstack{Rot./Trans.\\($^\circ$/cm)}
        & Method
        & PSNR $\uparrow$
        & SSIM $\uparrow$
        & LPIPS $\downarrow$ \\
        \midrule

        \multirow{2}{*}{$0/0$}
        & 3DGS     & 28.08 & 0.852 & 0.253 \\
        & CF-Splat & 29.16 & 0.854 & 0.232 \\
        \addlinespace[2pt]

        \multirow{2}{*}{$0.1/0.5$}
        & 3DGS     & 26.56 & 0.815 & 0.293 \\
        & CF-Splat & 28.07 & 0.823 & 0.267 \\
        \addlinespace[2pt]

        \multirow{2}{*}{$0.3/2$}
        & 3DGS     & 23.00 & 0.751 & 0.378 \\
        & CF-Splat & 25.08 & 0.759 & 0.333 \\
        \addlinespace[2pt]

        \multirow{2}{*}{$1.0/5$}
        & 3DGS     & 19.62 & 0.714 & 0.449 \\
        & CF-Splat & 21.91 & 0.687 & 0.397 \\
        \bottomrule
    \end{tabular}
\end{table}

\begin{figure}[!t]
    \centering
    \includegraphics[width=3.5in]{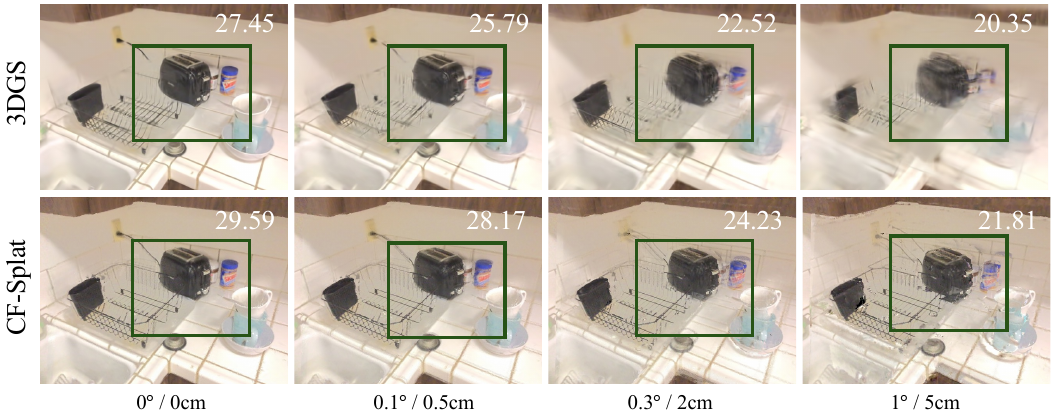}
    \caption{Visualized comparison of 3DGS and CF-Splat under increasing camera pose perturbations, on the \textit{apt0} scene from BundleFusion~\cite{dai2017bundlefusion}. PSNR of visualized test view is reported for each noise setting.}
    \label{fig:pose_noise}
\end{figure}

We first evaluate robustness to camera pose perturbations. Starting from the ground-truth poses for this ablation only, we apply four levels of rotation and translation noise with per-axis standard deviations of
$0^\circ/0$ cm,
$0.1^\circ/0.5$ cm,
$0.3^\circ/2$ cm, and
$1^\circ/5$ cm.
We compare CF-Splat with 3DGS under the same perturbation settings, with quantitative and qualitative results reported in Table~\ref{tab:pose_noise} and Fig.~\ref{fig:pose_noise}, respectively. Compared with 3DGS, CF-Splat maintains better reconstruction quality than 3DGS as pose noise increases. Under inaccurate camera poses, the per-view iterative updates in 3DGS are driven by inconsistent photometric constraints, which can be absorbed by the coupled Gaussian geometry and appearance parameters and lead to blurred renderings. In contrast, CF-Splat jointly incorporates all observations into a global closed-form formulation on the given mesh, reducing such view-specific overfitting and yielding more stable appearance estimates under pose perturbations.

\subsubsection{Mesh Noise}
\begin{table}[t]
    \centering
    \setlength{\tabcolsep}{3.0pt}
    \caption{Effect of Mesh Reconstruction Errors Induced by Different TSDF voxel sizes on the \textit{apt0} scene from BundleFusion~\cite{dai2017bundlefusion}. Best Results Are Highlighted in Bold.}
    \label{tab:mesh_noise}
    \begin{tabular}{@{}lccccc@{}}
        \toprule
        Voxel size & Vertices (M) & PSNR $\uparrow$ & SSIM $\uparrow$ & LPIPS $\downarrow$ & Error (cm) $\downarrow$ \\
        \midrule
        0.5 cm & 11.92 & \textbf{29.17} & \textbf{0.872} & \textbf{0.212} & \textbf{8.29} \\
        1 cm & 2.39 & 29.10 & 0.869 & 0.219 & 8.98 \\
        2 cm & 0.48 & 29.07 & 0.867 & 0.222 & 9.86 \\
        4 cm & 0.10 & 28.51 & 0.863 & 0.223 & 11.81 \\
        \bottomrule
    \end{tabular}
\end{table}

\begin{figure}[!t]
    \centering
    \includegraphics[width=3.5in]{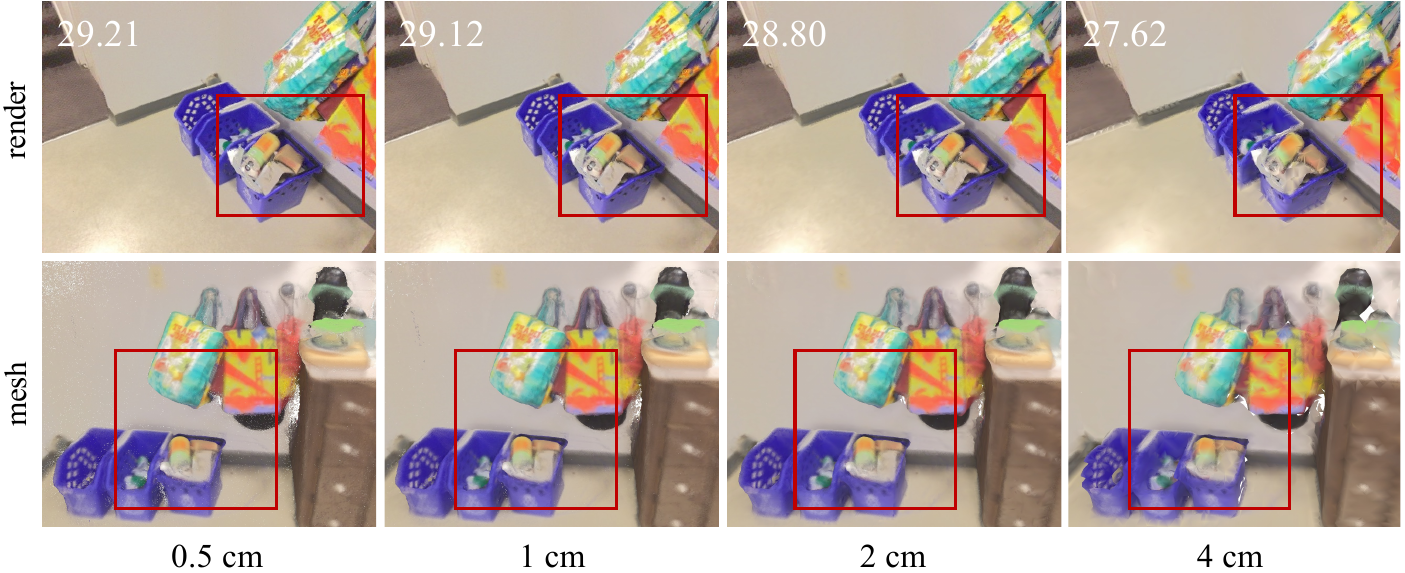}
    \caption{Render (top) and mesh (bottom) results of CF-Splat using input meshes reconstructed with different TSDF voxel sizes. PSNR is reported for each setting.}
    \label{fig:mesh_noise}
\end{figure}

To evaluate robustness to inaccuracies in the input geometry, we reconstruct the input meshes using TSDF voxel sizes of $0.5$ cm, $1$ cm, $2$ cm, and $4$ cm. We quantify the geometric error using the mean one-way surface distance from the reference mesh to the TSDF-reconstructed mesh, where smaller voxel sizes lead to higher geometric accuracy. Table~\ref{tab:mesh_noise} reports the corresponding vertex counts, geometric errors, and rendering performance, while Fig.~\ref{fig:mesh_noise} provides qualitative comparisons of the rendered results and reconstructed mesh regions.

Despite the increasing geometric error, CF-Splat exhibits only moderate degradation in rendering quality. The results indicate that the proposed closed-form appearance estimation remains stable across meshes with different geometric accuracies, demonstrating robustness to errors of the input mesh.

\subsection{Ablation Study}
\label{sec:ablation}
We further ablate four design choices of CF-Splat: the maximum SH degree,
gradient-guided triangle densification in Section~\ref{sec:densification},
the Block-Jacobi preconditioner in Section~\ref{sec:pcg}, and the global
closed-form formulation in Section~\ref{sec:global_lls}.

\subsubsection{SH Degree}

\begin{table}[t]
  \centering
  
  \setlength{\tabcolsep}{3.0pt}
  \caption{Ablation of the maximum SH degree $L$ on the
  \textit{backpack} scene from OmniObject3D~\cite{wu2023omniobject3d}.}
  \label{tab:sh_degree_ablation}
  \resizebox{\columnwidth}{!}{%
  \begin{tabular}{ccccccc}
    \toprule
    SH Degree
    & PSNR$\uparrow$
    & SSIM$\uparrow$
    & LPIPS$\downarrow$
    & Time (s)$\downarrow$
    & GPU (GB)$\downarrow$
    & FPS$\uparrow$ \\
    \midrule
    0 & 35.39 & 0.968 & 0.062
      & \textbf{0.96} & \textbf{1.153} & \textbf{3027.6} \\
    1 & 38.46 & 0.979 & 0.054
      & 0.98 & 1.154 & 3018.0 \\
    2 & 38.63 & 0.982 & 0.043
      & 1.19 & 1.157 & 2874.4 \\
    3 & \textbf{39.61} & \textbf{0.989} & \textbf{0.025}
      & 1.53 & 1.160 & 2853.3 \\
    \bottomrule
  \end{tabular}%
  }
\end{table}

We first evaluate the maximum SH degree $L$ on the \textit{backpack} scene. As shown in Table~\ref{tab:sh_degree_ablation}, increasing $L$
generally improves NVS quality by expanding the directional
appearance capacity of the per-vertex SH representation in
\eqref{eq:sh}. The improvement comes with an increase in computation
and memory, since the SH state dimension $q$ also grows.

\subsubsection{Gradient-Guided Adaptive Densification}

\begin{table}[t]
  \centering
  \setlength{\tabcolsep}{3.0pt}
  \caption{Ablation of Gradient-Guided Triangle Densification on The
  \textit{backpack} Scene from OmniObject3D~\cite{wu2023omniobject3d}.}
  \label{tab:dense_round_ablation}
  \resizebox{\columnwidth}{!}{%
  \begin{tabular}{cccccccc}
    \toprule
    Rounds
    & Vertices (K)
    & PSNR$\uparrow$
    & SSIM$\uparrow$
    & LPIPS$\downarrow$
    & Time (s)$\downarrow$
    & GPU (GB)$\downarrow$
    & FPS$\uparrow$ \\
    \midrule
     0 &  22.4 & 39.31 & 0.986 & 0.027
       & \textbf{0.81} & \textbf{1.146} & \textbf{3959.9} \\
    10 &  51.4 & 39.61 & 0.989 & 0.025
       & 1.53 & 1.160 & 2853.3 \\
    20 &  77.2 & 39.66 & 0.989 & 0.023
       & 2.43 & 1.174 & 2159.6 \\
    40 & 125.9 & \textbf{39.67} & \textbf{0.990} & \textbf{0.023}
       & 4.28 & 1.192 & 1581.8 \\
    \bottomrule
  \end{tabular}%
  }
\end{table}

\begin{figure}[!t]
    \centering
    \includegraphics[width=3.2in]{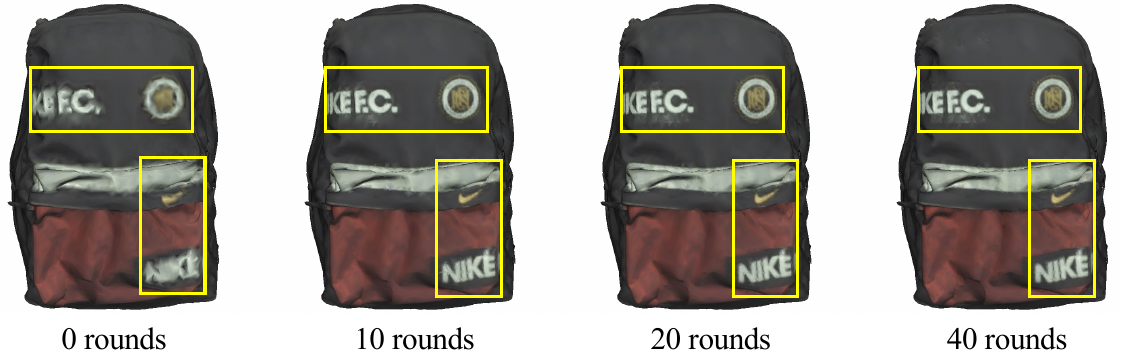}
    \caption{Qualitative effect of gradient-guided adaptive densification
    on the \textit{backpack} scene from
    OmniObject3D~\cite{wu2023omniobject3d}. The highlighted regions show
    the appearance details.}
    \label{fig:densification}
\end{figure}

We next evaluate gradient-guided adaptive densification on the \textit{backpack} scene, varying only the number of densification rounds. Table~\ref{tab:dense_round_ablation}
and Fig.~\ref{fig:densification} show that increasing spatial vertex support
improves fine-scale appearance reconstruction, with the gains gradually
saturating as the mesh becomes denser.

\subsubsection{Block-Jacobi Preconditioner}

\begin{figure}[!t]
    \centering
    \includegraphics[width=3.5in]{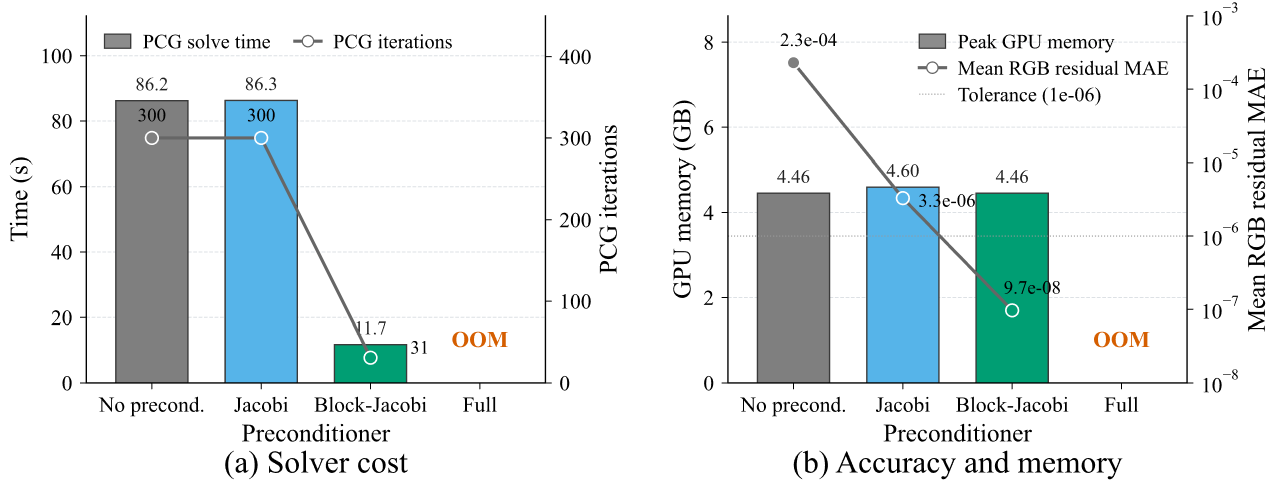}
    \caption{Preconditioner ablation on the \textit{apt0} scene from
    BundleFusion~\cite{dai2017bundlefusion}, using a PCG tolerance of
    $10^{-6}$ and a maximum of 300 iterations. No precond. means the proposed PCG solver without any preconditioner.
    (a) PCG solve time and iteration count.
    (b) Peak GPU memory and mean RGB residual MAE.
    OOM exceeds the available GPU memory.}
    \label{fig:jacobi_comparison}
\end{figure}

We compare the matrix-free solver without preconditioning, with scalar
Jacobi, with the proposed Block-Jacobi preconditioner, and with a full dense
preconditioner on the \textit{apt0} scene.

As shown in Fig.~\ref{fig:jacobi_comparison}, the unpreconditioned and scalar
Jacobi variants require substantially more iterations and fail to satisfy the
prescribed tolerance within the common maximum iterations. In contrast,
Block-Jacobi converges more efficiently while maintaining a comparable memory
footprint.

\subsubsection{Global Closed-Form Formulation}

\begin{figure}[!t]
    \centering
    \includegraphics[width=3.5in]{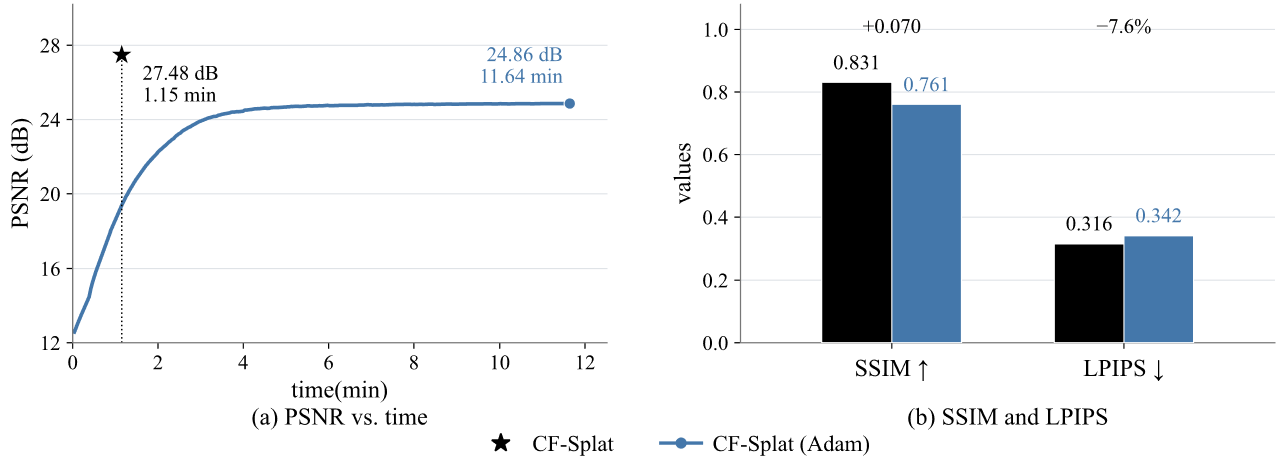}
    \caption{Comparison between CF-Splat and CF-Splat (Adam) on the \textit{college} scene from Oxford Spires~\cite{tao2026oxford}.
    (a) PSNR vs. computation time.
    (b) Final SSIM and LPIPS.}
    \label{fig:cf_adam}
\end{figure}

Finally, we isolate the benefit of the global closed-form
formulation by comparing CF-Splat with CF-Splat (Adam). The only difference between them is the optimization strategy, as introduced in Section~\ref{sec:baselines}.

Fig.~\ref{fig:cf_adam} shows that the Adam variant improves progressively
with optimization but converges much more slowly, while CF-Splat reaches a
better appearance estimate with substantially less computation. CF-Splat exploits the global LLS objective directly rather than relying on
a finite sequence of learning-rate-dependent gradient updates.

\subsection{Discussion}
Similar to triangle-based splatting methods~\cite{held2026triangle,held2025meshsplatting}, CF-Splat solves a slightly different problem comparing to existing Gaussian splatting methods in its optimization formulation. Although all methods share the same high-level objective of estimating scene appearance for NVS, CF-Splat exploits the linear structure of the appearance model and formulates the problem as a global LLS system over the input mesh, rather than optimizing an $\alpha$-blending-based nonlinear objective. Therefore, the global optimality guaranteed by CF-Splat is specific to our global LLS formulation and should not be interpreted as global optimality over all possible splatting objectives. This also explains why our goal is not to dominate every baseline in image-space NVS quality, but to achieve competitive or superior rendering quality under a substantially more structured optimization formulation.

Another important distinction is that CF-Splat preserves the geometry of the input mesh and focuses on appearance estimation over this surface representation. Apart from appearance-driven densification, the underlying geometry is not jointly optimized during scene fitting. This deliberately simplifies the splatting problem, but also avoids degrading geometry solely for marginal improvements in image-space rendering quality. Such a property is particularly valuable in robotic pipelines, where the reconstructed geometry may also serve downstream tasks and therefore should remain consistent with the output of upstream mapping or reconstruction modules.

Finally, the use of matrix-free PCG does not conflict with the closed-form nature of CF-Splat. The underlying appearance estimation problem remains a well-posed global LLS problem with a unique global minimizer, while the proposed PCG solver serves only as an efficient numerical procedure for solving the resulting LLS system at scale. In this sense, the closed-form formulation defines the optimization structure, whereas the matrix-free solver provides a practically stable implementation of that structure.

% \subsection{Discussion}

\section{Conclusion}
\label{sec:conclusion}

In this work, we presented CF-Splat, a novel closed-form triangle-splatting
framework for efficient scene appearance estimation over a
known triangular mesh. Given the mesh geometry and calibrated cameras,
CF-Splat reveals that rasterized pixel colors are linear with respect to the
per-vertex SH coefficients, allowing multi-view appearance estimation to be
formulated as a global LLS problem with a unique global
minimizer. Gradient-guided triangle densification adaptively
allocates spatial appearance capacity without altering the mapped surface,
while observation-adaptive SH regularization improves robustness under uneven
directional support. To scale the resulting global system, the proposed triangle-structured matrix-free PCG solver exploits triangle-local observation
support and avoids explicitly constructing the global normal system while
converging to the same LLS solution within the prescribed numerical
tolerance. Experiments across four real-world
datasets demonstrate that CF-Splat achieves competitive or superior NVS quality while substantially reducing computation time and peak GPU memory consumption on a consumer-grade laptop. 
These results show that CF-Splat provides a new closed-form perspective on splatting-based scene representation, demonstrating that high-quality appearance estimation can be achieved through a globally structured linear formulation rather than iterative nonlinear scene fitting.

The current formulation assumes relatively accurate mesh geometry and
camera calibration, so the errors in the upstream
map are not intentionally corrected. Future work will therefore investigate uncertainty-aware appearance estimation under imperfect geometry and poses, as well as incremental
formulations that update the global appearance state as new observations
become available.

\small
\bibliographystyle{ieeetr}
\bibliography{ref}

@article{seeley1966spherical,
  title={Spherical harmonics},
  author={Seeley, Robert T},
  journal={The American Mathematical Monthly},
  volume={73},
  number={4P2},
  pages={115--121},
  year={1966},
  publisher={Taylor \& Francis}
}

@book{muller2006spherical,
  title={Spherical harmonics},
  author={M{\"u}ller, Claus},
  volume={17},
  year={2006},
  publisher={Springer}
}

@article{kerbl20233d,
  title={3D Gaussian splatting for real-time radiance field rendering.},
  author={Kerbl, Bernhard and Kopanas, Georgios and Leimk{\"u}hler, Thomas and Drettakis, George},
  journal={ACM Trans. Graph.},
  volume={42},
  number={4},
  pages={139--1},
  year={2023}
}

@inproceedings{huang20242d,
  title={2d gaussian splatting for geometrically accurate radiance fields},
  author={Huang, Binbin and Yu, Zehao and Chen, Anpei and Geiger, Andreas and Gao, Shenghua},
  booktitle={ACM SIGGRAPH 2024 conference papers},
  pages={1--11},
  year={2024}
}

@article{chen2024pgsr,
  title={Pgsr: Planar-based gaussian splatting for efficient and high-fidelity surface reconstruction},
  author={Chen, Danpeng and Li, Hai and Ye, Weicai and Wang, Yifan and Xie, Weijian and Zhai, Shangjin and Wang, Nan and Liu, Haomin and Bao, Hujun and Zhang, Guofeng},
  journal={IEEE Transactions on Visualization and Computer Graphics},
  year={2024},
  publisher={IEEE}
}

@article{du2024mvgs,
  title={Mvgs: Multi-view-regulated gaussian splatting for novel view synthesis},
  author={Du, Xiaobiao and Wang, Yida and Yu, Xin},
  journal={arXiv preprint arXiv:2410.02103},
  year={2024}
}

@article{waczynska2024games,
  title={Games: Mesh-based adapting and modification of gaussian splatting},
  author={Waczy{\'n}ska, Joanna and Borycki, Piotr and Tadeja, S{\l}awomir and Tabor, Jacek and Spurek, Przemys{\l}aw},
  journal={arXiv preprint arXiv:2402.01459},
  year={2024}
}

@inproceedings{held2026triangle,
  title={Triangle splatting for real-time radiance field rendering},
  author={Held, Jan and Vandeghen, Renaud and Deliege, Adrien and Hamdi, Abdullah and Giancola, Silvio and Rebain, Daniel and Cioppa, Anthony and Ghanem, Bernard and Vedaldi, Andrea and Tagliasacchi, Andrea and others},
  booktitle={2026 International Conference on 3D Vision (3DV)},
  pages={1248--1257},
  year={2026},
  organization={IEEE}
}

@inproceedings{schoenberger2016sfm,
    author={Sch\"{o}nberger, Johannes Lutz and Frahm, Jan-Michael},
    title={Structure-from-Motion Revisited},
    booktitle={Conference on Computer Vision and Pattern Recognition (CVPR)},
    year={2016},
}

@inproceedings{schoenberger2016mvs,
    author={Sch\"{o}nberger, Johannes Lutz and Zheng, Enliang and Pollefeys, Marc and Frahm, Jan-Michael},
    title={Pixelwise View Selection for Unstructured Multi-View Stereo},
    booktitle={European Conference on Computer Vision (ECCV)},
    year={2016},
}

@article{mildenhall2021nerf,
  title={Nerf: Representing scenes as neural radiance fields for view synthesis},
  author={Mildenhall, Ben and Srinivasan, Pratul P and Tancik, Matthew and Barron, Jonathan T and Ramamoorthi, Ravi and Ng, Ren},
  journal={Communications of the ACM},
  volume={65},
  number={1},
  pages={99--106},
  year={2021},
  publisher={ACM New York, NY, USA}
}

@article{bobrow2023,
  title={Colonoscopy 3D video dataset with paired depth from 2D-3D registration},
  author={Bobrow, Taylor L and Golhar, Mayank and Vijayan, Rohan and Akshintala, Venkata S and Garcia, Juan R and Durr, Nicholas J},
  journal={Medical Image Analysis},
  pages={102956},
  year={2023},
  publisher={Elsevier},
}

@inproceedings{wu2023omniobject3d,
    author = {Tong Wu and Jiarui Zhang and Xiao Fu and Yuxin Wang and Jiawei Ren, 
    Liang Pan and Wayne Wu and Lei Yang and Jiaqi Wang and Chen Qian and Dahua Lin and Ziwei Liu},
    title = {OmniObject3D: Large-Vocabulary 3D Object Dataset for Realistic Perception, 
    Reconstruction and Generation},
    booktitle={IEEE/CVF Conference on Computer Vision and Pattern Recognition (CVPR)},
    year={2023}
}

@inproceedings{kato2018neural,
  title={Neural 3d mesh renderer},
  author={Kato, Hiroharu and Ushiku, Yoshitaka and Harada, Tatsuya},
  booktitle={Proceedings of the IEEE conference on computer vision and pattern recognition},
  pages={3907--3916},
  year={2018}
}

@inproceedings{liu2019soft,
  title={Soft rasterizer: A differentiable renderer for image-based 3d reasoning},
  author={Liu, Shichen and Li, Tianye and Chen, Weikai and Li, Hao},
  booktitle={Proceedings of the IEEE/CVF international conference on computer vision},
  pages={7708--7717},
  year={2019}
}

@inproceedings{ren2026fastgs,
  title={Fastgs: Training 3d gaussian splatting in 100 seconds},
  author={Ren, Shiwei and Wen, Tianci and Fang, Yongchun and Lu, Biao},
  booktitle={Proceedings of the IEEE/CVF Conference on Computer Vision and Pattern Recognition},
  pages={26094--26103},
  year={2026}
}

@article{held2025meshsplatting,
  title={Meshsplatting: Differentiable rendering with opaque meshes},
  author={Held, Jan and Son, Sanghyun and Vandeghen, Renaud and Rebain, Daniel and Gadelha, Matheus and Zhou, Yi and Cioppa, Anthony and Lin, Ming C and Van Droogenbroeck, Marc and Tagliasacchi, Andrea},
  journal={arXiv preprint arXiv:2512.06818},
  year={2025}
}

@article{tao2026oxford,
  title={The oxford spires dataset: Benchmarking large-scale lidar-visual localisation, reconstruction and radiance field methods},
  author={Tao, Yifu and Mu{\~n}oz-Ba{\~n}{\'o}n, Miguel {\'A}ngel and Zhang, Lintong and Wang, Jiahao and Fu, Lanke Frank Tarimo and Fallon, Maurice},
  journal={The International Journal of Robotics Research},
  volume={45},
  number={6},
  pages={839--857},
  year={2026},
  publisher={SAGE Publications Sage UK: London, England}
}

@article{liu2020general,
  title={A general differentiable mesh renderer for image-based 3D reasoning},
  author={Liu, Shichen and Li, Tianye and Chen, Weikai and Li, Hao},
  journal={IEEE Transactions on Pattern Analysis and Machine Intelligence},
  volume={44},
  number={1},
  pages={50--62},
  year={2020},
  publisher={IEEE}
}

@article{vizzo2022vdbfusion,
  title={Vdbfusion: Flexible and efficient tsdf integration of range sensor data},
  author={Vizzo, Ignacio and Guadagnino, Tiziano and Behley, Jens and Stachniss, Cyrill},
  journal={Sensors},
  volume={22},
  number={3},
  pages={1296},
  year={2022},
  publisher={MDPI}
}

@inproceedings{curless1996volumetric,
  author    = {Brian Curless and Marc Levoy},
  title     = {A Volumetric Method for Building Complex Models
               from Range Images},
  booktitle = {Proceedings of the 23rd Annual Conference on
               Computer Graphics and Interactive Techniques},
  pages     = {303--312},
  year      = {1996},
  doi       = {10.1145/237170.237269}
}

@article{dai2017bundlefusion,
  author  = {Dai, Angela and Nie{\ss}ner, Matthias and Zollh{\"o}fer, Michael and Izadi, Shahram and Theobalt, Christian},
  title   = {{BundleFusion}: Real-Time Globally Consistent 3D Reconstruction Using On-the-Fly Surface Re-Integration},
  journal = {ACM Transactions on Graphics},
  volume  = {36},
  number  = {4},
  pages   = {76:1--76:18},
  year    = {2017},
  doi     = {10.1145/3072959.3054739}
}

@inproceedings{oleynikova2017voxblox,
  author    = {Oleynikova, Helen and Taylor, Zachary and Fehr, Marius and Nieto, Juan and Siegwart, Roland},
  title     = {{Voxblox}: Incremental 3D Euclidean Signed Distance Fields for On-Board {MAV} Planning},
  booktitle = {2017 IEEE/RSJ International Conference on Intelligent Robots and Systems (IROS)},
  pages     = {1366--1373},
  year      = {2017},
  doi       = {10.1109/IROS.2017.8202315}
}

@article{liu2023large,
  title={Large-scale LiDAR consistent mapping using hierarchical LiDAR bundle adjustment},
  author={Liu, Xiyuan and Liu, Zheng and Kong, Fanze and Zhang, Fu},
  journal={IEEE Robotics and Automation Letters},
  volume={8},
  number={3},
  pages={1523--1530},
  year={2023},
  publisher={IEEE}
}

@article{kingma2014adam,
  title={Adam: A method for stochastic optimization},
  author={Kingma, Diederik P and Ba, Jimmy},
  journal={arXiv preprint arXiv:1412.6980},
  year={2014}
}

@inproceedings{hanson2025speedy,
  title={Speedy-splat: Fast 3d gaussian splatting with sparse pixels and sparse primitives},
  author={Hanson, Alex and Tu, Allen and Lin, Geng and Singla, Vasu and Zwicker, Matthias and Goldstein, Tom},
  booktitle={2025 IEEE/CVF Conference on Computer Vision and Pattern Recognition (CVPR)},
  pages={21537--21546},
  year={2025},
  organization={IEEE}
}

@inproceedings{mallick2024taming,
  title={Taming 3dgs: High-quality radiance fields with limited resources},
  author={Mallick, Saswat Subhajyoti and Goel, Rahul and Kerbl, Bernhard and Steinberger, Markus and Carrasco, Francisco Vicente and De La Torre, Fernando},
  booktitle={SIGGRAPH Asia 2024 Conference Papers},
  pages={1--11},
  year={2024}
}

@inproceedings{azinovic2022neural,
  title     = {Neural RGB-D Surface Reconstruction},
  author    = {Azinovi{\'c}, Dejan and Martin-Brualla, Ricardo and Goldman, Dan B. and Nie{\ss}ner, Matthias and Thies, Justus},
  booktitle = {Proceedings of the IEEE/CVF Conference on Computer Vision and Pattern Recognition (CVPR)},
  year      = {2022},
  pages     = {6290--6301}
}

@inproceedings{yang2023unisim,
  title     = {UniSim: A Neural Closed-Loop Sensor Simulator},
  author    = {Yang, Ze and Chen, Yun and Wang, Jingkang and Manivasagam, Sivabalan and Ma, Wei-Chiu and Yang, Anqi Joyce and Urtasun, Raquel},
  booktitle = {Proceedings of the IEEE/CVF Conference on Computer Vision and Pattern Recognition (CVPR)},
  year      = {2023},
  pages     = {1389--1399}
}

@article{adamkiewicz2022vision,
  title   = {Vision-Only Robot Navigation in a Neural Radiance World},
  author  = {Adamkiewicz, Michal and Chen, Timothy and Caccavale, Adam and Gardner, Rachel and Culbertson, Preston and Bohg, Jeannette and Schwager, Mac},
  journal = {IEEE Robotics and Automation Letters},
  volume  = {7},
  number  = {2},
  pages   = {4606--4613},
  year    = {2022},
  doi     = {10.1109/LRA.2022.3150497}
}

@article{wisth2022vilens,
  title={Vilens: Visual, inertial, lidar, and leg odometry for all-terrain legged robots},
  author={Wisth, David and Camurri, Marco and Fallon, Maurice},
  journal={IEEE Transactions on Robotics},
  volume={39},
  number={1},
  pages={309--326},
  year={2022},
  publisher={IEEE}
}

@inproceedings{ruan2023slamesh,
  title={Slamesh: Real-time lidar simultaneous localization and meshing},
  author={Ruan, Jianyuan and Li, Bo and Wang, Yibo and Sun, Yuxiang},
  booktitle={2023 IEEE International Conference on Robotics and Automation (ICRA)},
  pages={3546--3552},
  year={2023},
  organization={IEEE}
}

@article{jia2025cad,
  title={Cad-mesher: A convenient, accurate, dense mesh-based mapping module in slam for dynamic environments},
  author={Jia, Yanpeng and Cao, Fengkui and Wang, Ting and Tang, Yandong and Shao, Shiliang and Liu, Lianqing},
  journal={IEEE Transactions on Multimedia},
  year={2025},
  publisher={IEEE}
}

@article{tian2022kimera,
  title={Kimera-multi: Robust, distributed, dense metric-semantic slam for multi-robot systems},
  author={Tian, Yulun and Chang, Yun and Arias, Fernando Herrera and Nieto-Granda, Carlos and How, Jonathan P and Carlone, Luca},
  journal={IEEE transactions on robotics},
  volume={38},
  number={4},
  year={2022},
  publisher={IEEE}
}

@article{liu2024dynameshslam,
  title={DynaMeshSLAM: A mesh-based dynamic visual SLAMMOT method},
  author={Liu, Yang and Guo, Chi and Luo, Yarong and Wang, Yingli},
  journal={IEEE Robotics and Automation Letters},
  volume={9},
  number={6},
  pages={5791--5798},
  year={2024},
  publisher={IEEE}
}

@inproceedings{lee2024compact,
  title     = {Compact 3D Gaussian Representation for Radiance Field},
  author    = {Lee, Joo Chan and Rho, Daniel and Sun, Xiangyu
               and Ko, Jong Hwan and Park, Eunbyung},
  booktitle = {Proceedings of the IEEE/CVF Conference on Computer Vision
               and Pattern Recognition (CVPR)},
  pages     = {21719--21728},
  year      = {2024},
  doi       = {10.1109/CVPR52733.2024.02052}
}

@inproceedings{fan2024lightgaussian,
  title     = {LightGaussian: Unbounded 3D Gaussian Compression with
               15x Reduction and 200+ FPS},
  author    = {Fan, Zhiwen and Wang, Kevin and Wen, Kairun and Zhu, Zehao
               and Xu, Dejia and Wang, Zhangyang},
  booktitle = {Advances in Neural Information Processing Systems},
  volume    = {37},
  pages     = {140138--140158},
  year      = {2024}
}

@inproceedings{pehlivan2026matrixfree,
  title     = {Matrix-Free Second-Order Optimization of Gaussian Splats
               with Residual Sampling},
  author    = {Pehlivan, Hamza and Boscolo Camiletto, Andrea
               and Foo, Lin Geng and Habermann, Marc
               and Theobalt, Christian},
  booktitle = {International Conference on 3D Vision (3DV)},
  pages     = {18--28},
  year      = {2026},
  doi       = {10.1109/3DV69130.2026.00009}
}

@inproceedings{lan2025_3dgs2,
  title     = {3DGS$^2$: Near Second-order Converging 3D Gaussian Splatting},
  author    = {Lan, Lei and Shao, Tianjia and Lu, Zixuan and Zhang, Yu
               and Jiang, Chenfanfu and Yang, Yin},
  booktitle = {SIGGRAPH Conference Papers},
  year      = {2025},
  articleno = {99},
  numpages  = {10},
  doi       = {10.1145/3721238.3730687}
}

@inproceedings{hollein2025_3dgslm,
  title     = {3DGS-LM: Faster Gaussian-Splatting Optimization
               with Levenberg-Marquardt},
  author    = {H{\"o}llein, Lukas and Bo\v{z}i\v{c}, Alja\v{z}
               and Zollh{\"o}fer, Michael and Nie{\ss}ner, Matthias},
  booktitle = {Proceedings of the IEEE/CVF International Conference
               on Computer Vision (ICCV)},
  pages     = {26740--26750},
  year      = {2025},
  doi       = {10.1109/ICCV51701.2025.02482}
}

@article{hsiao2026_3dgs2tr,
  title   = {3DGS$^2$-TR: Scalable Second-Order Trust-Region Method
             for 3D Gaussian Splatting},
  author  = {Hsiao, Roger and Fang, Yuchen and Huang, Xiangru
             and Li, Ruilong and Rabeti, Hesam and Gojcic, Zan
             and Lavaei, Javad and Demmel, James and Shao, Sophia},
  journal = {arXiv preprint arXiv:2602.00395},
  year    = {2026}
}

@inproceedings{huang2026opt3dgs,
  title     = {Opt3DGS: Optimizing 3D Gaussian Splatting with Adaptive
               Exploration and Curvature-Aware Exploitation},
  author    = {Huang, Ziyang and Chen, Jiagang and Liu, Jin and Ji, Shunping},
  booktitle = {Proceedings of the AAAI Conference on Artificial Intelligence},
  volume    = {40},
  number    = {7},
  pages     = {5230--5238},
  year      = {2026},
  doi       = {10.1609/aaai.v40i7.37438}
}

@article{sharma2005ciede2000,
  author  = {Sharma, Gaurav and Wu, Wencheng and Dalal, Edul N.},
  title   = {The CIEDE2000 Color-Difference Formula: Implementation Notes,
             Supplementary Test Data, and Mathematical Observations},
  journal = {Color Research \& Application},
  volume  = {30},
  number  = {1},
  pages   = {21--30},
  year    = {2005},
  doi     = {10.1002/col.20070}
}

@inproceedings{wood2000surface,
  author    = {Wood, Daniel N. and Azuma, Daniel I. and Aldinger, Ken
               and Curless, Brian and Duchamp, Tom and Salesin, David H.
               and Stuetzle, Werner},
  title     = {Surface Light Fields for 3D Photography},
  booktitle = {Proceedings of the 27th Annual Conference on Computer
               Graphics and Interactive Techniques},
  pages     = {287--296},
  year      = {2000},
  doi       = {10.1145/344779.344925}
}

@inproceedings{chen2002light,
  author    = {Chen, Wei-Chao and Bouguet, Jean-Yves and Chu, Michael H.
               and Grzeszczuk, Radek},
  title     = {Light Field Mapping: Efficient Representation and Hardware
               Rendering of Surface Light Fields},
  booktitle = {Proceedings of the 29th Annual Conference on Computer
               Graphics and Interactive Techniques},
  pages     = {447--456},
  year      = {2002},
  doi       = {10.1145/566570.566601}
}

@inproceedings{palma2013surface,
  author    = {Palma, Gianpaolo and Desogus, Nicola and Cignoni, Paolo
               and Scopigno, Roberto},
  title     = {Surface Light Field from Video Acquired in Uncontrolled Settings},
  booktitle = {2013 Digital Heritage International Congress (DigitalHeritage)},
  volume    = {1},
  pages     = {31--38},
  year      = {2013},
  publisher = {IEEE},
  doi       = {10.1109/DigitalHeritage.2013.6743710}
}

@article{vanhoey2013robust,
  author  = {Vanhoey, Kenneth and Sauvage, Basile and
             G{\'e}nevaux, Olivier and Larue, Fr{\'e}d{\'e}ric and
             Dischler, Jean-Michel},
  title   = {Robust Fitting on Poorly Sampled Data for Surface Light Field
             Rendering and Image Relighting},
  journal = {Computer Graphics Forum},
  volume  = {32},
  number  = {6},
  pages   = {101--112},
  year    = {2013},
  doi     = {10.1111/cgf.12073}
}

\end{document}